%% file: main.tex
\documentclass{biometrika}
\usepackage{amsmath}
\usepackage{newtxtext}
\usepackage[subscriptcorrection]{newtxmath}
\usepackage[subscriptcorrection]{newtxmath}
\usepackage{subcaption}

\graphicspath{{./art/}}

\usepackage{natbib}
\usepackage{preamble}
\usepackage{enumitem}
\usepackage[section]{placeins}
\usepackage{diagbox}
\usepackage{caption}     
\usepackage{graphicx}
\newcommand{\xapprox}[1]{%
  \mathrel{\overset{#1}{\scalebox{1.8}[1]{$\approx$}}}%
}

\newcommand{\ignore}[1]{}

\newcommand{\eps}{\mathrm{eps}}

\renewcommand{\cS}{\mathbb{S}}
\renewcommand{\cO}{\mathbb{O}}

\usepackage[ruled,vlined]{algorithm2e}
\DontPrintSemicolon
\SetKwInput{KwInput}{Input}
\SetKwInput{KwOutput}{Output}

\usepackage{tcolorbox}
\newtcolorbox{mybox}{
    colback=pink!20, 
    colframe=black,  
    fonttitle=\bfseries, 
    coltitle=black,
    boxrule=0.5pt,  
    arc=3pt,      
    left=5pt,     
    right=5pt,    
    top=5pt,      
    bottom=5pt,   
}

\makeatletter
\def\tableofcontents{%
  \section*{Contents}
  \@starttoc{toc}
}
\makeatother

\usepackage[
  colorlinks=true,
  linkcolor=blue,
  citecolor=purple,
  urlcolor=blue,
  pdfborder={0 0 0}
]{hyperref}

\begin{document}
    \linespread{1.3}\selectfont

    \title{Test for partial effects in Fr\'echet regression on Bures-Wasserstein manifolds}
    \author{Haoshu Xu and Hongzhe Li\\
    University of Pennsylvania}
    \date{}

    \maketitle


    \begin{abstract}
      We propose a novel test for assessing partial effects in Fr\'echet regression with responses lying on the Bures-Wasserstein manifold. 
      Under the null hypothesis, we show that the statistic admits a degenerate V-statistic approximation whose limiting distribution is a weighted mixture of chi-squared random variables, with weights determined by the eigenvalues of an integral operator associated with a reproducing kernel Hilbert space (RKHS) kernel.
      We establish the asymptotic validity and consistency of the proposed test. Its finite-sample performance is examined through simulation studies. 
      We apply the proposed test to study the effect of age, while controlling for other covariates, on gene co-expression structure in single-cell data.\\

      \noindent \textit{Some key words:} Fr\'echet differentials; Optimal transport; 
      Reproducing kernel Hilbert space;  V-statistic; Riemannian  gradient descent; Single cell genomics      .
    \end{abstract}
 \section{Introduction}
    \label{sec:introduction}
    \input{intro.tex}
    
    \section{Fr\'echet regression on the Bures--Wasserstein manifold}
    \label{sec: formulation}
    \input{formulation.tex}

    \section{Test Statistic and Asymptotic Theory}
    \label{sec:asymptotic}
    \input{test.tex}

    \section{Numerical simulations}\label{sec: simulation}
    \input{simulation.tex}

    \section{Application to single cell co-expression analysis}\label{sec: application}
    \input{singlecell.tex}

\section*{Acknowledgment}
This research is supported by NIH grants. We thank the reviewers for many helpful comments. 

 \section*{Supplementary material} 
The Supplementary Material contains simulation results for the sample-splitting test, detailed descriptions of the genes and samples included in the single-cell application, complete proofs of all theorems, and additional technical lemmas.

    \vskip 0.2in
    \bibliographystyle{biometrika}
    \bibliography{bib}

\newpage
\clearpage

\setcounter{page}{1}

\renewcommand{\thepage}{S\arabic{page}}
\setcounter{page}{1}

\renewcommand{\thepage}{S\arabic{page}}


\setcounter{figure}{0}

\setcounter{table}{0}


\renewcommand{\thefigure}{S\arabic{figure}}

\renewcommand{\thetable}{S\arabic{table}}
   \section*{Supplementary material}
   
     
    \appendix

    \section{Additional simulations}
    \label{sec: appendix_simulation}
    \input{simulation_appdx.tex}

    \section{Additional single-cell data analysis}
    \label{sec: appendix_singlecell}

\input{singlecell_appdx.tex}

    \section{Proof of Theorem \ref{thm: est}}
    \label{sec: prf_est}
    \input{prf_est.tex}
    
    \section{Proof of Theorem \ref{thm: null}}
    \label{sec: prf_null}
    \input{prf_null.tex}

    \section{Proof of Corrolary \ref{cor: glb_null}}
    \label{sec: prf_glb_null}
    \input{global_null.tex}

    \section{Proof of Theorem \ref{thm: size}}
    \label{sec: prf_size}
    \input{prf_size.tex}

    \section{Proof of Theorem \ref{thm: power}}
    \label{sec: prf_power}
    \input{prf_power.tex}


    \section{Proof of Proposition \ref{prop: example}}
    \label{sec: prf_example}
    \input{prf_example.tex}

    \section{Technical lemmas}
    \label{sec: tech_lemmas}
    \input{tech_lemmas.tex}

\end{document}

%% file: intro.tex
In many modern applications, positive definite matrices are often used to summarize the marginal covariance structure among sets of variables. Examples include  medical imaging \citep{dryden09, fillard07}, neuroscience \citep{friston11_brain,kong20,kong21} and gene coexpression analysis in single cell genomics.  A central challenge in these fields is how to perform regression analysis where the covariance matrix serves as the outcome variable in relation to a set of Euclidean covariates and how to test for the association between these matrix and covariates. 

Several regression approaches for covariance-matrix outcomes have been proposed. \citet{Chiu1996} model the elements of the log-covariance matrix as a linear function of the covariates, though at the cost of estimating a large number of parameters; \citet{HoffNiu} express the covariance matrix as a quadratic function of the covariates; and \citet{Zou2017} link the matrix outcome to a linear combination of covariate-derived similarity matrices, studying the asymptotic properties of the resulting estimators. Each of these imposes a specific parametric form on how the covariance matrix varies with the covariates. Fr\'echet regression instead of positing a parametric form, defines the regression through the geometry of the response space. \citet{XuLi2025} studied Fr\'echet regression with a covariance-matrix response under the Bures--Wasserstein metric---a special case of the general framework of \citet{muller16}---and developed a formal test for the overall association between covariance matrices and a Euclidean vector of covariates. Although a covariance matrix can be identified with a centred Gaussian distribution, the Wasserstein $F$-test of \citet{petersen21_WFtest} does not carry over to this setting: it is restricted to univariate distributions, whose $2$-Wasserstein geometry embeds isometrically into a flat Hilbert space of quantile functions, whereas the Bures--Wasserstein manifold of $d\times d$ matrices is curved for $d\ge2$ and admits no such embedding.

The test of \citet{XuLi2025}, however, targets only the global null of no covariate effect. In many applications one instead needs to test the effect of a single predictor, or a subset, while conditioning on the remaining covariates or confounders. For example, in single-cell analysis one may wish to test the association between the gene co-expression matrix and a covariate of interest while conditioning on the proportion of a particular cell type. In this paper, we develop such a test of partial effects, extending the framework of \citet{XuLi2025}. 
Under the null hypothesis, our test statistic admits a degenerate V-statistic approximation whose limiting distribution is a weighted mixture of chi-squared random variables, with weights determined by the eigenvalues of an integral operator associated with an RKHS kernel. We establish the asymptotic validity and consistency of the proposed test.

Throughout the paper, 
we use $[n]$ to denote~$\cbr{1,\ldots,n}$.
Given any vector $x \in \RR^p$, we let $\vx=(1, x^\top)^\top \in \RR^{p+1}$.
We denote by $\cS^d, \cS^d_+, \cS^d_{++}$ the sets of all $d \times d$ symmetric, positive semi-definite, and positive definite matrices, respectively. For any real numbers $a<b$, we define $\cS^d(a,b):= \cbr{S \in \cS^d: a I_d \prec S \prec b I_d}$. When the dimension $d$ is clear from context, we omit the subscript for simplicity. We use $\Fnorm{\cdot}$ to denote the matrix Frobenius norm and $\inner{\cdot}{\cdot}_{\mathrm{F}}$ to denote the matrix Frobenius inner product.

%% file: formulation.tex
\subsection{General formulation}
\label{subsec: model formulation}
The Fr\'echet regression model \citep{muller16} extends classical linear regression to responses taking values in general metric spaces by relying solely on the geometry induced by a distance, rather than a linear structure. A prominent example arises when the response lies in the Bures-Wasserstein manifold of symmetric positive-definite (SPD) matrices $\cS^d_{++}$, equipped with the Wasserstein distance
\begin{align}
    W(A,B) = \sbr[\big]{\tr{A} + \tr{B} - 2 \tr \rbr[\big]{(A^{1/2}BA^{1/2})^{1/2}}}^{1/2}, \quad \text{for any } A,B \in \cS^d_{++}.
    \label{eqn: wasserstein_distance}
\end{align}
Let $(X,Q) \in \RR^p \times \cS^d_{++}$ denote a random covariate-response pair. The model is based on the conditional Fr\'echet mean, defined as
\begin{align*}
    \EE_{\text{Fr\'echet}} \sbr{Q| X=x} := \argmin_{S \in \cS^d_{++}} \EE_{Y|X=x} \sbr{W^2(S,Q) | X=x}.
\end{align*}
The following result, due to \citet{aap21}, guarantees existence and uniqueness under mild moment assumptions.
\begin{lemma} \label{lem: exist}
    On the Bures-Wasserstein manifold $(\cS^d_{++},W)$, the following hold: (i) If \ $\EE \tr Q < +\infty$, then the Fr\'echet mean $\EE_{\text{\normalfont Fr\'echet}} Q$ exists uniquely. (ii) If \ $\EE[\tr Q | X=x] < +\infty$, then the conditional Fr\'echet mean $\EE_{\text{\normalfont Fr\'echet}} \sbr{Q | X=x}$ exists uniquely.
\end{lemma}
We write 
\begin{align*}
    Q^*(x):= \EE_{\mathrm{\text{Fr\'echet}}} \sbr{Q|X=x}.
\end{align*}
Fr\'echet regression models $Q^*(x)$ as the minimizer of a weighted Fr\'echet objective,
\begin{equation}
    Q^*(x) = \argmin_{S \in \cS^d_{++}} F(x,S), \qquad F(x,S):=\EE_{(X,Q)} \sbr[\big]{w(x,X) W^2(S,Q)},
    \label{eqn: frechet_assumption}
\end{equation}
where the bilinear weight function $w: \RR^p \times \RR^p \to \RR$ is given by
\begin{equation*}
    w(x, X)=1+(x-\mu)^{\top} \Sigma^{-1}(X-\mu), \quad \mu=\EE X, \;\ \Sigma=\operatorname{Var}(X).
\end{equation*}
Although $Q^*(x)$ does not admit a closed-form expression in general, it reduces to linear regression on matrix square roots when $Q$ is supported on commuting matrices \citep{XuLi2025}.

Given observations $\cbr{(X_i,Q_i)}_{i=1}^n$, we estimate $Q^*(x)$ by the Fr\'echet estimator
\begin{align}
    \label{eqn: estimator}
    \hQ_n(x) := \argmin_{S \in \cS^d_{++}} F_n(x,S), \qquad F_n(x,S):=\frac{1}{n} \sum_{i=1}^n w_n(x,X_i) W^2(S,Q_i),
\end{align}
where
\begin{align*}
    w_n(x,X_i) := 1 + (x-\hmu)^{\top} \hSigma^{-1}(X_i-\hmu), \quad \hmu = n^{-1} \sum_{i=1}^n X_i, \quad \hSigma = n^{-1} \sum_{i=1}^n (X_i - \hmu)(X_i - \hmu)^\top.
\end{align*}
\citet{XuLi2025} established uniform convergence of this estimator to $Q^*(x)$.
Based on this estimator, they proposed a statistical test for the global null hypothesis
\begin{align}
    \cH_{\text{global}}: Q(x)\equiv Q^*_0 \;\ \text{for some unknown} \;\ Q_0^* \in \cS^d_{++},
    \label{eqn: global_null}
\end{align}
under which the asymptotic null distribution of the test statistic is a weighted sum of independent chi-square random variables.

\subsection{Partial-effect setup}
We focus on assessing partial effect by partitioning the covariate vector as  $X=(Y,Z)$, where $Y \in \RR^{p_1}$ denotes covariates to be adjusted for and $Z \in \RR^{p-p_1}$ represents the covariate of interest.
Accordingly, the mean vector and covariance matrix of $X$ admit the block
decompositions
\begin{align*}
    \mu = \begin{pmatrix}
        \mu_Y \\
        \mu_Z   
    \end{pmatrix}, \quad
    \Sigma = \begin{pmatrix}
        \Sigma_{YY} & \Sigma_{YZ} \\
        \Sigma_{ZY} & \Sigma_{ZZ}
    \end{pmatrix}.
\end{align*}
We use the same block notation for the sample mean $\hmu$ and sample covariance matrix $\hSigma$.

To test whether the component $Z$ has any effect on the conditional Fr\'echet mean of $Q$ beyond that explained by $Y$, we consider the null hypothesis
\begin{equation}
H_0:\quad Q^*(y,z) = Q^*(y,z') \quad \text{for all } y\in\mathbb{R}^{p_1} \text{ and all } z,z'\in\mathbb{R}^{p-p_1}. \label{eqn: null}
\end{equation}
Under the null \eqref{eqn: null}, it is natural to consider an estimator that depends only on the covariate $Y$,
\begin{align}
    \tQ_n(y):= \argmin_{S \in \cS^d_{++}} \frac{1}{n} \sum_{i=1}^n \tw_n(y,Y_i) W^2(S,Q_i),
    \label{eqn: tqn}
\end{align}
where the associated weight function $\tw_n: \RR^{p_1} \times \RR^{p_1} \to \RR$ is defined by
\begin{align}
    \tw_n(y,Y):= 1 + (y-\hmu_Y)^\top \hSigma_{YY}^{-1}(Y-\hmu_Y).
\end{align}
The estimator $\tQ_n(y)$ is a natural analogue of \eqref{eqn: estimator} that involves only the covariate $Y$. It can be viewed as an M-estimator of the population quantity
\begin{align}
    \tQ^*(y) := \argmin_{S \in \cS^d_{++}} \tF(y,S), \quad \text{where } \tF(y,S):=\EE_{(Y,Q)} \sbr{\tw(y,Y) W^2(S,Q)}.
    \label{eqn: tQ_tF}
\end{align}
Here the corresponding population weight function $\tw: \RR^{p_1} \times \RR^{p_1} \to \RR$ is given by
\begin{align}
    \tw(y,Y):= 1 + (y-\mu_Y)^\top \Sigma_{YY}^{-1}(Y-\mu_Y).
    \label{eqn: wt_y}
\end{align}

To further justify the introduction of $\tQ_n(y)$, we establish the following connection between~$Q^*$ and~$\tQ^*$, and between $\hQ_n$ and $\tQ_n$.

\begin{property}
    \label{prpt: sub_model}
    For any covariate value $x=(y,z)\in \RR^p$, define
    \begin{align}
        \tx:= \rbr{y, \tz} \in \RR^p, \quad \text{where } \tz := \mu_Z + \Sigma_{ZY} \Sigma_{YY}^{-1} (y-\mu_Y);
        \label{eqn: tx}
        \\
        \hx:= \rbr{y, \hz} \in \RR^p, \quad \text{where } \hz := \hmu_Z + \hSigma_{ZY} \hSigma_{YY}^{-1} (y-\hmu_Y).
        \label{eqn: hx}
    \end{align}
    Then for any $x=(y,z), X=(Y,Z) \in \RR^p$,
    \begin{align*}
        w(x, X) = \tw(y, Y), \qquad w_n(x, X) = \tw_n(y, Y).
    \end{align*}
    Consequently,
    \begin{align*}
        Q^*(\tx) = \tQ^*(y),
        \qquad
        \hQ_n(\hx) = \tQ_n(y).
    \end{align*}
\end{property}

Property~\ref{prpt: sub_model} is a direct consequence of Lemma~\ref{lem: matrix}. It is purely algebraic and does not require any distributional assumptions. 
In particular, it shows that $\tQ^*(y)$ is well defined under the Fr\'echet regression model for the full covariate vector $X$, and that $\tQ_n(y)$ converges to $\tQ^*(y)$ under the same assumptions imposed on the original estimator $\hQ_n(x)$; see Theorem~\ref{thm: est}.



\subsection{Optimal transport maps and differentials}
\label{sec:ot_diff}
The Fr\'echet estimator \eqref{eqn: estimator} is an M-estimator defined on the Bures-Wasserstein manifold $(\cS^d_{++},W)$. To study its asymptotic properties, we require notions of first- and second-order differentials for functions defined on this manifold. Since $\cS^d_{++}$ is an open subset of the normed linear space $(\cS^d,\Fnorm{\cdot})$, these differentials can be formulated using standard functional calculus on Banach spaces. We therefore provide a concise, self-contained introduction in this section; for additional background, see \citet{dudley}.

Let $(\cV,\norm{\cdot}_{\cV})$ be a normed linear space and denote by $L(\cS^d,\cV)$ the space of bounded linear operators from $\cS^d$ to $\cV$. A function $f:\cS^d_{++}\to\cV$ is said to be Fr\'echet \emph{differentiable} at $S\in\cS^d_{++}$ if there exists a bounded linear operator $df(S)\in L(\cS^d,\cV)$ such that
 \begin{align*}
    \lim_{\substack{H\to 0\\ H\in\cS^d, S+H\in\cS^d_{++}}}
    \frac{\norm{f(S+H)-f(S)-df(S)(H)}_{\cV}}{\Fnorm{H}} = 0 .
\end{align*}
The operator $df(S)$ is called the (Fr\'echet) \emph{derivative} of $f$ at $S$.
If $f$ is Fr\'echet differentiable at every $S\in\cS^d_{++}$, then the map $df:\cS^d_{++}\to L(\cS^d,\cV)$ is called the Fr\'echet derivative of $f$.

For $k=1,2,\ldots$, define recursively
\begin{align*}
    L^{k+1}(\cS^d,\cV):=L(\cS^d,L^k(\cS^d,\cV)),
\end{align*}
equipped with the operator norm. For $k\ge2$, the function $f$ is said to be (Fr\'echet) \emph{differentiable of order $k$} if $f$ is differentiable of order $k-1$ and the map $d^{k-1}f:\cS^d_{++}\to L^{{k-1}}(\cS^d,\cV)$ is Fr\'echet differentiable. The resulting $k$th-order derivative is denoted by $d^k f$.

We now consider the squared Wasserstein distance. For any fixed $Q\in\cS^d_{++}$, the map
\begin{align*}
    W^2(\cdot,Q): \cS^d_{++} \to \RR
\end{align*}
is Fr\'echet differentiable; see Lemma~\ref{lem: Q} in Appendix~\ref{sec: tech_lemmas}. Its derivative at $S\in\cS^d_{++}$ satisfies
\begin{align*}
    dW^2(S,Q)(H) = \inner[\big]{I_d - T_S^Q}{H}_{\mathrm F}, \qquad \forall H\in\cS^d,
\end{align*}
where $\inner{\cdot}{\cdot}_{\mathrm{F}}$ denotes the Frobenius inner product, and $T_S^Q\in\cS^d$ is the optimal transport map from the centered Gaussian distribution $N(0,S)$ to $N(0,Q)$, given explicitly by
\begin{align*}
    T_S^Q
    = Q^{1/2}\rbr[\big]{Q^{1/2} S Q^{1/2}}^{-1/2}Q^{1/2}
    = S^{-1/2}\rbr[\big]{S^{1/2} Q S^{1/2}}^{1/2}S^{-1/2}.
\end{align*}
Thus, the derivative of $W^2(\cdot,Q)$ at $S$ can be identified with the matrix
\begin{align}
    d W^2(S,Q) = I_d - T_S^Q \in \cS^d.
    \label{eqn: dW2}
\end{align}

For any fixed $Q\in\cS^d_{++}$, the map $T_{\cdot}^Q:\cS^d_{++}\to\cS^d$ is itself Fr\'echet differentiable. Its derivative at $S$, denoted by $dT_S^Q$, is a linear operator on $\cS^d$. Although its explicit expression is involved, it is well defined; see Lemma~\ref{lem: Q} in Appendix~\ref{sec: tech_lemmas} for details. Consequently, the squared Bures-Wasserstein distance $W^2(\cdot,Q)$ is twice Fr\'echet differentiable, with second derivative given~by
\begin{align}
    d^2 W^2(S,Q) = - dT_S^Q.
    \label{eqn: d2W2}
\end{align}

It follows that, under mild regularity conditions, the Fr\'echet objective functions $F(x,\cdot)$ and $F_n(x,\cdot)$ defined in~\eqref{eqn: frechet_assumption} and~\eqref{eqn: estimator} are twice Fr\'echet differentiable for each fixed $x$. Their first and second derivatives at $S\in\cS^d_{++}$ are given by
\begin{equation}
    \begin{aligned}
        dF(x,S) &= \EE_{(X,Q)} \sbr[\big]{w(x,X) \rbr[\big]{I_d - T_S^Q}}, \qquad
        d^2 F(x,S) = - \EE_{(X,Q)} \sbr[\big]{w(x,X) dT_S^Q}, \\
        dF_n(x,S) &= \frac{1}{n} \sum_{i=1}^n w_n(x,X_i) \rbr[\big]{I_d - T_S^{Q_i}}, \qquad
        d^2 F_n(x,S) = - \frac{1}{n} \sum_{i=1}^n w_n(x,X_i) dT_S^{Q_i}.
    \end{aligned}
    \label{eqn: d12F}
\end{equation}

In particular, since $w(\mu,X)\equiv 1$ and $w_n(\hmu,X)\equiv 1$ by
\eqref{eqn: frechet_assumption} and~\eqref{eqn: estimator}, we obtain
\begin{align}
    d^2 F(\mu,S) &= - \EE_{Q} \sbr[\big]{dT_S^Q}, \qquad 
    d^2 F_n(\hmu,S) = - \frac{1}{n} \sum_{i=1}^n dT_S^{Q_i}.
    \label{eqn: d2_mu}
\end{align}
The population second differential $d^2F(x,S)$ appears in Assumption~\ref{assumption: minimizer_local}, where we impose a positive-definiteness condition at
the minimizer $Q^*(x)$. Its empirical counterpart $d^2F_n(x,S)$ is used to construct the proposed
partial-effect test statistic $\cT_n$, while the special case $d^2F_n(\hmu,S)$ enters the global test
statistic $\cT_{\text{global}}$ defined in Section~\ref{sec: test_statistic}.

\subsection{Assumptions}
Recall that $\cS^d, \cS^d_{++}$ and $\cS^d(a,b)$ denote the sets of $d \times d$ symmetric, positive-definite matrices and symmetric matrices with eigenvalues in $(a,b)$, respectively, as defined in Section~\ref{sec:introduction}.

The assumptions required for partial-effect testing are identical to those used to establish consistency and validity of the global test in \citet{XuLi2025}, and are summarized here for completeness.
Assumption~\ref{assumption: X} and~\ref{assumption: bdd_Q} impose a light-tail condition on the covariate $X$ and a boundedness condition on the response $Q$, respectively.
Assumption~\ref{assumption: minimizer_global} and~\ref{assumption: minimizer_local}  specify global and local regularity conditions for the objective function $F(x,\cdot)$ defined in \eqref{eqn: frechet_assumption}, in a neighborhood of its minimizer~$Q^*(x)$.
For detailed discussions and technical justifications, see \citet{XuLi2025}.
\begin{assumption}
    The covariate vector $X$ is sub-Gaussian with sub-Gaussian norm satisfying $\norm{X}_{\psi_2}\leq C_{\psi_2}$, and its covariance matrix satisfies $\lambda_{\min}(\Sigma)\geq c_{\Sigma}$ for constants $C_{\psi_2},c_{\Sigma} >0$.
    \label{assumption: X}
\end{assumption}

\begin{assumption}
    Given $X=x \in \supp(X)$, the eigenvalues of $Q$ are bounded away from $0$ and infinity  in the sense that
    \begin{equation*}
        \PP\rbr{Q \in \cS^d \rbr[\big]{\gamma_{\Lambda}(\norm{x-\mu})^{-1},\gamma_{\Lambda}(\norm{x-\mu})} | X=x} = 1,
    \end{equation*}
    where
    $\gamma_{\Lambda}(t) := c_\Lambda \rbr{t \vee 1}^{C_\Lambda}$ for constants $c_\Lambda \geq 1$ and $C_\Lambda \geq 0$.
    \label{assumption: bdd_Q}
\end{assumption}

\begin{assumption}
    The Fr\'echet regression model~\eqref{eqn: frechet_assumption} holds for all $x \in \text{supp}(X)$.
    \label{assumption: frechet}
\end{assumption}

The estimator $\hQ_n(x)$ is an M-estimator obtained by minimizing the empirical
objective function associated with $F(x,\cdot)$. Accordingly, Assumption \ref{assumption: minimizer_global} concerns the global behavior of $F(x,\cdot)$ outside local neighborhoods of its minimizer $Q^*(x)$ for each $x$, and is motivated by the well-separated-minimizer condition commonly used in M-estimation. 

\begin{assumption}
    \label{assumption: minimizer_global}
    For any $x\in \supp(X)$, the function $F(x,\cdot):\cS^d_{++}\to \RR$ has a unique minimizer. Moreover, there exist constants $\alpha_F \geq 1$ and $\delta_F > 0$ such that for any $x \in \supp(X)$ and any $0 \leq \delta\leq \delta_F \leq \Delta$,
    \begin{equation*}
        \inf \cbr{F(x,S) - F(x,Q^*(x)): \delta \leq \Fnorm{S-Q^*(x)} \leq \Delta} \geq \frac{\delta^{\alpha_F}}{\gamma_F(\norm{x-\mu},\Delta)}\ ,
    \end{equation*}
    where $\gamma_F(t_1,t_2) := c_F \rbr{t_1 \vee 1}^{C_F} \rbr{t_2 \vee 1}^{C_F}$
    for constants $c_F > 0, C_F \geq 0$. 
\end{assumption}

The local regularity condition for M-estimators in Euclidean space typically requires the
positive-definiteness of the Hessian of the objective function at the minimizer \citep{vdv}.
In our setting, the Fr\'echet objective $F(x,\cdot)$ is defined on the Bures--Wasserstein manifold
$\cS^d_{++}$. Since $\cS^d_{++}$ is an open subset of the normed linear space $(\cS^d,\Fnorm{\cdot})$,
we formulate the analogue of a Hessian using second-order Fr\'echet differentials; see
Section~\ref{sec:ot_diff} for the definitions and for the explicit expressions of $dF(x,S)$ and
$d^2F(x,S)$ in terms of the optimal transport map $T_S^Q$ and its differential $dT_S^Q$.

Throughout, we identify linear functionals on $\cS^d$ with matrices in $\cS^d$ via the Frobenius inner product by the Riesz representation theorem.
With this identification, the second differential $d^2F(x,S):\cS^d\to\cS^d$ is a symmetric linear
operator (Section~\ref{sec:ot_diff}) and plays the role of a curvature (Hessian) operator for the
Fr\'echet objective.
Assumption~\ref{assumption: minimizer_local} imposes a positive-definiteness condition on
this operator at the minimizer $Q^*(x)$.

\begin{assumption}
    \label{assumption: minimizer_local}
    For any $x \in \supp(X)$, consider the symmetric linear operator
    \begin{align*}
        \EE_{(X,Q)} \rbr{-w(x,X)dT_{Q^*(x)}^{Q}}: \cS^d \to \cS^d,
    \end{align*}
    which is the second differential of $F(x,\cdot)$ at $Q^*(x)$. This operator has a lower bound for its minimum eigenvalue given by
    \begin{align*}
        \lambda_{\min}\rbr{-\EE w(x,X)dT_{Q^*(x)}^{Q}} &\geq \frac{1}{\gamma_{\lambda}(\norm{x-\mu})}\ ,
    \end{align*}
    where $\gamma_{\lambda}(t) := c_{\lambda} \rbr{t \vee 1}^{C_{\lambda}}$ 
    for constants $c_{\lambda} > 0$ and $C_{\lambda} \geq 0$.
\end{assumption}

We close this section with two examples illustrating the assumptions above. In the commutative
design (Example~\ref{example_c}), the response matrices share a common eigenbasis; in the non-commutative design (Example~\ref{example_nc}), the
eigenbasis varies across observations. Proposition~\ref{prop: example} verifies
Assumptions~\ref{assumption: X}--\ref{assumption: minimizer_local} for the commutative design; both are used in the simulations of Section~\ref{sec: simulation}.

\begin{example}[commutative case]
    \label{example_c}

    Let $X_i=(X_{i1}\ \cdots\ X_{ip})$, $i\in[n]$, be i.i.d.\ covariates in $\RR^p$ with
    $X_i\sim\mathrm{Uniform}[-1,1]^p$, and take $p_y=p_z=2$ so that $p=4$. The response matrices
    $Q_1,\dots,Q_n\in\RR^{d\times d}$ are generated as
    \begin{align*}
        Q_i = U\,V_i\,f(X_i;\delta_z)^2\,V_i\,U^\top,
    \end{align*}
    where $U\in \cO_d$ follows the Haar measure, $V_i$ is diagonal with i.i.d.\ entries
    $V_{i,kk}\sim\mathrm{Uniform}[0.9,1.1]$, and $U$ and $\{X_i,V_i\}_{i\in[n]}$ are mutually
    independent. The map $f(\cdot;\delta_z):\RR^p\to\RR^{d\times d}$ takes values in the diagonal
    matrices, with
    \begin{align*}
        f(x;\delta_z)_{kk} = 1.5 + \frac{k}{2} + 0.1\sum_{j=1}^{p_y} y_j
            + \delta_z\sum_{l=1}^{p_z} z_l, \qquad x=(y,z),
    \end{align*}
    where $\delta_z\in[-1/4,1/4]$ controls the deviation from the null hypothesis~\eqref{eqn: null}, with
    $\delta_z=0$ corresponding to the null. We verify in
    Proposition~\ref{prop: example} that $(X,Q)$ satisfies the Fr\'echet regression model with $Q^*(X_i)=U f(X_i;\delta_z)^2 U^\top$
    and Assumptions~\ref{assumption: X}--\ref{assumption: minimizer_local}.
\end{example}

\begin{example}[non-commutative case]
    \label{example_nc}

    Let $X=(X_1\ \cdots\ X_p)\sim\mathrm{Uniform}[-1,1]^p$, with $p_y=p_z=2$. The response
    matrix $Q\in\RR^{d\times d}$ ($d$ even) is generated as
    \begin{align*}
        Q = U\,V\,g(X;\delta_z)^2\,V\,U^\top,
    \end{align*}
    where $U,V,X$ are mutually independent. The random orthogonal matrix $U\in\cO_d$ is block
    diagonal, $U=\mathrm{diag}\big(U^{(1)},\dots,U^{(\lfloor d/2\rfloor)}\big)$, with the
    $U^{(m)}$ i.i.d.\ $2\times2$ Haar-distributed orthogonal blocks; $V$ is diagonal with i.i.d.\
    entries $V_{kk}\sim\mathrm{Uniform}[0.9,1.1]$; and $g(\cdot;\delta_z):\RR^p\to\RR^{d\times d}$
    takes values in the diagonal matrices, with
    \begin{align*}
        g(x;\delta_z)_{kk} = 1.5 + 0.5\left\lceil k/2\right\rceil + 0.1\sum_{j=1}^{p_y} y_j
            + \delta_z\sum_{l=1}^{p_z} z_l, \qquad x=(y,z),
    \end{align*}
    where $\delta_z\in[-1/4,1/4]$ controls the deviation from the null hypothesis~\eqref{eqn: null},
    with $\delta_z=0$ corresponding to the null. By construction, 
    $g(x;\delta_z)^2$ is a stationary point of the Fr\'echet objective $F(x,\cdot)$.
\end{example}

\begin{proposition}
    \label{prop: example}
    Under the commutative design of Example~\ref{example_c} (with $p_y=p_z=2$ and
    $\delta_z\in[-1/4,1/4]$), the pair $(X,Q)$ satisfies the Fr\'echet regression model with
    conditional Fr\'echet mean $Q^*(x)=U f(x;\delta_z)^2 U^\top$, together with
    Assumptions~\ref{assumption: X}--\ref{assumption: minimizer_local}.
\end{proposition}


For the non-commutative design of Example~\ref{example_nc}, a direct calculation, as in the proof
of Proposition~\ref{prop: example}, verifies Assumptions~\ref{assumption: X}--\ref{assumption: bdd_Q} and shows that $g(x;\delta_z)^2$ is a stationary point of
$F(x,\cdot)$. Establishing Assumptions~\ref{assumption: minimizer_global}--\ref{assumption:
minimizer_local} is substantially harder: the Fr\'echet weight $w(x,X)$ takes both signs, and the
commutative reduction of Example~\ref{example_c} no longer applies. We thus leave a full
verification of the non-commutative case to future work and use Example~\ref{example_nc} as a
non-commutative setting in the simulations of Section~\ref{sec: simulation}.

%% file: test.tex
\subsection{Test statistic}
\label{sec: test_statistic}
We begin by recalling the global test statistic proposed in \citet{XuLi2025}:
\begin{align}
    \cT_{\text{global}} = \sum_{i=1}^n \Fnorm{\hat{H} \rbr{\hQ_{n}(X_i) - \hQ_{n}(\hmu)}}^2, \qquad \hat{H} =-\frac{1}{n} \sum_{i=1}^n dT_{\hQ_{n}(\bar{X})}^{Q_i},
    \label{eqn: xuli25_test}
\end{align}
where $\hat{H}: \cS^d \to \cS^d$ is a linear operator on the space of symmetric $d\times d$ matrices
(the differential $dT_S^Q$ is defined in Section~\ref{sec:ot_diff}). 
By \eqref{eqn: d2_mu}, $\hat H$ coincides with the empirical second differential
$d^2F_n(x,S)$ evaluated at the reference point $(x,S)=(\hmu,\hQ_n(\hmu))$.
Consequently, for each~$i$, 
$\hat{H}\rbr{\hQ_n(X_i)-\hQ_n(\hmu)} \in \cS^d$, and its Frobenius norm is squared in the definition of $\cT_{\text{global}}$.

Intuitively, $\cT_{\text{global}}$ aggregates the deviations of the fitted responses $\hQ_{n}(X_i)$ from the estimated Fr\'echet mean $\hQ_{n}(\hmu)$, with $\hat{H}$ acting as a curvature adjustment to account for the local geometry of the Wasserstein manifold.
Under the null hypothesis of no covariate effect, these deviations are expected to be small, yielding a statistic of order $O_p(1)$.

We now adapt this idea to partial-effect testing.
For each observation $X_i=(Y_i,Z_i)$, let $\hX_i$ denote the associated reference covariate
defined in~\eqref{eqn: hx} of Property~\ref{prpt: sub_model}, obtained by replacing $Z_i$
with its linear predictor given $Y_i$, at which the corresponding reference fit is evaluated.
We then define the test statistic
\begin{equation}
    \begin{aligned}
        \cT_n &= \sum_{i=1}^{n} \Fnorm{ \hat{H}_{n}(\hX_i) \rbr{\hQ_n(X_i) - \hQ_n(\hX_i) }}^2,\\
        &\text{where} \;\ \hat{H}_n(\hX_i) = -\frac{1}{n} \sum_{j=1}^{n} w_n(\hX_i,X_j) dT^{Q_j}_{\hQ_n(\hX_i)}.
    \end{aligned}
    \label{eqn: test_statistic}
\end{equation}
For any $x\in\RR^p$, the operator $\hat{H}_n(x):\cS^d\to\cS^d$ is linear.
By construction and \eqref{eqn: d2_mu}, $\hat{H}_n(\hX_i)$ coincides with the empirical second differential
$d^2F_n(\hX_i,\hQ_n(\hX_i))$ of the Fr\'echet objective evaluated at the reference point $(\hX_i,\hQ_n(\hX_i))$.
Consequently,
\begin{align*}
    \hat{H}_{n}(\hX_i) \rbr{\hQ_n(X_i) - \hQ_n(\hX_i) } \in \cS^d,
\end{align*}
whose squared Frobenius norm enters the definition of $\cT_n$.

Intuitively, $\cT_n$ compares the fitted response at the observed covariate $X_i$
with the corresponding reference fit evaluated at $\hX_i$, thereby isolating the
effect of $Z$ after adjusting for $Y$.
By Property~\ref{prpt: sub_model}, evaluating $\hQ_n(\hX_i)$ is algebraically equivalent to fitting a reduced Fr\'echet regression model that depends only on $Y_i$.
To account for the non-isotropic geometry of the Bures--Wasserstein manifold,
this difference is scaled by the local curvature operator $\hat{H}_n(\hX_i)$.

Under the null hypothesis~\eqref{eqn: null}, the uniform convergence of $\hQ_n(\cdot)$
established in Lemma~\ref{lem: xu25} implies that
$\hQ_n(X_i)$ and $\hQ_n(\hX_i)$ are uniformly close in $i$.
As a result, the test statistic $\cT_n$ is of order $O_p(1)$ under the null.
Moreover, $\cT_n$ naturally extends the global test statistic:
when no covariate adjustment is performed—specifically, when $Z=X$—
the statistic $\cT_n$ reduces to $\cT_{\mathrm{global}}$.

To justify the use of $\hX_i$ as the reference covariate in~\eqref{eqn: test_statistic},
we show in Theorem~\ref{thm: est} that evaluating the Fr\'echet regression estimator at
$\hX$ is asymptotically equivalent to evaluating the population regression function at
the corresponding population reference covariate $\tx$.
Together with the identification in Property~\ref{prpt: sub_model}, this implies that
$\hQ_n(\hX_i)$ is asymptotically equivalent to the fitted value from the reduced
Fr\'echet regression of $Q$ on $Y$ alone.
Consequently, $\hQ_n(\hX_i)$ serves as a natural reference fit for isolating the
partial effect of $Z$ after adjusting for $Y$, both under the null and under alternatives.

\begin{theorem} \label{thm: est}
    Let $0 \leq L_n \leq C_L (\log n)^{1/2}$ for some constant $C_L>0$.
    Suppose Assumption~\ref{assumption: X}-\ref{assumption: minimizer_local} hold. Then for any fixed $\tau > 0$, with probability at least $1-C_3 n^{-\tau}$,
    \begin{align*}
        \sup_{\norm{x -\mu}\leq L_n} \Fnorm{\hQ_n(\hx) - Q^*(\tx)} \leq  C_1 \frac{ \log^{C_2} {n}}{\sqrt{n}}
    \end{align*}
    for some constants $C_1, C_2, C_3 > 0$ independent of $n$.
    Here $x=(y,z)$; $\tx=(x,\tz)$ and $\hx = (y,\hz)$ are defined in \eqref{eqn: tx} and \eqref{eqn: hx}, respectively.
\end{theorem}

\begin{remark}
    Theorem~\ref{thm: est} is nontrivial in the following two aspects.

    \begin{enumerate}[leftmargin=2.5em, label=(\roman*)]
        \item Theorem~\ref{thm: est} is not a direct consequence of uniform convergence result in  Lemma~\ref{lem: xu25}. Indeed, 
        uniform convergence of $\hQ_n(x)$ implies that, with probability at least $1- O(n^{-\tau})$,
        \begin{align*}
            \sup_{\norm{x -\mu}\leq L_n} \Fnorm{\hQ_n(\hx) - Q^*(\hx)} \lesssim \frac{\polylog{n}}{\sqrt{n}}.
        \end{align*}
        Although $\hx \xrightarrow{p} \tx$, controlling the difference $\Fnorm{Q^*(\hx)-Q^*(\tx)}$ would require additional smoothness assumptions on the map $x \mapsto Q^*(x)$, which are not imposed here.

        \item It may appear tempting to argue that $\tQ_n(y)\xrightarrow{p}\tQ^*(y)$ uniformly by noting that $\tQ_n(y)$ is an M-estimator of $\tQ^*(y)$ and mimicking the proof of
        Lemma~\ref{lem: xu25} in \citet{XuLi2025}. However, such an argument would require
        imposing additional analogues of Assumptions~\ref{assumption: X}--\ref{assumption: minimizer_local}
        formulated directly for the reduced covariate $Y$ alone, which is undesirable.
    \end{enumerate}
\end{remark}

\begin{remark}
    we compare $\cT_n$~\eqref{eqn: test_statistic} with its sample-splitting counterpart $\cT_{n,\text{split}}$ from \citet{thesis}. 
    \begin{itemize}[leftmargin=2em]
        \item \textit{Construction.}
        $\cT_{n, \text{split}}$ is defined as follows. The index set $[n]=\{1,\ldots,n\}$ is partitioned uniformly at random, independently of the data, into two disjoint subsets $\cA$ and $\cB$ with $\cA \cup \cB = [n]$, $\cA \cap \cB = \emptyset$, $|\cA| = \lfloor n/2 \rfloor$ and $|\cB| = \lceil n/2 \rceil$.
        Let $\hmu_{\cA} = |\cA|^{-1} \sum_{i \in \cA} X_i$ and $\hSigma_{\cA} = |\cA|^{-1} \sum_{i \in \cA} (X_i - \hmu_{\cA})(X_i - \hmu_{\cA})^\top$ denote the sample mean and covariance over $\cA$, and let $\hQ_{\cA}(\cdot)$ be the Fr\'echet regression estimator~\eqref{eqn: estimator} fitted on $\cA$,
        \begin{align*}
            \hQ_{\cA}(x) := \argmin_{S \in \cS^d_{++}} F_{\cA}(x,S), \qquad F_{\cA}(x,S) := \frac{1}{|\cA|} \sum_{i \in \cA} w_{\cA}(x,X_i) W^2(S,Q_i),
        \end{align*}
        where $w_{\cA}(x,X_i) := 1 + (x-\hmu_{\cA})^{\top} \hSigma_{\cA}^{-1}(X_i-\hmu_{\cA})$.
        The statistic $\cT_{n, \text{split}}$ then evaluates the deviation from the null on the held-out set $\cB$,
        \begin{align*}
            \cT_{n, \text{split}} &= \sum_{i \in \cB}^{} \Fnorm{ \hat{H}_{\cA}(\hX_i) \rbr{\hQ_{\cA}(X_i) - \hQ_{\cA}(\hX_i) }}^2,\\
            &\text{where} \;\ \hat{H}_{\cA}(\hX_i) = -\frac{1}{\abs{\cA}} \sum_{j \in \cA}^{} w_{\cA}(\hX_i,X_j) dT^{Q_j}_{\hQ_{\cA}(\hX_i)} \ .
        \end{align*}

        \item \textit{Motivation and efficiency.} The two statistics $\cT_n$ and $\cT_{n, \text{split}}$ share the same asymptotic null distribution. Splitting makes the estimator $\hQ_{\cA}(\cdot)$ independent of the evaluation points $\{X_i\}_{i \in \cB}$, which simplifies the derivation of the null distribution in \citet{thesis}. The price is efficiency: 
        intuitively, $\cT_{n, \text{split}}$ only uses half of the data for testing and its power is correspondingly lower (Figure~\ref{fig: split}).
        By contrast, $\cT_n$ uses the full sample for both estimation
        and evaluation. This reuse introduces dependence between the fitted responses and the evaluation points that complicates the analysis; we resolve it by reducing $\cT_n$ to a third-order V-statistic and applying the Hoeffding decomposition to decouple these dependencies.
    \end{itemize}
\end{remark}

\subsection{Asymptotic null distribution}

For any $x \in \RR^p$, define $\vx=(1,x^\top)^\top \in \RR^{p+1}$, corresponding to the
augmented covariate vector with an intercept. Let $\Xi=\EE \vX \vX^\top \in \RR^{(p+1) \times (p+1)}$. We now establish the asymptotic null distribution of our test statistic.
\begin{theorem} \label{thm: null}
    Suppose Assumption \ref{assumption: X}-\ref{assumption: minimizer_local} hold. Under the null hypothesis \eqref{eqn: null}, the test statistic $\cT_n$ converges in distribution to
    \begin{align}
        \cT_n \xrightarrow{d} \sum_{i=1}^{\infty} \lambda_i w_i,
        \label{eqn: null_dist}
    \end{align}
    where $\{w_i\}_{i \geq 1}$ are independent $\chi_1^2$ random variables, and $\cbr{\lambda_i}_{i \geq 1}$ are the eigenvalues of the integral operator with kernel 
    $\cK: \rbr{\RR^p \times \cS^d_{++}} \times \rbr{\RR^p \times \cS^d_{++}} \to \RR$ defined as
    \begin{align}
        \cK(x,S;x',S') := \EE_{X} \inner{\tau(x,S;X)}{\tau(x',S';X)}_{\mathrm{F}}. \label{eqn: cK}
    \end{align}
    Here $\inner{\cdot}{\cdot}_{\mathrm{F}}$ is the Frobenius inner product, and $\tau: \RR^p \times \bbS^d_{++} \times \RR^p \to \RR^{d \times d}$ is the influence function defined by
    \begin{align*}
        \tau(x,S;X) &= \tau_0(x,S;X) - \tau_1(x,S;X) + A(X) \tau_0(x,S;X),
    \end{align*}
    with components
    \begin{align*}
        \tau_0(x,S;X) &= \vX^\top \Xi^{-1} \vx \rbr[\big]{T^{S}_{Q^*(X)} - I_d} - \EE'' \sbr{\vX^\top \Xi^{-1}\rbr{\vx \vx^\top - \Xi} \Xi^{-1}\vec{X}'' \rbr[\big]{T^{Q''}_{Q^*(X)} - I_d}},\\
        \tau_1(x,S;X) &= \vY^\top \Xi_{YY}^{-1} \vy \rbr[\big]{T^{S}_{Q^*(X)} - I_d} - \EE'' \sbr{\vY^\top \Xi_{YY}^{-1}\rbr{\vy \vy^\top - \Xi_{YY}} \Xi_{YY}^{-1} \vec{Y}'' \rbr[\big]{T^{Q''}_{Q^*(X)} - I_d}},\\
        A(X) &= \sbr{\EE'' \rbr[\big]{\vX^\top \Xi^{-1} \vX'' - \vY^\top \Xi_{YY}^{-1} \vY''} dT^{Q''}_{Q^*(X)}} \sbr{ -\EE'' \vX^\top \Xi^{-1} \vX'' dT^{Q''}_{Q^*(X)} }^{-1},
    \end{align*}
    where $\tX := \rbr{Y, \mu_Z + \Sigma_{ZY} \Sigma_{YY}^{-1} (Y-\mu_Y)}$ is defined as in \eqref{eqn: tx}, and $\EE''$ denotes expectation with respect to an independent copy $(X'',Q'')$ of $(X,Q)$.
\end{theorem}

\begin{remark} \label{remark: integral_operator}
    $\{\lambda_i\}_{i \geq 1}$ in Theorem~\ref{thm: null} are the eigenvalues of 
    the self-adjoint integral operator $\mathscr{T}_{\cK}: L_2(\RR^p \times \cS^d_{++}; \PP) \to L_2(\RR^p \times \cS^d_{++}; \PP)$ associated with the kernel $\cK$, defined by
    \begin{align*}
        \mathscr{T}_{\cK} f := \int \cK(\cdot, \cdot; x', Q') f(x', Q') \, d\PP(x', Q'),
    \end{align*}
    where $L_2(\RR^p \times \cS^d_{++}; \PP)$ denote the space of square-integrable functions on $\RR^p \times \cS^d_{++}$ with respect to the probability measure $\PP$ (see Appendix~\ref{sec: tech_lemmas} for RKHS background). Under the assumptions of Theorem~\ref{thm: null}, we have $\tr \mathscr{T}_{\cK} = \EE_{(X,Q)\otimes X'} \Fnorm{\tau(X,Q;X')}^2 < \infty$, 
    where $X'$ is an independent copy of $X$.  Hence the induced integral operator $\mathscr{T}_{\cK}$ is a trace-class operator on $L_2(\RR^p \times \cS^d_{++}; \PP)$. Moreover, since $\EE \sum_{i=1}^{\infty} \lambda_i w_i =  \tr \mathscr{T}_{\cK} < +\infty$, the series $\sum_{i=1}^{\infty} \lambda_i w_i$ is a well-defined non-negative random variable with finite expectation.
\end{remark}

Since $\cT_n$ generalizes the global test statistic \eqref{eqn: xuli25_test}, we show in Corrolary \ref{cor: glb_null} that Theorem~\ref{thm: null} recovers the corresponding asymptotic null distribution in \citet{XuLi2025}.
\begin{corollary}
    \label{cor: glb_null}
    Consider testing the global effect \eqref{eqn: global_null} where $p_1=p$ and $x=z$. Suppose Assumption \ref{assumption: X}-\ref{assumption: minimizer_local} hold, and that $X$ and $Q$ are independent conditionally on $Q^*(X)$. Then, under the global null hypothesis \eqref{eqn: global_null},  the eigenvalues $\lambda_1 \geq \lambda_2 \geq \ldots 0$ satisfy the following properties:
    \begin{itemize}[leftmargin=2em]
        \item $\lambda_{pd^2+1}=0$, that is, the operator has rank at most $p d^2$.
        \item The eigenvalues
        \begin{align*}
            \cbr{\lambda_{p k + 1} : k = 0,1,\ldots,d^2-1}
        \end{align*}
        coincide exactly with the eigenvalues of the $d^2 \times d^2$ covariance matrix
        \begin{align*}
            \EE \vecc \rbr{T_{Q^*(\mu)}^Q - I_d} \vecc \rbr{T_{Q^*(\mu)}^Q - I_d}^\top,
        \end{align*}
        where $\vecc$ denotes the column-wise vectorization operator.
        \item For each $k = 0,1,\ldots,d^2-1$,
        \begin{align*}
            \lambda_{p k + 1} = \lambda_{p k + 2} = \cdots = \lambda_{p k + p},
        \end{align*}
        so that each nonzero eigenvalue has multiplicity $p$.
    \end{itemize}
\end{corollary}

To estimate the eigenvalues $\{\lambda_j\}_{j \geq 1}$,  we employ a kernel matrix approach. For notational clarity, define the empirical covariance matrices
\begin{align*}
    \hXi:= n^{-1} \sum_{i=1}^{n} \vX_i \vX_i^\top, \qquad \hXi_{YY}:=n^{-1} \sum_{i=1}^{n} \vY_i \vY_i^\top.
\end{align*}
The estimated eigenvalues $\{\hlambda_j\}_{j = 1}^{n}$ are obtained as the eigenvalues of the $n \times n$ kernel matrix
\begin{align*}
    \hbK_{n} := \frac{1}{n} \sbr{\hat{\cK}_n (X_i,Q_i;X_j,Q_j)}_{i,j \in [n]} \in \RR^{n \times n},
\end{align*}
where the kernel function $\hat{\cK}_n: (\RR^p \times \cS^d_{++}) \times  (\RR^p \times \cS^d_{++}) \to \RR$ is given by
\begin{align*}
    \hat{\cK}_{n}(x,S;x',S') = \frac{1}{n} \sum_{k=1}^{n} \inner{\htau(x,S;X_k)}{\htau(x',S';X_k)}_{\mathrm{F}}.
\end{align*}
The function $\htau: \RR^p \times \cS^d_{++} \times \RR^p \to \RR^{d \times d}$ here captures the influence of each observation and is defined as a plug-in estimator of~$\tau$:
\begin{align*}
    \htau(x,S;X) &= \htau_0(x,S;X) - \htau_1(x,S;X) + \hA(X) \htau_0(x,S;X),\\
    \htau_0(x,S;X) &= \vX^\top \hXi^{-1} \vx \rbr[\big]{T^{S}_{\hQ_{n}(X)} - I_d}\\
    & - \sbr{\vX^\top \hXi^{-1} \rbr[\big]{\vx \vx^\top - \hXi} \hXi^{-1} \otimes \id}\times \sbr[\Big]{ \frac{1}{n_1} \sum_{i=1}^{n_1} \vec{X}_i \otimes \rbr[\big]{T^{Q_i}_{\hQ_{n}(X)} - I_d}},\\
    \htau_1(x,S;X) &= \vY^\top \hXi_{YY}^{-1} \vy \rbr[\big]{T^{S}_{\hQ_{n}(X)} - I_d}\\
    & - \sbr{\vY^\top \hXi_{YY}^{-1}\rbr[\big]{\vy \vy^\top - \hXi_{YY}} \hXi_{YY}^{-1} \otimes \id}\times \sbr[\Big]{ \frac{1}{n} \sum_{i=1}^{n} \vec{Y}_i \otimes \rbr[\big]{T^{Q_i}_{\hQ_{n}(X)} - I_d}},\\
    \hA(X) &= \sbr[\Big]{\frac{1}{n}  \sum_{i=1}^{n} \rbr[\big]{\vX^\top \hXi^{-1}\vX_i - \vec{\hX}^\top \hXi^{-1}\vX_i } dT^{Q_i}_{\hQ_{n}(X)}} \sbr[\Big]{ -\frac{1}{n} \sum_{i=1}^{n}  w_{n}(X,X_i)dT^{Q_i}_{\hQ_{n}(X)} }^{-1},
\end{align*}
where $\hX := (Y, \hmu_Z + \hSigma_{ZY} \hSigma_{YY}^{-1} (Y-\hmu_Y))$.

Given the eigenvalue estimates~$\{\hlambda_j\}_{j=1}^{n}$, let $\hq_{1-\alpha}$ denote the $(1-\alpha)$-quantile of
\begin{align}
    \sum_{j = 1}^{n} \hlambda_j w_j,
    \label{eqn: est_null}
\end{align}
where $\{w_j\}_{j=1}^n$ are independent $\chi_1^2$ random variables.
We now define our level-$\alpha$ test $\Phi_{n,\alpha}$ by
\begin{align}
    \Phi_{n,\alpha}:= \1 \rbr{\cT_n > \hq_{1-\alpha}}.
    \label{eqn: test}
\end{align}

\begin{theorem} \label{thm: size}
    Suppose Assumptions~\ref{assumption: X}-\ref{assumption: minimizer_local} hold. Under the null hypothesis \eqref{eqn: null}, the test $\Phi_{n,\alpha}$ attains the nominal level asymptotically:
    \begin{align*}
        \lim_{n \to \infty}\PP \rbr{\cT_n > \hq_{1-\alpha}} = \alpha.
    \end{align*}
\end{theorem}



\begin{remark}
    To simplify notation in this remark, let $\cR:= \RR^p \times \cS^d_{++}$, write $r=(x,S) \in \cR$ for a generic element, and set $R_i=(X_i,Q_i)$, $i \in [n]$.
    The limiting law \eqref{eqn: null_dist} is the infinite series $\sum_{j\ge1}\lambda_j w_j$ in the
    eigenvalues of $\mathscr{T}_{\cK}: L_2(\cR; \PP) \to L_2(\cR; \PP)$ (the integral operator associated with kernel $\cK$ from Remark~\ref{remark: integral_operator}), whereas the calibration quantile $\widehat q_{1-\alpha}$ is
    computed from the finite sum \eqref{eqn: est_null} using the eigenvalues
    $\{\widehat\lambda_j\}_{j=1}^n$ of the $n\times n$ matrix $\widehat{\bK}_n$. 
    The proof of
    Theorem~\ref{thm: size} rests on spectral closeness of $\widehat{\bK}_n$ to  $\mathscr{T}_{\cK}$ 
    (in the sense that,
    $\sum_{j=1}^n\widehat\lambda_j \xrightarrow{p} \sum_{j\ge1}\lambda_j$ and
    $\sum_{j\ge1}(\widehat\lambda_j-\lambda_j)^2 \xrightarrow{p} 0$ with
    $\widehat\lambda_j:=0$ for $j>n$; see Appendix~\ref{sec: prf_size} for details). 
    We now give some intuition for this closeness, which can be understood as a consequence of the following three-step approximation
    \[
        \begin{aligned}
        &\mathscr{T}_{\cK} \;\xapprox{\ (\mathrm i)\ }\; \bK_n
                    \;\xapprox{\ (\mathrm{ii})\ }\; \bar{\bK}_n
                    \;\xapprox{\ (\mathrm{iii})\ }\; \widehat{\bK}_n,\\[2pt]
        &[\bK_n]_{ij}=\tfrac1n\mathcal K(R_i,R_j),\quad
        [\bar{\bK}_n]_{ij}=\tfrac1n\bar{\cK}_n(R_i,R_j),\quad
        [\widehat{\bK}_n]_{ij}=\tfrac1n\widehat{\mathcal K}_n(R_i,R_j),
        \end{aligned}
    \]
    where $\cK(r; r')=\EE_X \Finner{\tau(r;X)}{\tau(r';X)}$ from \eqref{eqn: cK},
    \begin{align*}
        \bar{\cK}_n(r,r')&:=n^{-1}\sum_{k=1}^n\langle\tau(r;X_k),\tau(r';X_k)\rangle_{\mathrm{F}} \quad \text{and}\quad
        \widehat{\cK}_n(r,r'):= n^{-1}\sum_{k=1}^n\langle\htau(r;X_k),\htau(r';X_k)\rangle_{\mathrm{F}}.
    \end{align*}
    \begin{enumerate}[label=(\roman*),leftmargin=*]
        \item passes via the kernel trick \citep{rkhs10}. Let $\cH_{\cK}$ be the RKHS of
        $\cK$ and $\mathscr S_n:\cH_{\cK}\to\mathbb R^n$ the sampling operator
        $(\mathscr S_nf)_i=n^{-1/2}f(R_i)$, with adjoint $\mathscr S_n^\dagger$. By the reproducing property, one has
        \[
        \mathscr T_n:=\mathscr S_n^\dagger \mathscr S_n
        =\frac1n\sum_{i=1}^n\langle\,\cdot\,,\mathcal K_{R_i}\rangle_{\cH_{\cK}}\,\mathcal K_{R_i},
        \qquad \mathcal K_{R_i}:=\mathcal K(R_i,\cdot),
        \]
        an operator on $\cH_{\cK}$ whose mean $\mathbb E\,\mathscr T_n$ shares its nonzero
        eigenvalues $\{\lambda_j\}$ with the integral operator $\mathscr T_{\mathcal K}$ of
        Theorem~\ref{thm: null} \citep[Prop.~8]{rkhs10}. Meanwhile, the nonzero eigenvalues of
        $\mathscr T_n=\mathscr S_n^\dagger\mathscr S_n$ coincide with those of
        $\mathscr S_n\mathscr S_n^\dagger=\bK_n$; since $\mathscr T_n$ concentrates around its mean $\mathbb E\,\mathscr T_n$, the spectral closeness (i) follows.

        \item is bridged through a matrix $\bar{\bK}^{\,-}_n$, with entries
        \begin{align*}
            n[\bar{\bK}^{\,-}_n]_{ij}=
            \begin{cases}
                \tfrac1{n-2}\sum_{k\ne i,j} \langle\tau(R_i;X_k),\tau(R_j;X_k)\rangle_{\mathrm{F}},& i\ne j\\
                \tfrac1{n-1}\sum_{k\ne i} \langle\tau(R_i;X_k),\tau(R_i;X_k)\rangle_{\mathrm{F}},& i=j
            \end{cases}
        \end{align*}
        i.e. leave-two-out on the off-diagonal and leave-one-out on the diagonal. Both approximations are then tractable: 
        \begin{itemize}[leftmargin=1em, label=-]
            \item $\bar{\bK}^{\,-}_n$ differs from $\bar{\bK}_n$ only in the deleted summands and scaling, a negligible perturbation;
            \item $\bar{\bK}^{\,-}_n\approx\bK_n$ follows from concentration since conditionally on $(R_i,R_j)$, $[\bar{\bK}^{\,-}_n]_{ij}$ averages i.i.d.\ terms with mean $[\bK_n]_{ij}$.
        \end{itemize}

        \item is controlled by the convergence $\htau\to \tau$.
    \end{enumerate}
\end{remark}

\subsection{Asymptotic power of the proposed test}
We now analyze the power of the test $\Phi_{n,\alpha}$ against a sequence of alternative hypotheses.
To formalize the alternatives, we first introduce the relevant hypothesis space. Let $\mathscr{P}$ denote the class of probability measures satisfying the regularity conditions imposed throughout the paper:
\begin{equation*}
    \mathscr{P}:= \cbr{\PP \in \cP_2 \rbr[\big]{\RR^p \times \cS^d_{++}}: \PP \text{ satisfies } \text{Assumption } \ref{assumption: X} - \ref{assumption: minimizer_local} }.
\end{equation*}

We consider alternatives under which the conditional independence in \eqref{eqn: null} fails by at least a prescribed amount. Specifically, define the sequence of alternative hypothesis classes
\begin{align*}
    H_{1,n}: \PP \in \mathscr{P}_{F}(a_n) :=  \cbr{\tPP \in \mathscr{P}: \EE_{(X,Q) \sim \widetilde{\PP}} \Fnorm[\big]{Q^*(X)-Q^*\rbr[\big]{(Y, \mu_Z +  \Sigma_{ZY}\Sigma_{YY}^{-1}(Y-\mu_Y) )}}^2 \geq a_n^2}
\end{align*}

The following theorem establishes uniform consistency of the proposed test against
this class of alternatives.

\begin{theorem} \label{thm: power}
    Consider a sequence of alternative hypotheses $H_{1,n}$ with $a_n \gtrsim n^{\alpha_F - 1/2}$ for some constant $\alpha_F>0$. Then the worst-case power converges uniformly to one:
    \begin{align*}
        \lim_{n \to \infty} \inf_{\PP \in \mathscr{P}_F(a_n)}\PP\rbr{\cT_n > \hq_{1-\alpha}} = 1.
    \end{align*}
\end{theorem}
The intuition underlying Theorem~\ref{thm: power} is as follows.
Under alternatives in $\mathscr{P}_F(a_n)$, the test statistic~$\cT_n$ grows on the
order of $n a_n^2$, whereas the critical value $\hq_{1-\alpha}$ increases only at a
polylogarithmic rate in $n$. As a result, the test rejects the null hypothesis with
probability tending to one.

%% file: simulation.tex
To compute the Fr\'echet estimator $\hQ_n(x)$ in \eqref{eqn: estimator}, we employ a Riemannian gradient descent algorithm summarized in Algorithm~\ref{alg:gd}. At each iteration, the algorithm evaluates the gradient $G$ of the empirical Fr\'echet objective $F_n(x,S)$ in the tangent space at the current iterate $S$. A gradient step is then taken in the tangent space, and the update
$S \gets (I_d -\eta G) S (I_d -\eta G)$
maps the iterate back to the manifold $\cS^d_{++}$ via the exponential map. The procedure is iterated until convergence, as measured by the operator norm of the gradient, or until a prescribed maximum number of iterations is reached. 

In the numerical experiments reported in Example~\ref{example_c} and~\ref{example_nc} below, we use
the initialization $S_0 = I_d$ and a constant step size $\eta = 1$, which are found to perform
well in practice; see \citet{XuLi2025} for further details.

\begin{algorithm}[!htbp]
    \caption{Gradient descent for computing $\hQ_n(x)$ in Fr\'echet regression}
    \label{alg:gd}

    \KwInput{Covariates~$\{X_i\}_{i=1}^n$, responses~$\{Q_i\}_{i=1}^n$,
    covariate~$x$, learning rate~$\eta$,
    initialization~$S_0 \in \cS^d_{++}$, maximum iterations~$T$, threshold~$\eps$.}

    $S \gets S_0$\;

    \For{$t=1,\ldots,T$}{
    $G \gets \frac{1}{n}\sum_{i=1}^n
    w_{n}(x,X_i)\bigl(I_d - T_S^{Q_i}\bigr)$ \;
    \If{$\Onorm{G} < \eps$}{
        \textbf{break}\;
    }
    $S \gets (I_d -\eta G) S (I_d -\eta G)$\;
    
    }

    \KwOutput{$S$}
\end{algorithm}

In the remainder of this section, we present numerical experiments to illustrate the finite-sample performance of the proposed test. 
We consider two representative settings that follow the Fr\'echet regression model. 
Example~\ref{example_c} examines a commutative case in which the covariance matrices share a common eigenspace, reducing the Fréchet regression to a linear regression on matrix square roots. In contrast, Example~\ref{example_nc} investigates a non-commutative setting in which the covariance matrices have distinct eigenspaces.

    
    
    

\begin{figure}[!htb]
    \centering
    \includegraphics[width=\textwidth]{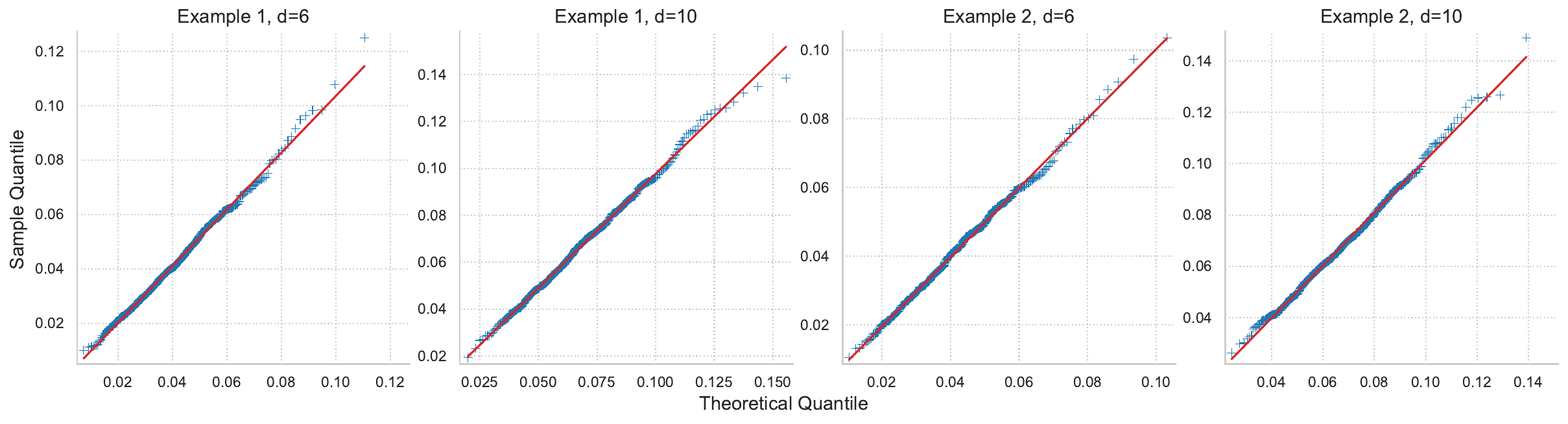}\\
    \includegraphics[width=\textwidth]{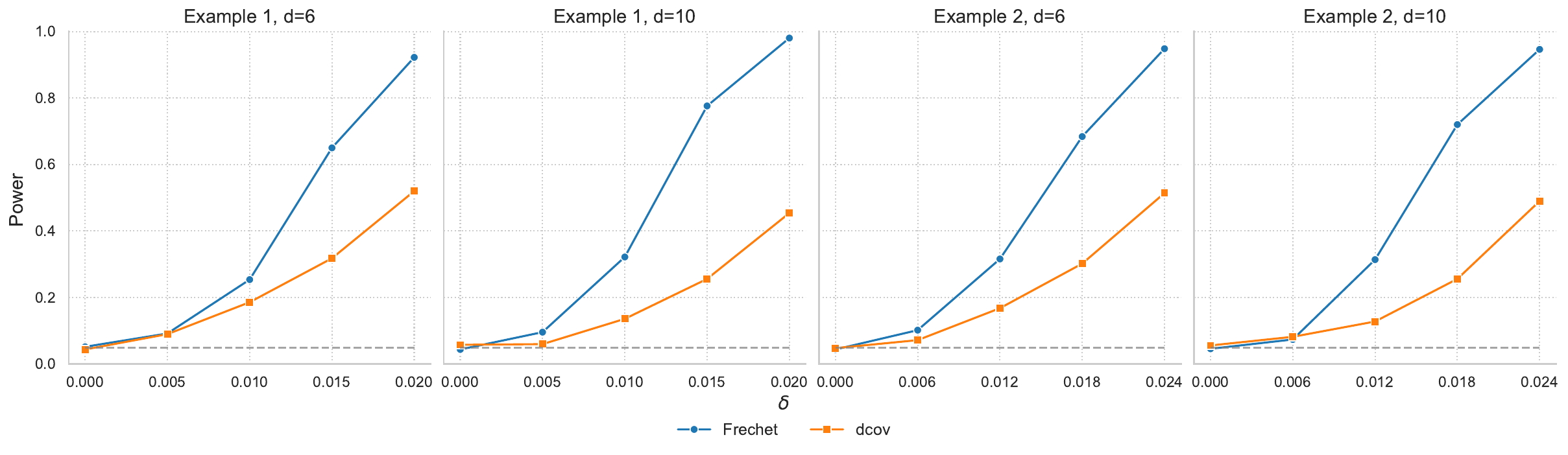}
    \caption{Top: Q-Q plots of $\cT_n$ under the null hypothesis, with sample size $n=200$ and covariate dimensions $p_y=p_z=2$. Each plot is based on 500 Monte Carlo replications. The red line indicates the least-squares linear fit between the empirical and theoretical quantiles. Bottom: Empirical power curves as functions of the effect size $\delta_z$, with sample size $n=200$ and covariate dimensions $p_y=p_z=2$. Each point is based on 500 Monte Carlo replications. The dashed line indicates the nominal level $\alpha=0.05$. }
    \label{fig: qq}
\end{figure}

\textit{Null calibration.}
We first verify that the test statistic $\cT_n$ tracks the asymptotic null distribution in Theorem~\ref{thm: null}. Fixing $\delta_z=0$ so that the null \eqref{eqn: null} holds. The top panle of Figure~\ref{fig: qq} gives Q-Q plots of $\cT_n$ against its limiting law, based on 500 Monte Carlo replications for both examples with sample size $n=200$, dimensions $d=6$ and $d=10$. The covariate dimensions are set to $p_y=p_z=2$. The Q-Q plots exhibit good agreement with the theoretical distribution, closely tracking the identity line. 

To further quantify this agreement, we perform a least-squares linear regression of the sample quantiles on the corresponding theoretical quantiles. The resulting slopes and intercepts are reported in Table~\ref{tab:qq-linear-fit}. Across all configurations, the estimated slopes are close to one and the intercepts are near zero, providing additional evidence for the accuracy of the proposed asymptotic approximation.

\begin{table}[!htb]
    \centering
    \tbl{Linear fit of Q-Q plots}
    {\begin{tabular}{lccccccccc}
        \hline
        & \multicolumn{4}{c}{Example~\ref{example_c}} & & \multicolumn{4}{c}{Example~\ref{example_nc}} \\
        \cline{2-5} \cline{7-10}
        Dimension $d$ & 6 & 10 & 20 & 30 & & 6 & 10 & 20 & 30 \\
        \hline
        $\beta_1$ (slope) 
        & 1.04 & 0.974 & 0.951 & 0.960 & & 1.01 & 1.03 & 0.970 &0.937\\
        $\beta_0 \;(\times 10^{-4})$ 
        & 6.14 & 2.83 & 40.3 & 66.9 & & 7.54 & 11.7  & 37.0 & 87.1\\
        \hline
    \end{tabular}}
    \label{tab:qq-linear-fit}
    \begin{tabnote}
        $\beta_1$ and $\beta_0$ denote the slope and intercept, respectively, of the least-squares linear fit between sample and theoretical quantiles.
    \end{tabnote}
\end{table}

\textit{Power against alternatives.} Next, we compare the empirical power of the proposed Fr\'echet test with that of the distance covariance test for partial correlations \citep{p_dcov14}, as implemented in the Python package $\texttt{dcor}$  \citep{dcor_package}. 
The bottom panel of Figure~\ref{fig: qq} reports power curves as functions of the partial-effect size $\delta_z$ under the same simulation settings as in the top panel. Each power estimate is based on 500 Monte Carlo replications. Since the proposed procedure is tailored  to the Fr\'echet regression framework, whereas the distance covariance test is fully nonparametric, we observe substantially higher power for the Fr\'echet test in this setting, as expected.

\ignore{
\begin{figure}[!htb]
    \centering
    \includegraphics[width=\textwidth]{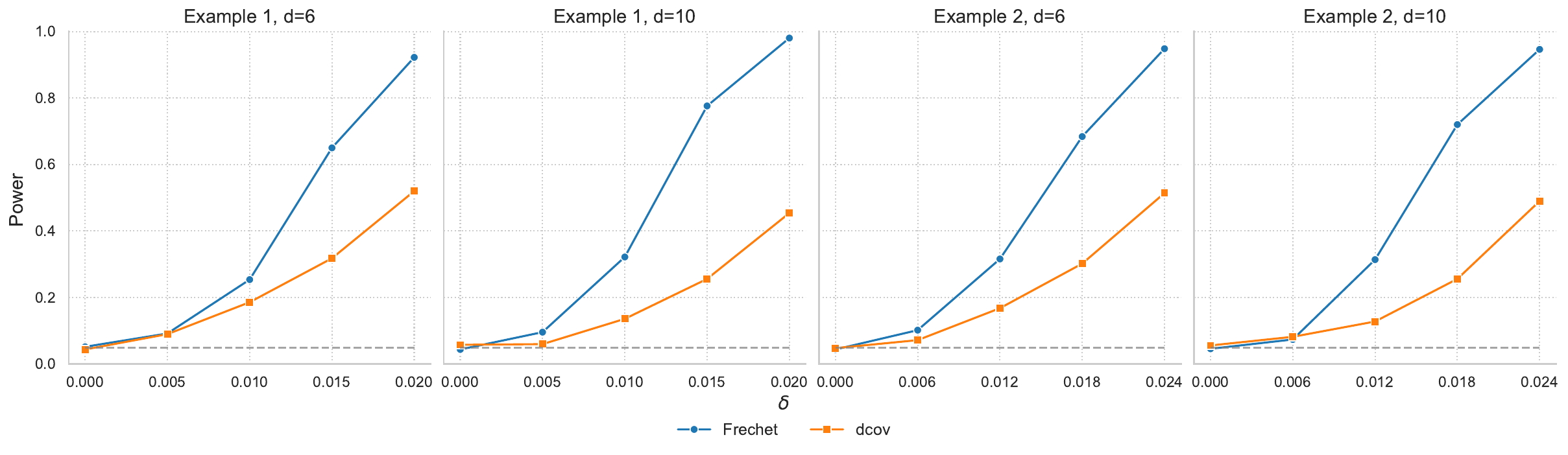}
    \caption{Empirical power curves as functions of the effect size $\delta_z$, with sample size $n=200$ and covariate dimensions $p_y=p_z=2$. Each point is based on 500 Monte Carlo replications. The dashed line indicates the nominal level $\alpha=0.05$.}
    \label{fig: power}
\end{figure}
}

\begin{figure}[!htb]
    \centering
    \includegraphics[width=.8\textwidth]{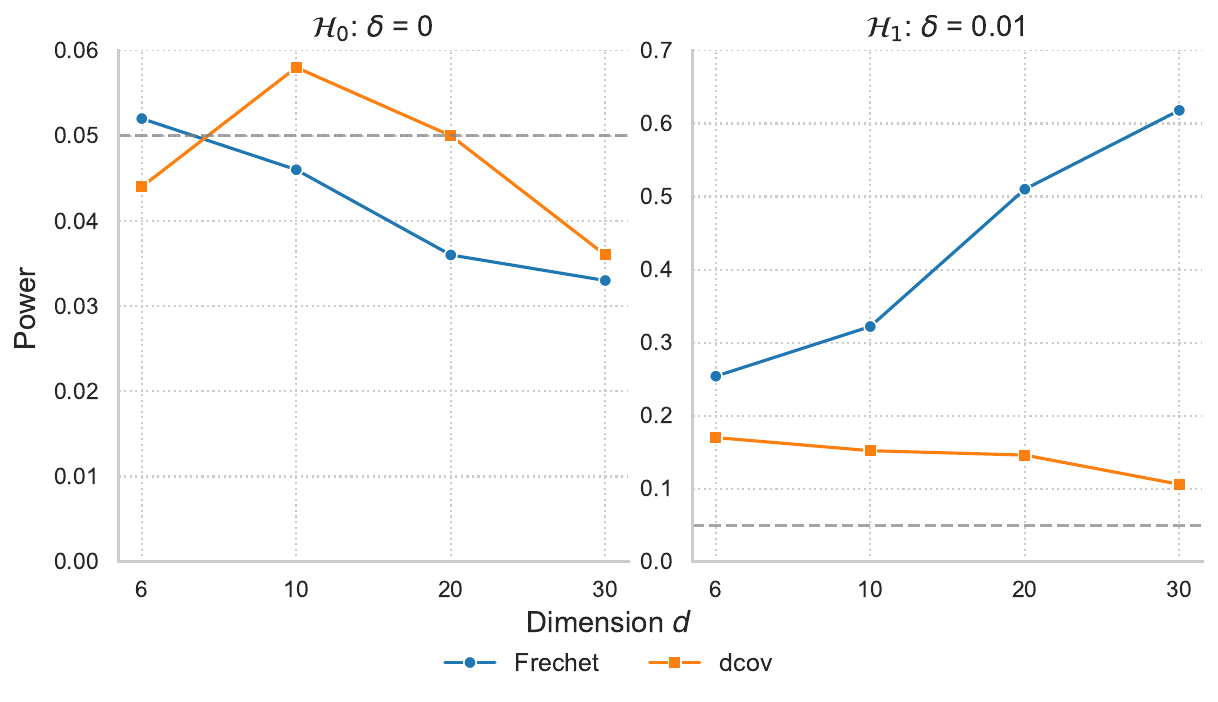}\\
     \includegraphics[width=.8\textwidth]{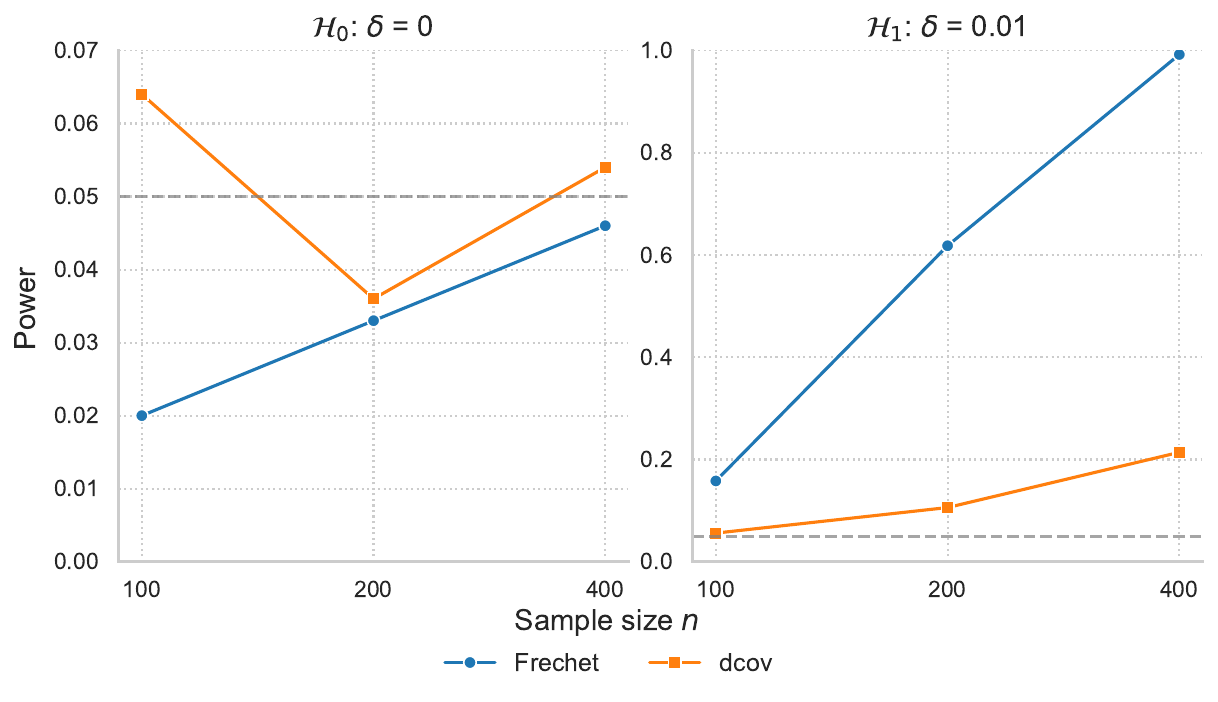}
    \caption{Top: empirical power as a function of the dimension $d$ of the covariance matrix, with sample size $n=200$ and covariate dimensions $p_y=p_z=2$. Bottom: Empirical power as a function of the sample size $n$, with covariance matrix dimension $d=30$ and covariate dimensions $p_y=p_z=2$.
    Each point is based on 500 Monte Carlo replications. The dashed line indicates the nominal level $\alpha=0.05$.}
    \label{fig: ds}
\end{figure}

\textit{Dimension $d$.} We also investigate the impact of dimensionality and sample size on test performance. The top panel of Figure~\ref{fig: ds} presents results for varying dimensions $d \in \cbr{6,10,20,30}$ with sample size fixed at $n=200$. As the dimension increases, the proposed test becomes slightly more conservative, but it continues to perform satisfactorily even at $d=30$. Moreover, for a fixed partial-effect size $\delta_z$, increasing the dimension leads to a more pronounced departure from the null hypothesis. The proposed Fr\'echet test captures this effect and attains higher power, whereas the power of the distance covariance test deteriorates with increasing dimension. Hence, while both methods exhibit sensitivity to increasing dimension, the impact appears to be more substantial for \texttt{dcov} in our simulations.

\textit{Sample size.} The bottom panel of Figure~\ref{fig: ds} further illustrates the effect of sample size, showing results for $n \in {100,200,400}$ with the dimension fixed at $d=30$. As expected, the performance of the proposed test improves with increasing sample size and becomes stable for $n \geq 200$.

\ignore{
\begin{figure}[!htb]
    \centering
    \includegraphics[width=.8\textwidth]{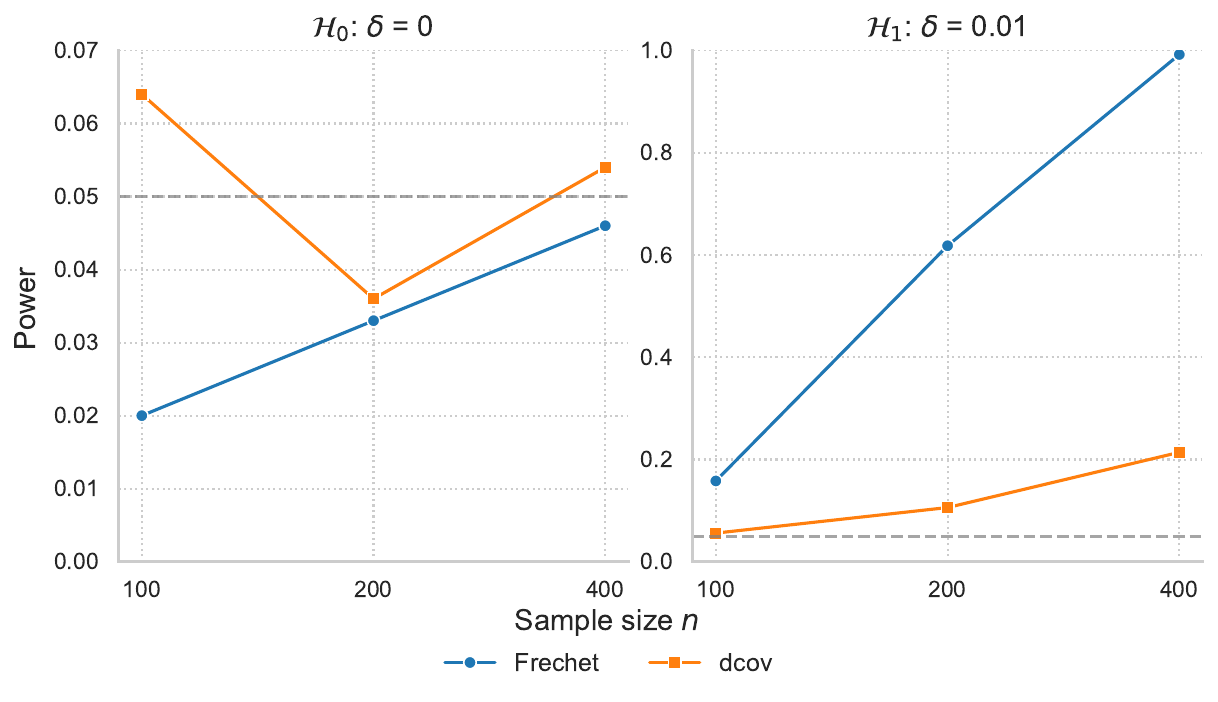}
    \caption{Empirical power as a function of the sample size $n$, with covariance matrix dimension $d=30$ and covariate dimensions $p_y=p_z=2$. Each point is based on 500 Monte Carlo replications. The dashed line indicates the nominal level $\alpha=0.05$.}
    \label{fig: ns}
\end{figure}
}

\begin{figure}[!htb]
    \centering
    \includegraphics[width=\textwidth]{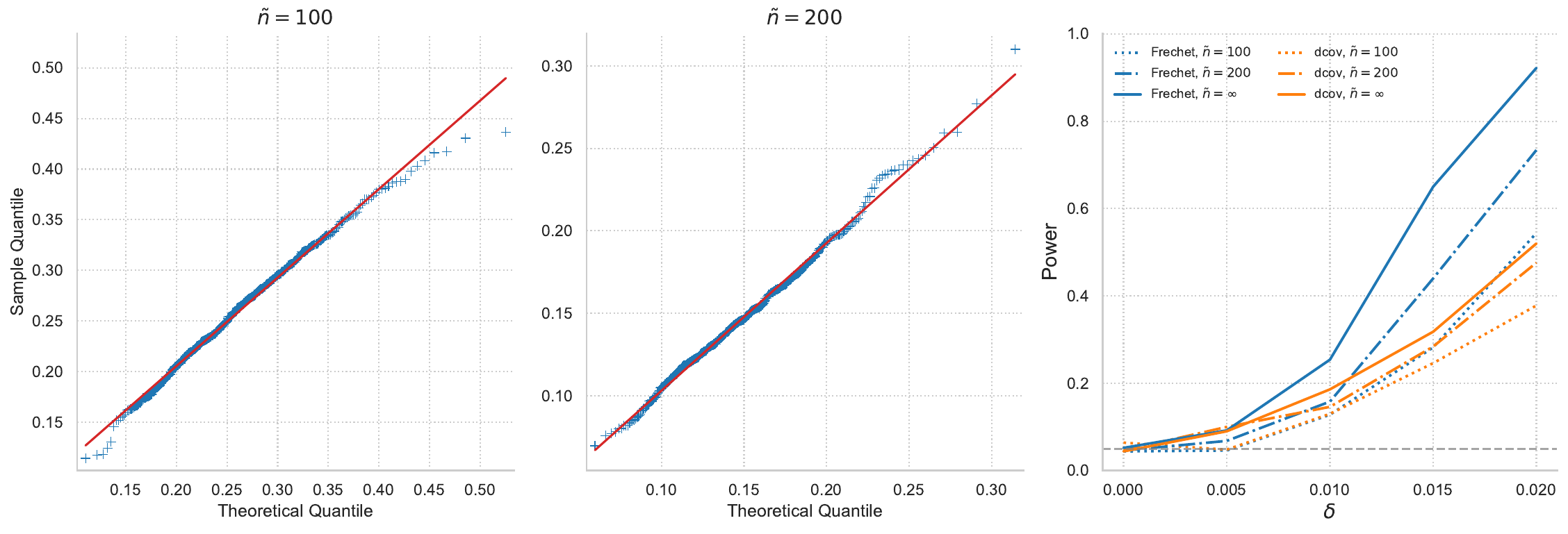}
    \caption{Performance of the proposed test when the responses are the plug-in
    covariances $\tQ_i$, under the setting of Example~\ref{example_c} with $d = 6$,
    and $n = 200$. Left and middle: Q-Q plots of $\tilde{\cT}_n$ under
    the null ($\delta_z = 0$) against the estimated null quantiles, for $\tn = 100$ and
    $\tn = 200$. Right: empirical power as a function of the effect size
    $\delta_z$ for the proposed test (Fr\'echet) and \texttt{dcov}, at
    $\tn \in \{100, 200, \infty\}$. Each point is based on 500 Monte Carlo replications.}
    \label{fig: n_tilde}
\end{figure}

\textit{Effect of estimated covariance matrices.} Finally, we investigate the size and power
of the proposed test when the response covariances are not observed but estimated.
Specifically, we consider the case where we observe
$\{(X_i; A_{i1}, A_{i2}, \ldots, A_{i \tn}) : i \in [n]\}$ with
$A_{i1}, A_{i2}, \ldots, A_{i \tn} \overset{i.i.d.}{\sim} N(0, Q_i)$, and take the plug-in
estimator $\tQ_i := \tn^{-1} \sum_{j \in [\tn]} A_{ij} A_{ij}^\top$. We then substitute
$\{(X_i, \tQ_i) : i \in [n]\}$ into the entire downstream pipeline: the
estimator~\eqref{eqn: estimator}, the test statistic~\eqref{eqn: test_statistic}, and the
eigenvalue estimates in~\eqref{eqn: est_null} that determine the calibrated quantile.
Figure~\ref{fig: n_tilde} reports, for
$\tn \in \{100, 200, \infty\}$\footnote{$\tn = \infty$ denotes the oracle case in which $Q_i$
is observed.} under the setting of Example~\ref{example_c} with $d = 6$, the Q-Q plot of the
observed test statistics against the estimated null quantiles, together with the power curves
of the proposed test (Fr\'echet) and \texttt{dcov}. The size remains close to the nominal
level across all values of $\tn$, while the power of the proposed test dominates
\texttt{dcov} throughout. The same phenomenon is observed for the global test of
\citet{XuLi2025}. We conjecture that, under the null, the test statistic computed from
$\{(X_i, \tQ_i)\}$ still converges to a weighted sum of independent $\chi^2_1$ variables,
now with weights that the plug-in eigenvalues continue to estimate consistently---which
would explain the observed size control. A theoretical justification is left for future work.

%% file: singlecell.tex
Aging is a complex process of accumulation of molecular, cellular, and organ damage, leading to loss of function and increased vulnerability to disease and death. Nutrient-sensing pathways, namely insulin/insulin-like growth factor signaling and target-of-rapamycin are conserved in various organisms. We are interested in understanding the co-expression structure of 61 genes in this KEGG nutrient-sensing pathways based on the recently published population scale single cell RNA-seq data of human peripheral blood mononuclear cells (PBMCs) from blood samples of over 982 healthy individuals with ages ranging from 20 to 90 \citep{yazar2022single}.

Our analysis focuses on CD4+ naive and central memory T (CD4NC) cells, the most abundant cell type in the cohort. Age-associated changes in CD4 T-cell
functionality have been linked to chronic inflammation and decreased immunity
\citep{aging2}. Of the $61$ pathway genes, $G = 51$ are expressed in this cell
type, and our analysis is restricted to these. To keep the individual-specific
covariances well estimated, we retain the $n = 181$ donors for whom at least
$400$ CD4NC cells are observed, and write $n_i \ge 400$ for the number of such
cells of donor $i \in [n]$. Figure~\ref{fig:append_cell} shows the distribution of the number of cells across these 181 donors. For each donor we record the covariate vector
$X_i = (\mathrm{Age}_i, \mathrm{Sex}_i, \mathrm{Prop}_i)^{\top}$, where $\mathrm{Prop}_i$ is the
proportion of CD4NC cells and sex is coded as $0$ for female and $1$ for male. Figure~\ref{fig:append_age} shows the distribution of age of the 181 donors in our analysis. 

Let $E_{ic} \in \RR^{G}$ denote the expression of the~$G$ genes in
 cell $c$ of donor $i$, for $c \in [n_i]$. We first estimate donor $i$'s across-cell sample covariance
\[
  \Sigma_i \;=\; \frac{1}{n_i - 1} \sum_{c=1}^{n_i}
    \bigl(E_{ic} - \bar E_i\bigr)\bigl(E_{ic} - \bar E_i\bigr)^{\top}
    \;\in\; \cS^{G}_{+},
  \qquad
  \bar E_i \;=\; \frac{1}{n_i} \sum_{c=1}^{n_i} E_{ic} .
\]
Since our theory is formulated at the population level and requires the true individual
covariances to be strictly positive definite, we further restrict attention to $d = 20$ genes selected as follows.
For gene $g \in [G]$ and donor $i$, let $s^{2}_{ig} = (\Sigma_i)_{gg}$ be the
cell-level variance of gene $g$ in donor $i$, and
$\underline{s}^{2}_{g} = \min_{i \in [n]} s^{2}_{ig}$ its minimum across donors.
We take $\mathcal{G}_d \subseteq [G]$, with $|\mathcal{G}_d| = d$, to index the
$d$ genes with the largest $\underline{s}^{2}_{g}$. The gene set $\mathcal{G}_d$ is common to all donors; the genes are listed in decreasing order of $\underline{s}^{2}_{g}$ in Table~\ref{tab:selected_genes}.
We find the resulting
principal submatrix
$\Sigma_i^{\mathcal{G}_d} := \bigl(\Sigma_i\bigr)_{\mathcal{G}_d, \mathcal{G}_d} \in \RR^{d \times d}$
to be strictly positive definite for every $i \in [n]$, and
$\{(X_i, \Sigma_i^{\mathcal{G}_d})\}_{i \in [n]}$ is the final dataset for our
Fr\'echet regression analysis.

We apply the proposed test to assess the effects of age, sex, and CD4NC cell proportion on the gene co-expression structure. We report the tests at $d = 20$ (Table~\ref{tab: singlecell_pvalues}, first row), with $d = 10$ and $15$ included to assess robustness to the number of genes retained.
When Fr\'echet regression is performed with age as the sole covariate, the proposed test~\eqref{eqn: test} reduces to the global test of \citet{XuLi2025}, yielding strong evidence of an association between gene co-expression structure and age (p-value = 0.0053).

We further examine whether this age effect persists after adjusting for sex, CD4NC cell proportion, or both. Specifically, we test the null hypothesis $\cH_0: Q^*(y,t)=Q^*(y,t')$ for all ages $t,t'$, where $y$ denotes sex, CD4NC cell proportion, or their concatenation. In every case, the association with age remains statistically significant after adjustment. In contrast, when testing for the effects of sex while adjusting for age and cell proportion, or for the effects of cell proportion while adjusting for age and sex, no significant associations are detected. These conclusions hold across $d \in \{10, 15, 20\}$; the results at $d = 10, 15$
confirm robustness to the number of genes retained.

\begin{table}[!htb]
    \centering
    \tbl{p-values of global and conditional tests}
    {
        \begin{tabular}{lcccccc}
            \hline
            \multicolumn{1}{c}{} &\multicolumn{6}{c}{Test}\\
            $d$ & $A$
            & $A \mid P$
            & $P \mid A$
            & $A \mid P,S$
            & $S \mid P,A$
            & $P \mid A,S$ \\
            \hline
            20
            & 0.0053 & 0.0056 & 1.0 & 0.0046 & 0.13 & 1.0 \\
            \hline
            15
            & 0.0013 & 0.0013 & 1.0 & 0.0008 & 0.11 & 1.0 \\
            \hline
            10
            & 0.0004 & 0.0003 & 1.0 & 0.0002 & 0.18 & 1.0 \\
            \hline
        \end{tabular}
    }
    \label{tab: singlecell_pvalues}
    \begin{tabnote}
        \textit{Notes.}
        $d$ = number of top genes selected;
        $A$ = Age; $S$ = Sex; $P$ = CD4NC cell proportion.
        For generic variables $X$ and $Y$, $X \mid Y$ denotes testing $X$
        while adjusting for $Y$; similarly, $X \mid Y,Z$ denotes testing $X$
        while adjusting for both $Y$ and $Z$.
    \end{tabnote}
\end{table}

\begin{figure}[p]
  \makebox[\textwidth][c]{%
    \rotatebox{270}{%
      \begin{minipage}{\textheight}
        \centering
        \includegraphics[width=\textheight,height=\textwidth,keepaspectratio]{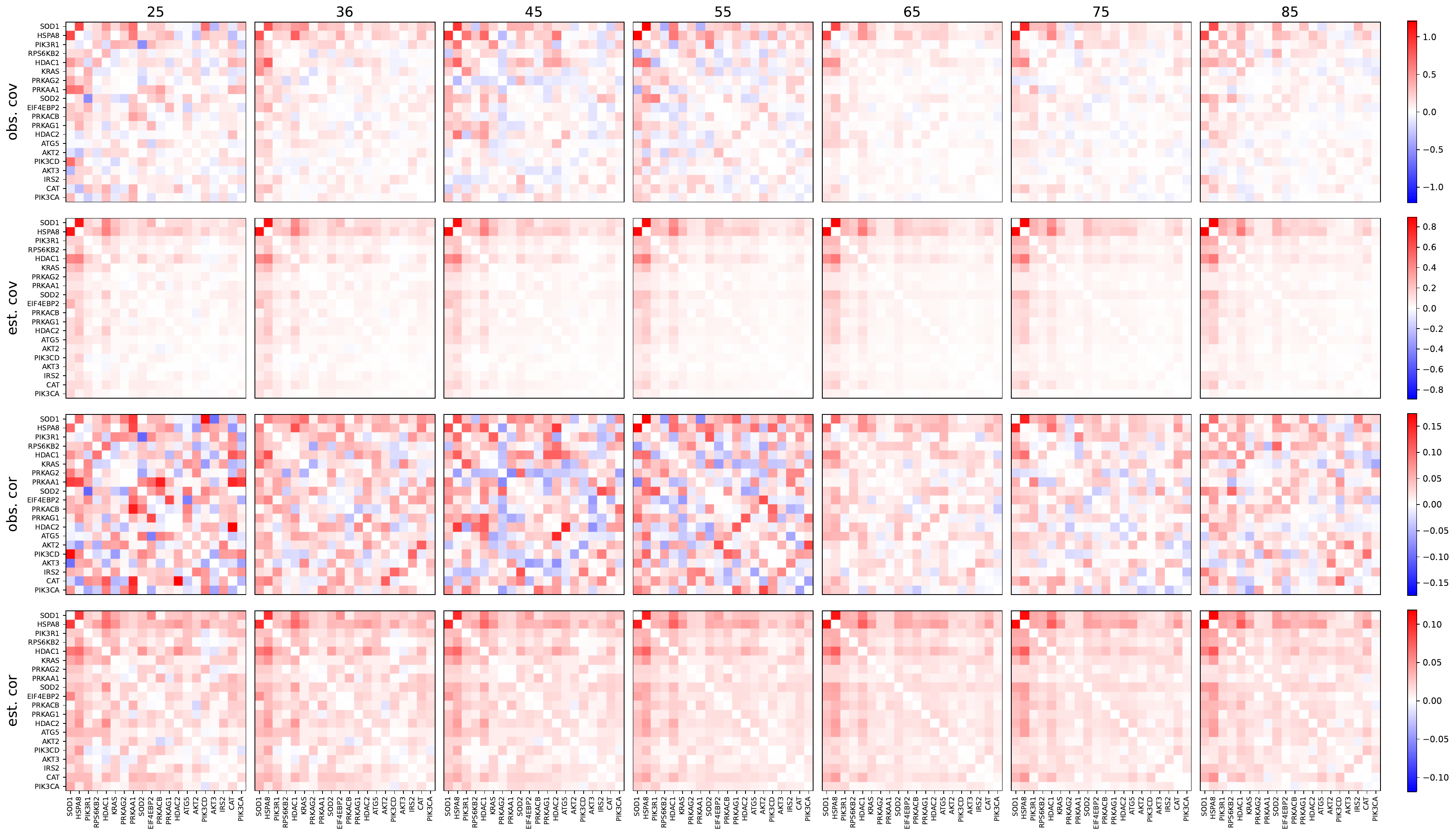}%
        \captionof{figure}{The observed responses $\hSigma^{\cG_d}(\mathrm{age})$ and the fitted covariances $\hQ_n(\mathrm{age})$ (top two rows), with the correlations obtained from each (bottom two rows), at ages $25, 36, 45, 55, 65, 75, 85$. Only off-diagonal entries are shown.}%
        \label{fig: data_mtrx}
      \end{minipage}%
    }%
  }
\end{figure}

To visualize the age dependence, Figure~\ref{fig: data_mtrx} summarizes the fit of the Fr\'echet regression
model with age as the sole covariate at ages 25, 36\footnote{Age $36$ is selected in place of $35$, for which no donor has $\ge 400$ CD4NC cells.}, 45, 55, 65, 75, 85 with $d=20$: the observed responses $\hSigma^{\cG_d}(\mathrm{age})$ and the estimated covariances $\widehat Q_n(\mathrm{age})$ in the first two rows, and their correlation counterparts in the last two. Here the observed response $\hSigma^{\cG_d}(\mathrm{age})$
is the average of the observed sample covariances over all donors of that age:
\begin{align*}
    \hSigma^{\cG_d}(\mathrm{age}) = \frac{1}{|\{i: \mathrm{Age}_i = \mathrm{age}\}|} \sum_{i: \mathrm{Age}_i = \mathrm{age}} \Sigma_i^{\mathcal{G}_d}.
\end{align*}
The fitted rows are visibly smoother than the observed rows, yet preserve the dominant co-expression block. Reading across ages reveals a gradual change in co-expression structure; this is discernible in the covariance rows and more pronounced in the correlation rows.

\begin{figure}[!htb]
    \centering
    \includegraphics[width=\textwidth]{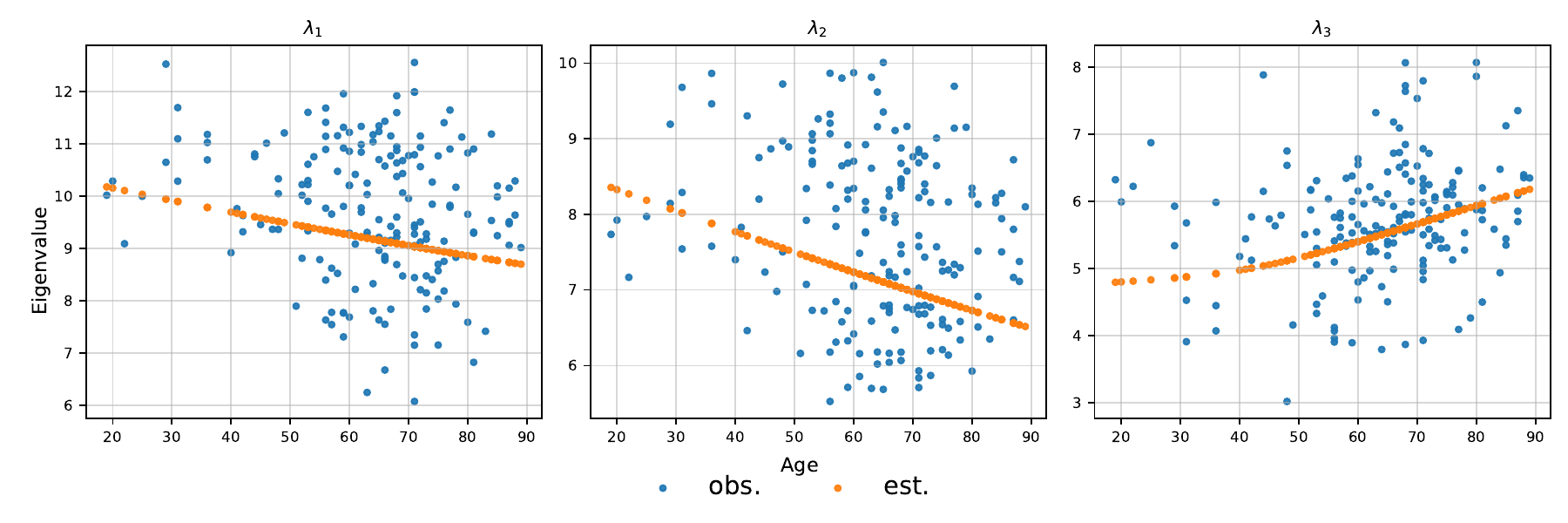}
    \caption{
        The three leading eigenvalues
        $\lambda_1, \lambda_2, \lambda_3$ (left to right) of the observed (blue) and fitted (red)
        covariances against age.
    }
    \label{fig: 3eigvals}
\end{figure}

Figure~\ref{fig: 3eigvals} views the age dependence through the eigenvalue spectrum, plotting the
three leading eigenvalues of the observed and fitted covariances against age: the two largest decrease and the third increases with age, and the Fr\'echet estimates track these trends.

\ignore{
\begin{remark}
By Property~\ref{prpt: sub_model}, fitting a Fr\'echet regression model with covariates $(\text{age}, z)$ and evaluating the fitted covariance at $(\text{age}, \hat z)$ yields the same results as in Figure~\ref{fig: data_mtrx} and \ref{fig: 3eigvals}, where $\hat z$ denotes the linear estimate of $z$ based on age.
\end{remark}
}




%% file: simulation_appdx.tex
\begin{figure}[!htb]
    \centering
    \includegraphics[width=.5\textwidth]{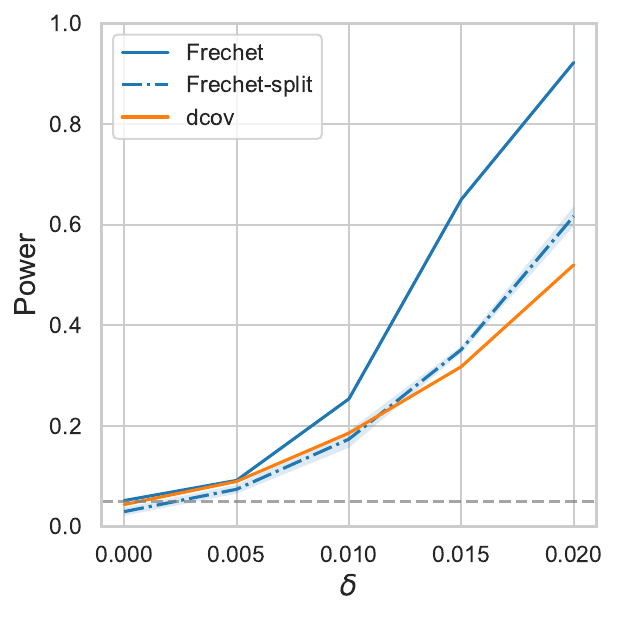}
    \caption{Empirical power of our test~\eqref{eqn: test} (Frechet), its sample-splitting counterpart (Frechet-split), and the distance-covariance test (dcov) as functions
    of the effect size $\delta_z$, for Example~\ref{example_c} with $d=6$, $n=200$ and
    $p_y=p_z=2$. Each point is based on 500 Monte Carlo replications. The
    sample-splitting curve is additionally averaged over five independent random splits; shaded bands show $\pm$ standard error.}
    \label{fig: split}
\end{figure}

\textit{Effect of sample splitting.} We assess the benefit of the split-free construction of $\cT_n$. Figure~\ref{fig: split} compares the split-free test with its sample-splitting counterpart and with \texttt{dcov}, under the setting of Example~\ref{example_c} with $d=6$. The sample-splitting curve is averaged over five independent random splits, with shaded bands of $\pm 1$ standard error. The split-free test dominates throughout, indicating that removing the split recovers the efficiency lost to data partitioning.

%% file: singlecell_appdx.tex
The selected genes are listed in decreasing order of $\underline{s}^{2}_{g}$ in Table~\ref{tab:selected_genes}, so the $d=10$
$d=15$ and $d=20$ subsets correspond to the first two, first three and all four rows, respectively.
\begin{table}[!htb]
    \centering
    \caption{Selected genes used in the single-cell data analysis.}
    \begin{tabular}{ccccc}
        \hline
        SOD1 & HSPA8 & PIK3R1 & RPS6KB2 & HDAC1 \\
        KRAS & PRKAG2 & PRKAA1 & SOD2 & EIF4EBP2 \\
        PRKACB & PRKAG1 & HDAC2 & ATG5 & AKT2 \\
        PIK3CD & AKT3 & IRS2 & CAT & PIK3CA \\
        \hline
    \end{tabular}
    \label{tab:selected_genes}
\end{table}

\begin{figure}[!htb]
    \centering
    \includegraphics[width=0.5\textwidth]{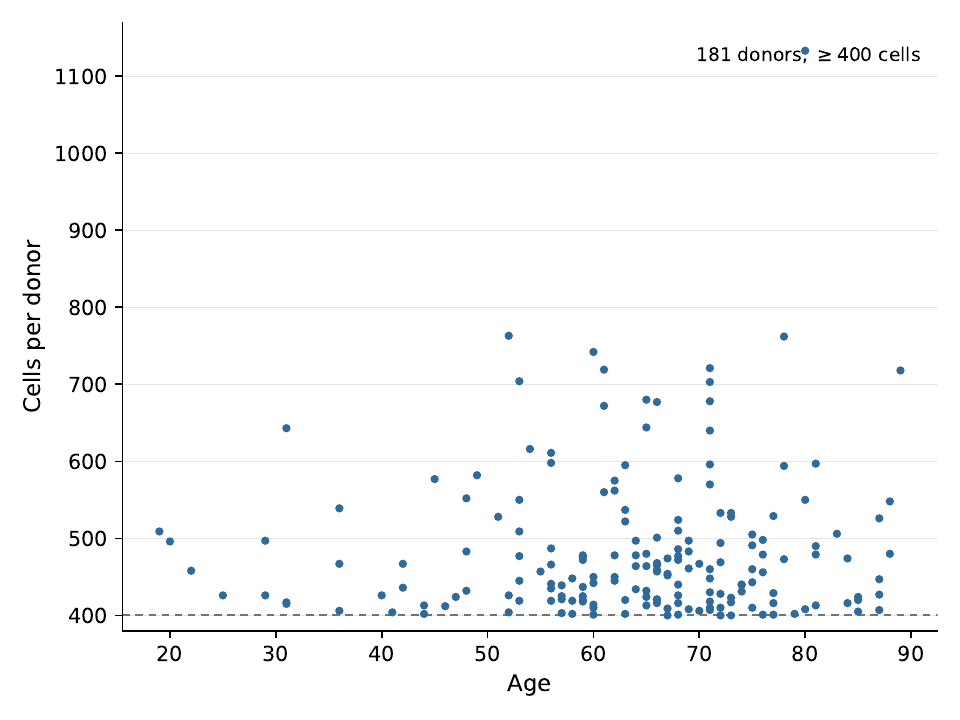}
    \caption{Distribution of the number of cells across the 181 donors included in the analysis.} \label{fig:append_cell}
\end{figure}

\begin{figure}[!htb]
    \centering
    \includegraphics[width=0.5\textwidth]{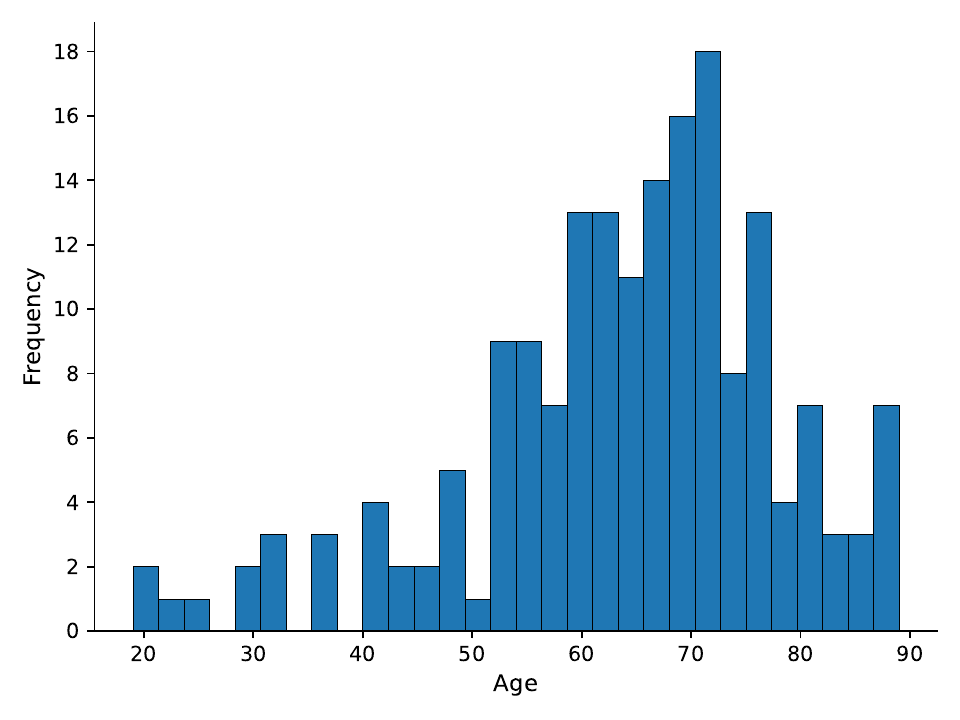}
    \caption{Age distribution of the 181 donors included in the analysis.} \label{fig:append_age}
\end{figure}

%% file: prf_est.tex
Recall that for any $\RR^p \ni x =(y,z)$, by definition we have
\begin{align*}
    \tx &= (y, \mu_Z+ \Sigma_{ZY} \Sigma_{YY}^{-1} (y-\mu_Y)),\\
    \hx &= (y, \hmu_Z + \hSigma_{ZY} \hSigma_{YY}^{-1} (y-\hmu_Y)).
\end{align*}
Lemma \ref{lem: matrix} implies that
\begin{align}
    F(\tx,S)&=\EE_{(X,Q)} \sbr{w(\tx,X) W^2(S,Q)} \nonumber\\
    &= \EE_{(Y,Q)} \sbr{\tw(y,Y)W^2(S,Q)}
    =: \tF(y,S) \label{eqn: prf_est_tF1},\\
    F_n(\hx,S)&= \frac{1}{n_1} \sum_{i=1}^{n_1} \sbr{w_n(\hx,X_i) W^2(S,Q_i)} \nonumber\\
    &= \frac{1}{n_1} \sum_{i=1}^{n_1} \sbr{\tw_n(y,Y_i) W^2(S,Q_i)}
    =: \tF_n(y,S). \label{eqn: prf_est_tFn1}
\end{align}
Hence
\begin{align*}
    Q^*(\tx) = \argmin_{S \in \cS^d_{++}} \tF(y,S),\\
    \hQ_n(\hx) = \argmin_{S \in \cS^d_{++}} \tF_n(y,S)
\end{align*}
Define $\tPP \in \cP(\RR^{p_1} \times \cS^d_{++})$ such that 
\begin{itemize}[leftmargin=*]
    \item for any Borel set $E_1 \in \cB(\RR^{p_1})$, 
    \begin{align*}
        \tPP_Y \sbr{Y \in E_1} = \PP_Y \sbr{Y \in E_1}
    \end{align*}
    \item for any Borel set $E_2 \in \cB(\cS^d_{++})$,
    \begin{align*}
        \tPP_Q \sbr{Q \in E_2  | Y=y} = \PP_Q \sbr{Q \in E_2 | X = (y, \mu_Z+ \Sigma_{ZY} \Sigma_{YY}^{-1} (y-\mu_Y))} 
    \end{align*}
\end{itemize}
Denote by $\tQ^*(y)$ the conditional expectation of $Q$ given $Y=y$ under $\tPP$, i.e.
\begin{align*}
    \tQ^*(y) &:= \argmin_{S \in \cS^d_{++}} \tEE_Q \sbr{W^2(S,Q) | Y=y}
\end{align*}

Then motivated by Lemma \ref{lem: xu25} from \cite{XuLi2025}, it suffices to show that $\tPP$ satifies the following conditions:
\begin{condition}  \label{assumption2: X}
    $Y$ is sub-Gaussian with $\norm{Y}_{\psi_2}\leq C_{\psi_2}$ and $\lambda_{\min}(\Sigma_{YY})\geq c_{\Sigma}$ for constants $C_{\psi_2},c_{\Sigma} >0$.
\end{condition}

\begin{condition} \label{assumption2: bdd_Q}
    Given $Y=y \in \supp Y$, the eigenvalues of $Q$ are bounded away from $0$ and infinity  in the sense that
    \begin{equation}
        \tPP\rbr{Q \in \cS^d \rbr{\tgamma_{\Lambda}(\norm{y-\mu_Y})^{-1},\tgamma_{\Lambda}(\norm{y-\mu_Y})} | Y=y} = 1
        \label{eqn2: asmptn_bdd_Q}
    \end{equation}
    where
    $\tgamma_{\Lambda}(t) := \tc_\Lambda \rbr{t \vee 1}^{\tC_\Lambda}$
    for constants $\tc_\Lambda \geq 1$ and $\tC_\Lambda \geq 0$.
\end{condition}

\begin{condition}  \label{assumption2: frechet}
    The Fr\'echet regression model holds. In other words, for any $y \in \supp Y$, the following equation is satisfied:
    \begin{equation}
        \tQ^*(y) = \argmin_{S \in \cS^d_{++}} \tF(y,S), \quad \text{where } \tF(y,S):=\EE_{(Y,Q)} \sbr{\tw(y,Y) W^2(S,Q)}
        \label{eqn2: frechet_assumption}
    \end{equation}
\end{condition}

\begin{condition} \label{assumption2: minimizer_global}
    For any $y \in \supp Y$, $\tF(y,\cdot):\cS^d_{++}\to \RR$ has a unique minimizer. Moreover, there exist constants $\talpha_F \geq 1$ and $\tdelta_F > 0$ such that for any $y \in \supp Y$ and any $(\delta,\Delta)$ that satisfies $0 \leq \delta\leq \delta_F \leq \Delta$,
    \begin{equation*}
        \inf \cbr{\tF(y,S) - \tF(y,\tQ^*(y)): \delta \leq \|S-\tQ^*(y)\|_{\mathrm{F}} \leq \Delta} \geq \frac{\delta^{\talpha_F}}{\tgamma_F(\norm{y-\mu},\Delta)}  
    \end{equation*}
    holds where
    $\tgamma_F(t_1,t_2) := \tc_F \rbr{t_1 \vee 1}^{\tC_F} \rbr{t_2 \vee 1}^{\tC_F}$
    for constants $\tc_F >0, \tC_F \geq 0$. 
\end{condition}

\begin{condition} \label{assumption2: minimizer_local}
    For any $y \in \supp Y$, consider the symmetric linear operator
    \begin{align*}
        \EE \rbr{-\tw(y,Y)dT_{\tQ^*(y)}^{Q}}: \cS^d \to \cS^d,
    \end{align*}
    which is the second differential of $\tF(y,\cdot)$ at $\tQ^*(y)$. This operator has a lower bound for its minimum eigenvalue given by:
    \begin{align}
        \lambda_{\min}\rbr{-\EE \tw(y,Y)dT_{\tQ^*(y)}^{Q}} &\geq \frac{1}{\tgamma_{\lambda}(\norm{y-\mu_Y})}, \label{eqn2: assmptn_min_eigen}
    \end{align}
    where 
    $\tgamma_{\lambda}(t) := \tc_{\lambda} \rbr{t \vee 1}^{\tC_{\lambda}}$
    for constants $\tc_{\lambda} > 0$ and $\tC_{\lambda} \geq 0$.
\end{condition}

\noindent \textbf{Condition \ref{assumption2: X}:} note that $Y$ is a subvector of $X$, hence Condition \ref{assumption2: X} follows from Assumption \ref{assumption: X}.

\noindent \textbf{Condition \ref{assumption2: bdd_Q}:} by definition of $\tPP$, we have for any $t>1$,
\begin{align*}
    &\tPP\sbr{Q \in \cS^d \rbr{t^{-1},t} | Y=y} \\
    = &\PP \sbr{Q \in \cS^d \rbr{t^{-1}, t} | X=(y, \mu_Z+ \Sigma_{ZY} \Sigma_{YY}^{-1} (y-\mu_Y))}.
\end{align*}
Then Condition \ref{assumption2: bdd_Q} follows by noticing that
\begin{align}
    \norm{y-\mu_Y} \leq 
    \norm{
        \begin{pmatrix}
            y\\
            \mu_Z+ \Sigma_{ZY} \Sigma_{YY}^{-1} (y-\mu_Y)
        \end{pmatrix}
        -
        \begin{pmatrix}
            \mu_Y\\
            \mu_Z
        \end{pmatrix}
    }
    \overset{(i)}{\lesssim} \norm{y-\mu_Y},
    \label{eqn: prf_est_normY}
\end{align}
where (i) follows from Assumption~\ref{assumption: X}.

\noindent \textbf{Condition \ref{assumption2: frechet}:} by definition of $\tQ^*(y)$, we have
\begin{align*}
    \tQ^*(y) &= \argmin_{S \in \cS^d_{++}} \tEE_Q \sbr{W^2(S,Q) | Y=y}\\
    &= \argmin_{S \in \cS^d_{++}} \EE_{Q} \sbr{W^2(S,Q) | X= (y, \mu_Z+ \Sigma_{ZY} \Sigma_{YY}^{-1} (y-\mu_Y))}\\
    &= Q^* \rbr{(y, \mu_Z + \Sigma_{ZY} \Sigma_{YY}^{-1} (y-\mu_Y))}.
\end{align*}
Moreover, Assumption~\ref{assumption: frechet} implies that
\begin{align*}
    Q^* \rbr{(y, \mu_Z + \Sigma_{ZY} \Sigma_{YY}^{-1} (y-\mu_Y))} 
    &= \argmin_{S \in \cS^d_{++}} F \rbr{(y, \mu_Z + \Sigma_{ZY} \Sigma_{YY}^{-1} (y-\mu_Y)), S}\\
    &= \argmin_{S \in \cS^d_{++}} \tF \rbr{y, S},
\end{align*}
where the last equality follows from \eqref{eqn: prf_est_tF1}. Hence Condition \ref{assumption2: frechet} holds.

\noindent \textbf{Condition \ref{assumption2: minimizer_global}, \ref{assumption2: minimizer_local}} follows by combining \eqref{eqn: prf_est_tF1}, \eqref{eqn: prf_est_normY} and Assumption~\ref{assumption: minimizer_global}, \ref{assumption: minimizer_local}.

Finally, applying Lemma \ref{lem: xu25} completes the proof.

%% file: prf_null.tex
We first present the following lemma, which simplifies the form of the test statistic. The proof is deferred to Section~\ref{sec: prf_reduction}.
\begin{lemma}[reduction] \label{lem: reduction}
    Instate the assumptions of Theorem \ref{thm: null}. Denote $R_i=(X_i,Q_i)$. Then we have
    \begin{align*}
        \cT_n = \frac{1}{n^2} \sum_{i,j,k \in [n]}^{} h (R_i, R_j; R_k) + o_p(1)
    \end{align*}
    where the function $h: \rbr{\RR^p \times \cS^d_{++}}^{3} \to \RR$ is defined as
    \begin{align*}
        h(r_1,r_2; r_3) = \inner{\tau(x_1,q_1; x_3)}{\tau(x_2,q_2; x_3)}
    \end{align*}
    for any $r_l = (x_l,q_l) \in \RR^p \times \cS^d_{++}$, $l \in \cbr{1,2,3}$.
\end{lemma}
With the symmetrization trick, we can further rewrite the leading term in Lemma \ref{lem: reduction} as a scaled V-statistic of order~3:
\begin{align*}
    \cT_n = \frac{1}{n^2} \sum_{i,j,k=1}^{n} \psi(R_i,R_j,R_k) + o_p(1)
\end{align*}
with symmetric kernel $\psi: \rbr{\RR^p \times \cS^d_{++}}^{3} \to \RR$ defined as
\begin{align*}
    \psi(r_1,r_2,r_3) &= \frac{1}{6} \sum_{\sigma} h(r_{\sigma(1)}, r_{\sigma(2)}; r_{\sigma(3)})\\
    &\overset{(i)}{=} \frac{1}{3} \sbr{h(r_1,r_2; r_3) + h(r_2,r_3; r_1) + h(r_3,r_1; r_2)}
\end{align*}
Here (i) follows from the fact that $h(\cdot,\cdot; \cdot)$ is symmetric in its first two arguments. Then the asymptotic distribution of $\cT_n$ can be derived by exploiting U-statistics theory in Appendix \ref{sec: tech_lemmas}. Specifically, by Lemma~\ref{lem: v_decomp}, we have
\begin{align*}
    \cT_n = \frac{(n-1)(n-2)}{n^2} \cdot nU_n^{(3)} + \frac{n-1}{n}\cdot 3 U_n^{(2)} + \frac{1}{n} U_n^{(1)} + o_p(1)
\end{align*}
where $U_n^{(3)}$, $U_n^{(2)}$, and $U_n^{(1)}$ are U-statistics with kernels
\begin{align*}
    \phi^{(3)}(r_1,r_2,r_3) &=  \psi(r_1,r_2,r_3),\\
    \phi^{(2)}(r_1,r_2) &= \frac{1}{2} \rbr{\psi(r_1,r_1,r_2) + \psi(r_1,r_2,r_2)},\\
    \phi^{(1)}(r_1) &= \psi(r_1,r_1,r_1).
\end{align*}
To get the asymptotic distribution of $nU_n^{(3)}$, $3U_n^{(2)}$, and $U_n^{(1)}/n$, note first that $U_n^{(1)}/n = o_p(1)$.

\noindent \textbf{For term $nU_n^{(3)}$,} we have
\begin{align*}
    \theta^{(3)} &= \EE \sbr{\phi^{(3)}(R_1,R_2,R_3)} = 0,\\
    \phi^{(3)}_1(r_1) &= \EE \sbr{\psi(r_1,R_2,R_3)} = 0, \;\ \text{and hence} \;\ \sbr{\sigma_1^{(3)}}^2 = 0,\\
    \phi^{(3)}_2(r_1,r_2) &= \EE \sbr{\psi(r_1,r_2,R_3)} = \frac{1}{3}\EE h(r_1,r_2;R_3) = \frac{1}{3} \cK(r_1;r_2)
\end{align*}
Here all three equalities exploit the fact that under the null hypothesis, we have $\EE_{R_1} {\tau(R_1;x)} = 0$ (see Lemma~\ref{lem: tau}).

Then from the asymptotic distribution of degenerate U-statistics (Lemma \ref{lem: u_asym_deg}), we obtain that
\begin{align*}
    nU_n^{(3)} \xrightarrow{d} \sum_{\nu=1}^{\infty} \lambda_\nu (Z_{\nu}^2 -1)
\end{align*}
where $Z_\nu$ are i.i.d. $N(0,1)$ and the $\lambda_\nu$ are the eigenvalues of $\cK$.

\noindent \textbf{For term $3U_n^{(2)}$,} we have
\begin{align*}
    \theta^{(2)} &= \EE \sbr{\phi^{(2)}(R_1,R_2)} = \EE \sbr{\psi(R_1,R_1,R_2)}\overset{(i)}{=}\frac{1}{3}\EE h(R_1,R_1;R_2) \overset{(ii)}{=}\frac{1}{3}\EE \cK(R_1;R_1)\overset{(iii)}{=}\frac{1}{3}\tr \cK
\end{align*}
Here (i) exploits  $\EE_{R_1} {\tau(R_1;x)} = 0$ again; (ii) follows from the definition of $\cK$; (iii) follows from the theory of integral operators.

Then from the asymptotic distribution of non-degenerate U-statistics (Lemma \ref{lem: u_asym_nondeg}), we obtain that
\begin{align*}
    3U_n^{(2)}=\tr \cK +o_p(1)
\end{align*}

Finally, combining the above results and the fact that $\tr \cK=\sum_{\nu=1}^{\infty} \lambda_\nu$, we complete the proof of Theorem~\ref{thm: null}.

\subsection{Proof of Lemma \ref{lem: reduction}}
\label{sec: prf_reduction}
First, note that Lemma \ref{lem: Q} implies that there exists an event $\cE_1$ that satisfies $\PP(\cE_1) \geq 1-O(n^{-100})$ such that
\begin{align*}
    \sup_{k \in [n]} \norm{\sbr{\frac{1}{n} \sum_{i=1}^{n} w_n(\hX_k,X_i) dT^{Q_i}_{\hQ_n(\hX_k)}} - \sbr{\frac{1}{n} \sum_{i=1}^{n} w_n(\hX_k,X_i) dT^{Q_i}_{Q^*(\hX_k)}}} \lesssim \frac{\polylog{n}}{\sqrt{n}}\\
    \sup_{k \in [n]} \Fnorm{\hQ_n(X_k) - Q^*(X_k)} \lesssim \frac{\polylog{n}}{\sqrt{n}}
\end{align*}
Hence we have
\begin{align*}
    \cT_n = \sum_{k=1}^{n} \Fnorm{\sbr{-\frac{1}{n} \sum_{i=1}^{n} w_n(\hX_k,X_i) dT^{Q_i}_{Q^*(\hX_k)}} \cdot \sbr{\hQ_n(X_k) - \hQ_n(\hX_k) }}^2 + o_p(1)
\end{align*}
Next, for any $x$, we have
\begin{align*}
    &\sbr{-\frac{1}{n} \sum_{i=1}^{n} w_n(\hx,X_i) dT^{Q_i}_{Q^*(\hx)}} \cdot \sbr{\hQ_n(x) - \hQ_n(\hx) }\\
    = &\sbr{-\frac{1}{n} \sum_{i=1}^{n} w_n(x,X_i) dT^{Q_i}_{Q^*(x)}} \cdot \sbr{\hQ_n(x) - Q^*(x) } -  \sbr{-\frac{1}{n} \sum_{i=1}^{n} w_n(\hx,X_i) dT^{Q_i}_{Q^*(\hx)}} \cdot \sbr{\hQ_n(\hx) - Q^*(x) }\\
    +& \sbr{\frac{1}{n} \sum_{i=1}^{n} \rbr{w_n(x,X_i) - w_n(\hx,X_i)} dT^{Q_i}_{Q^*(x)}} \cdot \sbr{\hQ_n(x) - Q^*(x) }
\end{align*}
Note that the optimality conditions at $x$ gives that
\begin{align*}
    \frac{1}{n} \sum_{i=1}^{n} w_n(x,X_i) \rbr{T^{Q_i}_{\hQ_n(x)} - I_d} = 0.
\end{align*}
which combined with Taylor expansions gives
\begin{align*}
    \frac{1}{n} \sum_{i=1}^{n} w_n(x,X_i) \rbr{T^{Q_i}_{Q^*(x)} - I_d} + \frac{1}{n} \sum_{i=1}^{n} w_n(x,X_i) dT^{Q_i}_{Q^*(x)} \cdot \rbr{\hQ_n(x) - Q^*(x)} + R_n(x) = 0
\end{align*}
where
\begin{align*}
    R_n(x) = \frac{1}{2n} \sum_{i=1}^{n} w_n(x,X_i) d^2 T^{Q_i}_{\tQ_n(x)} \cdot \rbr{\hQ_n(x) - Q^*(x)}^{\otimes 2}
\end{align*}
Here Lemma \ref{lem: Q} implies that with probability at least $1-O(n^{-100})$, we have
\begin{align*}
    \sup_{k \in [n]} \Fnorm{R_n(X_k)} \lesssim \frac{\polylog{n}}{n}
\end{align*}
Hence we obtain that for any $k \in [n]$, the following holds:
\begin{align*}
    &\sbr{-\frac{1}{n} \sum_{i=1}^{n} w_n(\hX_k,X_i) dT^{Q_i}_{Q^*(\hX_k)}} \cdot \sbr{\hQ_n(X_k) - \hQ_n(\hX_k) }\\
    = &\underbrace{\frac{1}{n} \sum_{i=1}^{n} w_n(X_k,X_i) \rbr{T^{Q_i}_{Q^*(X_k)} - I_d}}_{=:A_1(X_k)} - \underbrace{\frac{1}{n} \sum_{i=1}^{n} w_n(\hX_k,X_i) \rbr{T^{Q_i}_{Q^*(\hX_k)} - I_d}}_{A_1(\hX_k)} \\ 
    + &\underbrace{\sbr{\frac{1}{n} \sum_{i=1}^{n} \rbr{w_n(X_k,X_i) - w_n(\hX_k,X_i)} dT^{Q_i}_{Q^*(X_k)}} \cdot \sbr{\hQ_n(X_k) - Q^*(X_k) }}_{A_2(X_k,\hX_k)} + O_p\rbr{\frac{\polylog{n}}{n}}
\end{align*}
where the $O_p\rbr{n^{-1}\polylog{n}}$ term is uniform in $k$.

As a result, we have
\begin{align*}
    \cT_n = \sum_{k=1}^{n} \Fnorm{A_1(X_k) - A_1(\hX_k) + A_2(X_k,\hX_k)}^2 + o_p(1)
\end{align*}
We proceed to analyze the three terms $A_1(X_k)$, $A_1(\hX_k)$ and $A_2(X_k,\hX_k)$.

\noindent \textbf{Analysis of Term $A_2(X_k,\hX_k)$.} By applying a Taylor expansion to the optimality condition and invoking Lemma \ref{lem: Q}, we can obtain that
\begin{align*}
    &\sbr{\hQ_{n}(X_k) - Q^*(X_k) }\\
    = &\sbr{\frac{1}{n} \sum_{i=1}^{n} w_{n}(x,X_i) dT^{Q_i}_{Q^*(x)}}^{-1} \frac{1}{n} \sum_{i=1}^{n} w_{n}(X_k,X_i) \rbr{T^{Q_i}_{Q^*(X_k)} - I_d}
    + O_p\rbr{\frac{\polylog{n}}{n}}\\
    = &\sbr{ -\EE w(x,X')dT^{Q'}_{Q^*(x)} }^{-1} \frac{1}{n} \sum_{i=1}^{n} w_{n}(x,X_i) \rbr{T^{Q_i}_{Q^*(x)} - I_d} + O_p\rbr{\frac{\polylog{n}}{n}}
\end{align*}
where the $O_p\rbr{n^{-1}\polylog{n} }$ term is uniform in $k$. Also we have
\begin{align*}
    &\frac{1}{n} \sum_{i=1}^{n} \rbr{w_{n}(X_k,X_i) - w_{n}(\hX_k,X_i)} dT^{Q_i}_{Q^*(X_k)}\\
    = &\EE_{(X',Q')} \rbr{w(x,X') - w(\tx,X')} dT^{Q'}_{Q^*(x)} + O_p\rbr{\frac{\polylog{n}}{\sqrt{n}}}
\end{align*}
where $\tx=(y, \EE Z + \Sigma_{21}\Sigma_{11}^{-1}(y- \EE Y) )$ and the $O_p\rbr{\frac{\polylog{n}}{\sqrt{n}}}$ term is uniform in $k$. Combining the above two equations, we have
\begin{align*}
    &A_2(X_k,\hX_k)=\\
    & \sbr{\EE' \rbr{w(X_k,X') - w(\tX_k,X')} dT^{Q'}_{Q^*(x)}} \sbr{ -\EE' w(X_k,X')dT^{Q'}_{Q^*(X_k)} }^{-1} \frac{1}{n} \sum_{i=1}^{n} w_{n}(X_k,X_i) \rbr{T^{Q_i}_{Q^*(X_k)} - I_d} \\
    &+ O_p\rbr{\frac{\polylog{n}}{n}} 
\end{align*}
where $\EE'$ denotes the expectation with respect to the distribution of $(X',Q')$.
Here the $O_p\rbr{n^{-1} \polylog{n}}$ term is uniform in $k$.

\noindent \textbf{Analysis of Term $A_1(\hX_k)$.} By the definition of $w_{n}$, we have that
\begin{align*}
    A_1(\hx)=&\frac{1}{n} \sum_{i=1}^{n} w_{n}(\hx,X_i) \rbr{T^{Q_i}_{Q^*(\hx)} - I_d}\\
    = &\frac{1}{n} \sum_{i=1}^{n} \rbr{1+ (\hx-\bar{X})^\top \hSigma^{-1}(X_i - \bar{X}) } \rbr{T^{Q_i}_{Q^*(x)} - I_d}
\end{align*}
Recall that for any $x=(y,z) \in \RR^p$, $\hx \in \RR^p$ is defined as $\hx=(y,  \bar{Z} + \hSigma_{21} \hSigma_{11}^{-1} (y-\bar{Y}))$. Applying Lemma~\ref{lem: matrix} gives the following claim, which can be verified by direct calculation:
\begin{claim}[\ref{claim: Qy}]
    \label{claim: Qy}
    We have
    \begin{align*}
        (\hx-\bar{X})^\top \hSigma^{-1}(X_i - \bar{X}) = (y-\bar{Y}) \hSigma_{11}^{-1}(Y_i - \bar{Y})
    \end{align*}
\end{claim}
As a result, we have
\begin{align*}
    &\frac{1}{n} \sum_{i=1}^{n} w_{n}(\hx,X_i) \rbr{T^{Q_i}_{Q^*(\hx)} - I_d}\\
    = &\frac{1}{n} \sum_{i=1}^{n} \rbr{1+ (y-\bar{Y}) \hSigma_{11}^{-1}(Y_i - \bar{Y}) } \rbr{T^{Q_i}_{Q^*(x)} - I_d}\\
    = &\frac{1}{n} \sum_{i=1}^{n} \vy^\top \hat{\Xi}_{11}^{-1} \vY_i \rbr{T^{Q_i}_{Q^*(x)} - I_d}\\
    = & \frac{1}{n} \sum_{i=1}^{n} \vy^\top \Xi_{11}^{-1} \vY_i \rbr{T^{Q_i}_{Q^*(x)} - I_d} \\
    - &\frac{1}{n} \sum_{i=1}^{n} \sbr{\vy^\top \Xi_{11}^{-1}\rbr{\vY_i \vY_i^\top - \Xi_{11}} \Xi_{11}^{-1} \otimes \id}\times \sbr{\EE Y'\otimes \rbr{T^{Q'}_{Q^*(x)} - I_d}}\\
    + &\Delta_n(x)
\end{align*}
where
\begin{align*}
    \hDelta_n(x) = \sbr{\vy^\top \Xi_{11}^{-1}\rbr{\hXi_{11} - \Xi_{11}}\rbr{\hXi_{11}^{-1} - \Xi_{11}^{-1}}  \otimes \id}\times \sbr{\EE Y'\otimes \rbr{T^{Q'}_{Q^*(x)} - I_d}}
\end{align*}
given that $Q^*(\hx)=Q^*(x)$ under the null. Applying Lemma \ref{lem: Q} gives that with probability at least $1-O(n^{-100})$, we have
\begin{align*}
    \sup_{k \in [n]} \Fnorm{\hDelta_n(X_k)} \lesssim \frac{\polylog{n}}{n}
\end{align*}

\noindent \textbf{Analysis of Term $A_1(X_k)$.} Similarly, we have
\begin{align*}
    A_1(x) &= \frac{1}{n} \sum_{i=1}^{n} w_{n}(x,X_i) \rbr{T^{Q_i}_{Q^*(x)} - I_d}\\
    &= \frac{1}{n} \sum_{i=1}^{n} \vx^\top {\hXi}^{-1} \vX_i \rbr{T^{Q_i}_{Q^*(x)} - I_d}\\
    &= \frac{1}{n} \sum_{i=1}^{n} \vx^\top \Xi^{-1} \vX_i \rbr{T^{Q_i}_{Q^*(x)} - I_d} \\
    - &\frac{1}{n} \sum_{i=1}^{n} \sbr{\vx^\top \Xi^{-1} \rbr{\vX_i \vX_i^\top - \Xi} \Xi^{-1} \otimes \id}\times \sbr{\EE X'\otimes \rbr{T^{Q'}_{Q^*(x)} - I_d}}\\
    + & \Delta_n(x) 
\end{align*}
where
\begin{align*}
    \Delta_n(x) = \sbr{\vx^\top \Xi^{-1}\rbr{\hXi - \Xi}\rbr{\hXi^{-1} - \Xi^{-1}}  \otimes \id}\times \sbr{\EE X'\otimes \rbr{T^{Q'}_{Q^*(x)} - I_d}}
\end{align*}
Applying Lemma \ref{lem: Q} gives that with probability at least $1-O(n^{-100})$, we have
\begin{align*}
    \sup_{k \in [n]} \Fnorm{\Delta_n(X_k)} \lesssim \frac{\polylog{n}}{n}
\end{align*}

\noindent \textbf{Combining} results on $A_1(X_k)$, $A_1(\hX_k)$ and $A_2(X_k,\hX_k)$, we have
\begin{align*}
    \cT_n = \sum_{k=1}^{n} \Fnorm{\frac{1}{n} \sum_{i=1}^{n}\tau(X_i,Q_i;X_k) }^2 + o_p(1)
\end{align*}
where
\begin{align*}
    \tau(X,S;x) &= \tau_0(X,S;x) - \tau_1(X,S;x) + A(x) \tau_0(X,S;x)\\
    \tau_0(X,S;x) &= \vx^\top \Xi^{-1} \vX \rbr{T^{S}_{Q^*(x)} - I_d} - \sbr{\vx^\top \Xi^{-1}\rbr{\vX \vX^\top - \Xi} \Xi^{-1} \otimes \id}\times \sbr{\EE X'\otimes \rbr{T^{Q'}_{Q^*(x)} - I_d}}\\
    \tau_1(X,S;x) &= \vy^\top \Xi_{11}^{-1} \vY \rbr{T^{S}_{Q^*(x)} - I_d} - \sbr{\vy^\top \Xi_{11}^{-1}\rbr{\vY \vY^\top - \Xi_{11}} \Xi_{11}^{-1} \otimes \id}\times \sbr{\EE Y'\otimes \rbr{T^{Q'}_{Q^*(x)} - I_d}}\\
    A(x) &= \sbr{\EE_{(X',Q')} \rbr{w(x,X') - w(\tx,X')} dT^{Q'}_{Q^*(x)}} \cdot \sbr{ -\EE w(x,X')dT^{Q'}_{Q^*(x)} }^{-1}\\
    \tx &= (y, \mu_Z + \Sigma_{21} \Sigma_{11}^{-1} (y-\mu_Y))
\end{align*}

Finally, note that
\begin{align*}
    \sum_{k=1}^{n} \Fnorm{\frac{1}{n} \sum_{i=1}^{n}\tau(X_i,Q_i;X_k) }^2 &= \frac{1}{n^2} \sum_{i,j,k=1}^{n} \inner{\tau(X_i,Q_i;X_k)}{\tau(X_j,Q_j;X_k)}
\end{align*}
The proof is complete.

%% file: global_null.tex
For global testing, we have $z=x$. Hence under the assumptions of Corrolary \ref{cor: glb_null}, we have, for any $x,x' \in \RR^p$ and $S \in \cS^d_{++}$,
\begin{align*}
    \tau_0(x,S;x')&=\vX^\top \Xi^{-1} \vX' \rbr{T^{S}_{Q^*(\mu)} - I_d}=\sbr{1+(x-\mu)\Sigma^{-1}(X-\mu)}\rbr{T^{S}_{Q^*(\mu)} - I_d}\\
    \tau_1(x,S;x') &= T_{Q^*(\mu)}^S - I_d\\
    A(x) &= 0
\end{align*}
As a result, we have
\begin{align*}
    \tau(x,S;X) = (x-\mu)^\top\Sigma^{-1}(X-\mu)\rbr{T^{S}_{Q^*(\mu)} - I_d}
\end{align*}
which implies that
\begin{align*}
    \cK(x,S;x',S') &= (x-\mu)^\top\Sigma^{-1}(x'-\mu) \inner{T^{S}_{Q^*(\mu)} - I_d}{T^{S'}_{Q^*(\mu)} - I_d}_{\mathrm{F}}
\end{align*}
Note that under the null and conditional independence, $X \perp Q$. Then the desired result follows from the spectral analysis of the integral operator associated with this kernel.

%% file: prf_size.tex
Recall that $\lambda_1 \geq \lambda_2 \geq \cdots $ are eigenvalues of the integral operator $\cK$, let $S = \sum_{j=1}^{\infty} \lambda_j w_j$ where $\cbr{w_j}_{j \geq 1}$ are iid chi-squared random variables.

Note that for any $\epsilon>0$, we have
\begin{align*}
    \PP\rbr{\cT_n > \hq_{1-\alpha}} \leq \PP\rbr{\cT_n > q_{1-\alpha}-\epsilon} + \PP\rbr{\abs{\hq_{1-\alpha} - q_{1-\alpha}} > \epsilon}
\end{align*}
Since $\epsilon>0$ is arbitrary, taking $\limsup_{\epsilon \to 0} \limsup_{n \to \infty}$ on both sides gives
\begin{align*}
    \limsup_{n\to \infty} \PP\rbr{\cT_n > \hq_{1-\alpha}} &\leq \limsup_{\epsilon \to 0} \limsup_{n \to \infty} \PP\rbr{\cT_n > q_{1-\alpha}-\epsilon} + \limsup_{\epsilon \to 0} \limsup_{n \to \infty} \PP\rbr{\abs{\hq_{1-\alpha} - q_{1-\alpha}} > \epsilon}\\
    &\overset{(i)}{=} \limsup_{\epsilon \to 0}  \PP \rbr{S > q_{1-\alpha}-\epsilon} + \limsup_{\epsilon \to 0} \limsup_{n \to \infty} \PP\rbr{\abs{\hq_{1-\alpha} - q_{1-\alpha}} > \epsilon}\\
    &= \PP\rbr{S \geq q_{1-\alpha}} + \limsup_{\epsilon \to 0} \limsup_{n \to \infty} \PP\rbr{\abs{\hq_{1-\alpha} - q_{1-\alpha}} > \epsilon}\\
    &= \alpha + \limsup_{\epsilon \to 0} \limsup_{n \to \infty} \PP\rbr{\abs{\hq_{1-\alpha} - q_{1-\alpha}} > \epsilon}
\end{align*}
Here (i) follows from weak convergence $\cT_n \xrightarrow{d} S$ which is proved in Theorem \ref{thm: null} and the fact that $S$ have continuous distribution (if $X,Y$ are independent random variables and $X$ has continuous distribution, then $X+Y$ also has continuous distribution).

Similarly, we can show that
\begin{align*}
    \liminf_{n\to \infty} \PP\rbr{\cT_n > \hq_{1-\alpha}} \geq \alpha + \liminf_{\epsilon \to 0} \liminf_{n \to \infty} \PP\rbr{\abs{\hq_{1-\alpha} - q_{1-\alpha}} > \epsilon}
\end{align*}

Therefore, it suffices to show $\hq_{1-\alpha} \xrightarrow{p} q_{1-\alpha}$. Indeed, given $\hq_{1-\alpha} \xrightarrow{p} q_{1-\alpha}$, we have
\begin{align*}
    \lim_{n \to \infty} \PP\rbr{\abs{\hq_{1-\alpha} - q_{1-\alpha}} > \epsilon } = 0
\end{align*}
which implies that $\PP(\cT > \hq_{1-\alpha}) \to 0$ and completes the proof of Theorem \ref{thm: size}.

We now proceed to prove $\hq_{1-\alpha} \xrightarrow{p} q_{1-\alpha}$. Recall that $\hlambda_1 \geq \hlambda_2 \geq \cdots \geq \hlambda_n \geq 0$ are the eigenvalues of the matrix $\hbK_n \in \RR^{n \times n}$. By Lemma \ref{lem: quantile_conv}, it suffices to show that
\begin{align}
    \sum_{i=1}^{\infty} \lambda_j &< \infty \label{eqn: prf_sz_tr_k}\\
    \sum_{j=1}^{n} \hlambda_j &= O_p(1) \label{eqn: prf_sz_tr_hk}\\
    \abs{\sum_{j=1}^{n} \hlambda_j - \sum_{j=1}^{\infty} \lambda_j} &\xrightarrow{p} 0 \label{eqn: prf_sz_tr_diff}\\
    \sqrt{\sum_{j \geq 1}^{} \rbr{\hlambda_j - \lambda_j}^2 } &\xrightarrow{p} 0 \label{eqn: prf_sz_tr_l2}
\end{align}
Here in the last limit, we adopt the convention that $\hlambda_j = 0$ for $j>n$. Note that \eqref{eqn: prf_sz_tr_hk} is implied by \eqref{eqn: prf_sz_tr_k} and \eqref{eqn: prf_sz_tr_diff}, hence we only need to prove \eqref{eqn: prf_sz_tr_k}, \eqref{eqn: prf_sz_tr_diff} and \eqref{eqn: prf_sz_tr_l2}.

For notation simplicity, we denote $R_i = (X_i,Q_i)$ for any $i \in [n]$ in the rest of this proof.

\noindent \textbf{Proof of \eqref{eqn: prf_sz_tr_k}:} Note that
\begin{align*}
    \sum_{j=1}^{\infty} \lambda_j = \tr \cK &= \EE_R \cK(R;R)\\
    &= \EE_{(X,Q) \otimes X'} \Fnorm{\tau(X,Q;X')}^2\\
    &\overset{(i)}{\lesssim} \EE_{(X,Q) \otimes X'} \sbr{\rbr{1 \vee \norm{X}} \rbr{1 \vee \Onorm{Q} \vee \Onorm{Q^{-1}}} \rbr{1 \vee \norm{X'}}}^2\\
    &\overset{(ii)}{<} \infty
\end{align*}
Here (i) follows from Lemma \ref{lem: tau} and (ii) follows from Condition \ref{assumption: X} and \ref{assumption: bdd_Q}.

\noindent \textbf{Proof of \eqref{eqn: prf_sz_tr_diff} and \eqref{eqn: prf_sz_tr_l2}:} To prove \eqref{eqn: prf_sz_tr_diff} and \eqref{eqn: prf_sz_tr_l2}, we introduce more notations. Recall
\begin{align*}
    \hat{\cK}_n(r,r') = \frac{1}{n} \sum_{k=1}^{n} \inner{\htau(r;X_k)}{\htau(r';X_k)}
\end{align*}
Define
\begin{align*}
    &\bar{\cK}_n(r,r') = \frac{1}{n}\sum_{k=1}^{n} \inner{\tau(r;X_k)}{\tau(r';X_k)}\\
    &\bK_n \in \RR^{n \times n} \text{ with entries } \sbr{\bK_n}_{i,j} = \frac{1}{n} \cK(R_i,R_j)\\
    &\bar{\bK}_n \in \RR^{n \times n} \text{ with entries } \sbr{\bar{\bK}_n}_{i,j} = \frac{1}{n} \bar{\cK}_n(R_i,R_j)
\end{align*}
Let $\cbr{\tlambda_j}_{j \in [n]}$ and $\cbr{\barlambda_j}_{j \in [n]}$ be the eigenvalues of the matrix $\bK_n$ and $\bar{\bK}_n$ respectively. The notation is summarized in the following table:
\begin{table}[!htb]
    \tbl{Summary of eigenvalue notations}
    {\begin{tabular}{|c|c|c|c|c|}
        \hline
        Eigenvalue & $\lambda$ & $\tlambda$ & $\barlambda$ & $\hlambda$\\
        \hline
        Matrix/Operator & $\cK$ & $\bK_n$ & $\bar{\bK}_n$ & $\hbK_n$ \\
        \hline
        Entry $(i,j)$ & $-$ & $\cK(R_i,R_j)/n$ & $\bar{\cK}_n(R_i,R_j)/n$ & $\hat{\cK}_n(R_i,R_j)/n$\\
        \hline
    \end{tabular}}    
\end{table}
We approximate $\lambda$ by $\hlambda$ through the path
\begin{align*}
    \lambda \xleftarrow{\text{Lemma \ref{lem: approx_l_tl}}} \tlambda \xleftarrow{\text{Lemma \ref{lem: approx_tl_bl}}} \barlambda \xleftarrow{\text{Lemma \ref{lem: approx_bl_hl}}} \hlambda
\end{align*}

\begin{lemma}
    \label{lem: approx_l_tl}
    Instate the assumptions of Theorem \ref{thm: size}. Then there exists event $\cE_{\ref{thm: size},1}$ with probability at least $1 - O(n^{-100})$ under which
    \begin{align*}
    \abs{\sum_{j=1}^{\infty} \lambda_j - \sum_{j=1}^{n} \tlambda_j} &\lesssim \frac{\polylog{n}}{\sqrt{n}}\\
    \sqrt{\sum_{j \geq 1}^{} \rbr{\lambda_j - \tlambda_j}^2 } &\lesssim \frac{\polylog{n}}{\sqrt{n}}
    \end{align*}
\end{lemma}
See Appendix \ref{sec: prf_approx_l_tl} for proof.

\begin{lemma}
    \label{lem: approx_tl_bl}
    Instate the assumptions of Theorem \ref{thm: size}. Then there exists event $\cE_{\ref{thm: size},2}$ with probability at least $1 - O(n^{-98})$ under which
    \begin{align*}
        \abs{\sum_{j=1}^{n} \tlambda_j - \sum_{j=1}^{n} \barlambda_j} &\lesssim \frac{\polylog{n}}{\sqrt{n}}\\
        \sqrt{\sum_{j \geq 1}^{} \rbr{\tlambda_j - \barlambda_j}^2 } &\lesssim \frac{\polylog{n}}{\sqrt{n}}
    \end{align*}
\end{lemma}
See Appendix \ref{sec: prf_approx_tl_bl} for proof.

\begin{lemma}
    \label{lem: approx_bl_hl}
    Instate the assumptions of Theorem \ref{thm: size}. Then there exists event $\cE_{\ref{thm: size},3}$ with probability at least $1 - O(n^{-98})$ under which
    \begin{align*}
        \abs{\sum_{j=1}^{n} \barlambda_j - \sum_{j=1}^{n} \hlambda_j} &\lesssim \frac{\polylog{n}}{\sqrt{n}}\\
        \sqrt{\sum_{j \geq 1}^{} \rbr{\barlambda_j - \hlambda_j}^2 } &\lesssim \frac{\polylog{n}}{\sqrt{n}}
    \end{align*}
\end{lemma}
See Appendix \ref{sec: prf_approx_bl_hl} for proof. 

With Lemma \ref{lem: approx_l_tl}, \ref{lem: approx_tl_bl} and \ref{lem: approx_bl_hl}, \eqref{eqn: prf_sz_tr_diff} and \eqref{eqn: prf_sz_tr_l2} can be obtained from triangle inequalities.

Finally, combining the results above, the proof of Theorem \ref{thm: size} is now complete.


\subsection{Proof of Lemma \ref{lem: approx_l_tl}}
\label{sec: prf_approx_l_tl}
Recall notation $R=(X,Q)$. Denote $\cR = \RR^p \times \cS^d_{++}$.

\noindent \textbf{$\ell_2$ norm:} 
Let $T: L^2(\cV,\PP) \to L^2(\cV,\PP)$ be the integral operator induced by kernel $\cK$. Let $\cH$ be the RKHS associated with $\cK$. For any $r \in \cR$, let $\cK_{r} \in \cH$ be the function $\cK_{n}(r, \cdot)$. Define operators $\sT, \sT_{n}: \cH \to \cH$ as follows:
\begin{align*}
    \sT &= \int \inner{\cdot}{{\cK}_{r}}_{\cH} {\cK}_{r} \PP(dr)\\
    \sT_{n} &= \frac{1}{n} \sum_{i=1}^{n} \inner{\cdot}{{\cK}_{R_i}}_{\cH} {\cK}_{R_i}
\end{align*}
Then $\sT_{n}$ has the same strictly positive eigenvalues as $\bK_n$. Also, the operator $\sT$ has the same strictly positive eigenvalues $\lambda_1 \geq \lambda_2 \geq \cdots$ as $\cK$. The proof can be found in \citet{rkhs10}. Moreover, \citet{bhatia94,markus64} showed that
\begin{align*}
    \norm{\rbr{\lambda_j}_{j \geq 1} - \rbr{\tlambda_j}_{j \geq 1}}_{2} &\leq \norm{\sT - \sT_n}_{\mathrm{HS}}
\end{align*}

Then it suffices to upper bound $ \norm{\sT_n - \sT}_{\mathrm{HS}}$. Note that Lemma \ref{lem: tau} implies that for any $r = (x,q) \in \RR^p \times \cS^d_{++}$, we have
\begin{align*}
    \norm{\inner{\cdot}{{\cK}_{r}}_{\cH} {\cK}_{r}}_{\mathrm{HS}} &= \norm{{\cK}_{r}}_{\cH}^2
    = {\cK}(r;r) 
    \leq \sbr{\rbr{1 \vee \norm{x}} \rbr{1 \vee q} }^{C}
\end{align*}
Then from Assumption~\ref{assumption: X}, \ref{assumption: bdd_Q} and concentration in Hilbert space, there exists an event $\cE_{2,0}$ with probability at least $1 - O(n^{-100})$ such that
\begin{align*}
    \norm{\sT - \sT_n}_{\mathrm{HS}} &\lesssim \frac{\polylog{n}}{\sqrt{n}}
\end{align*}

\noindent \textbf{Difference of sum:} note that
\begin{align*}
    \sum_{j = 1}^{n} \tlambda_j - \sum_{j \geq 1}^{} \lambda_j
    &= \frac{1}{n} \sum_{i = 1}^{n} {\cK}(R_i,R_i) - \EE_{R_{1}} {\cK}(R_{1},R_{1})
\end{align*}
Then Conditions \ref{assumption: X}, \ref{assumption: bdd_Q} and concentration inequality (Lemma \ref{lem: arun22}) imply that there exists an event $\cE_{2,2}$ with probability at least $1 - O(n^{-100})$ under which we have
\begin{align*}
    \abs{\sum_{j = 1}^{n} \tlambda_j - \sum_{j \geq 1}^{} \lambda_j} &\lesssim \frac{\polylog{n}}{\sqrt{n}}
\end{align*}

\subsection{Proof of Lemma \ref{lem: approx_tl_bl}}
\label{sec: prf_approx_tl_bl}
It suffices to show that
\begin{align*}
    \abs{\tr \bK_n - \tr \bar{\bK}_n} &\lesssim \frac{\polylog{n}}{\sqrt{n}}\\
    \Fnorm{\bK_n - \bar{\bK}_n} &\lesssim \frac{\polylog{n}}{\sqrt{n}}
\end{align*}
Note that for each $i,j \in [n]$, the kernel $\bar{\cK}_n$ is weakly correlated with $R_i,R_j$ which makes the concentration analysis of $\bar{\bK}_n$ more involved. To this end, we introduce leave-two-out estimates. Recall that kernel matrix~$\bK_n \in \RR^{n \times n}$ has entries
\begin{align*}
    \sbr{\bK_n}_{ij} = \frac{1}{n} \cK(R_i,R_j) = \frac{1}{n} \EE_{X} \inner{\tau(R_i;X)}{\tau(R_j;X)}
\end{align*}
Define the leave-two-out estimate $\bar{\bK}_n^{-} \in \RR^{n \times n}$ with entries
\begin{align*}
    n\sbr{\bar{\bK}_n^{-}}_{ij} &= \frac{1}{n-2} \sum_{k \neq i,j}^{n} \inner{\tau(R_i;X_k)}{\tau(R_j;X_k)}, \quad i \neq j\\
    n\sbr{\bar{\bK}_n^{-}}_{ii} &= \frac{1}{n-1} \sum_{k \neq i}^{n} \inner{\tau(R_i;X_k)}{\tau(R_i;X_k)}, \quad i \in [n]
\end{align*}
\noindent \textbf{1. Bounding ($\bar{\bK}_n^- - \bK_n$).} Note that by independence, we have
\begin{align*}
    \EE \sbr{\sbr{\bar{\bK}_n^{-}}_{ij} \mid R_i,R_j} &= \sbr{\bK_n}_{ij}, \quad i \neq j\\
    \EE \sbr{\sbr{\bar{\bK}_n^{-}}_{ii} \mid R_i} &= \sbr{\bK_n}_{ii} , \quad i \in [n]
\end{align*}
Recall that we denote $R=(X,Q) \in \RR^p \times \cS^d_{++}$, with a little abuse of notation, we define $\norm{R}:= \norm{X} \vee \Onorm{Q} \vee \Onorm{Q^{-1}}$ (note that this is actually not a "norm"). Then
Lemma \ref{lem: tau} implies that there exists a constant $C_\tau>0$ such that for any $i,j,k \in [n]$,
\begin{align*}
    \abs{\inner{\tau(R_i;X_k)}{\tau(R_j;X_k)}} &\lesssim \sbr{\rbr{1 \vee \norm{X_k}} (1 \vee \norm{R_i})(1 \vee \norm{R_j})}^{C_\tau}
\end{align*}
Then concentration (Lemma \ref{lem: arun22}) together with Conditions \ref{assumption: X}, \ref{assumption: bdd_Q} and a union bound over all entries imply that there exists an event $\cE_{3}$ with probability at least $1 - O(n^{-98})$ such that under $\cE_3$, we have
\begin{align*}
    n\max_{i,j \in [n]} \abs{\sbr{\bar{\bK}_n^-}_{ij} - \sbr{\bK_n}_{ij}} &\lesssim \frac{\polylog{n}}{\sqrt{n}}
\end{align*}
\noindent \textbf{2. Bounding ($\bar{\bK}_n^- - \bar{\bK}_n$).} Note that for any $i \neq j \in [n]$,
\begin{align*}
    n\sbr{\bar{\bK}_n}_{ij} - n\sbr{\bar{\bK}_n^-}_{ij} &= \rbr{\frac{1}{n}-\frac{1}{n-2}} \sum_{k \neq i,j}^{n} \inner{\tau(R_i;X_k)}{\tau(R_j;X_k)} \\
    & + \frac{1}{n} \sbr{\inner{\tau(R_i;X_i)}{\tau(R_j;X_i)} + \inner{\tau(R_i;X_j)}{\tau(R_j;X_j)}}
\end{align*}
For any $i \in [n]$,
\begin{align*}
    n\sbr{\bar{\bK}_n}_{ii} - n\sbr{\bar{\bK}_n^-}_{ii} &= \rbr{\frac{1}{n}-\frac{1}{n-1}} \sum_{k \neq i}^{n} \inner{\tau(R_i;X_k)}{\tau(R_i;X_k)} + \frac{1}{n} \inner{\tau(R_i;X_i)}{\tau(R_i;X_i)}
\end{align*}
Therefore, Lemma \ref{lem: tau} together with Conditions \ref{assumption: X}, \ref{assumption: bdd_Q} and a union bound over all entries imply that there exists an event $\cE_4$ with probability at least $1 - O(n^{-99})$ such that under $\cE_4$, we have
\begin{align*}
    n \max_{i,j \in [n]} \abs{\sbr{\bar{\bK}_n}_{ij} - \sbr{\bar{\bK}_n^-}_{ij}} &\lesssim \frac{\polylog{n}}{n}
\end{align*}
Combining results in 1. and 2., we obtain that under the event $\cE_3 \cap \cE_4$ with probability at least $1 - O(n^{-98})$, we have
\begin{align*}
    n \max_{i,j \in [n]} \abs{\sbr{\bar{\bK}_n}_{ij} - \sbr{\bK_n}_{ij}} &\lesssim \frac{\polylog{n}}{\sqrt{n}}
\end{align*}
Finally, note that $\abs{\tr A} \leq n \norm{A}_{\max}$ and $\Fnorm{A} \leq n \norm{A}_{\max}$ for any $A \in \RR^{n \times n}$ ($\norm{A}_{\max}:= \max_{ij} \abs{A_{ij}}$).
We complete the proof.

\subsection{Proof of Lemma \ref{lem: approx_bl_hl}}
\label{sec: prf_approx_bl_hl}
It suffices to show that
\begin{align*}
    \abs{\tr \bar{\bK}_n - \tr \hbK_n} &\lesssim \frac{\polylog{n}}{\sqrt{n}}\\
    \Fnorm{\bar{\bK}_n - \hbK_n} &\lesssim \frac{\polylog{n}}{\sqrt{n}}
\end{align*}
Similar to the proof of Lemma \ref{lem: approx_tl_bl}, we first prove entry-wise concentration, and then union bound over all entries. The above two inequalities then follow from the relations $\abs{\tr A} \leq n \norm{A}_{\max}$ and $\Fnorm{A} \leq n \norm{A}_{\max}$ for any $A \in \RR^{n \times n}$ ($\norm{A}_{\max}:= \max_{ij} \abs{A_{ij}}$).

To get entry-wise concentration, note that Lemma \ref{lem: tau} implies that there exists an event $\cE_{\ref{lem: approx_bl_hl}, 1}$ with probability at least $1 - O(n^{-100})$ under which for any $i \in [n]$,
\begin{align*}
    &\abs{\inner{\htau(x,q;X_i)}{\htau(x',q';X_i)} - \inner{\tau(x,q;X_i)}{\tau(x',q';X_i)}}\\
    \lesssim  &\frac{\polylog{n}}{\sqrt{n}} \sbr{\rbr{1 \vee \norm{x}} \rbr{1 \vee \Onorm{q} \vee \Onorm{q^{-1}}} \rbr{1 \vee \norm{x'}} \rbr{1 \vee \Onorm{q'} \vee \Onorm{q'^{-1}}} }^{C_2}
\end{align*}
Hence under the event $\cE_{\ref{lem: approx_bl_hl}, 1}$, we have
\begin{align*}
    &n \abs{\sbr{\hbK_n}_{ij} - \sbr{\bar{\bK}_n}_{ij}} =\frac{1}{n} \sum_{k=1}^n \abs{\inner{\htau(X_i,Q_i;X_k)}{\htau(X_j,Q_j;X_k)} - \inner{\tau(X_i,Q_i;X_k)}{\tau(X_j,Q_j;X_k)}}\\
    &\lesssim \frac{\polylog{n}}{\sqrt{n}} \sbr{\rbr{1 \vee \norm{X_i}} \rbr{1 \vee \Onorm{Q_i} \vee \Onorm{Q_i^{-1}}} \rbr{1 \vee \norm{X_j}} \rbr{1 \vee \Onorm{Q_j} \vee \Onorm{Q_j^{-1}}} }^{C_2}
\end{align*}
Hence, Condition \ref{assumption: X} and \ref{assumption: bdd_Q} imply that for any $i,j \in [n]$, there exists event $\cE_{\ref{lem: approx_bl_hl}, ij} \subset \cE_{\ref{lem: approx_bl_hl}, 1}$ with probability $\PP(\cE_{\ref{lem: approx_bl_hl}, ij}) \geq 1 - O(n^{-100})$ under which $n \abs{\sbr{\hbK_n}_{ij} - \sbr{\bar{\bK}_n}_{ij}} \lesssim \polylog{n} n^{-1/2}$ (the constant in $\lesssim$ is independent of $i,j$). 

Taking union bound over all $i,j \in [n]$, we conclude that there exists event $\cE_{\ref{lem: approx_bl_hl}, 2} = \bigcap_{i,j \in [n]} \cE_{\ref{lem: approx_bl_hl}, ij}$ with probability at least $1 - O(n^{-98})$ under which
\begin{align*}
    n \max_{i,j \in [n]} \abs{\sbr{\hbK_n}_{ij} - \sbr{\bar{\bK}_n}_{ij}} &\lesssim \frac{\polylog{n}}{\sqrt{n}}
\end{align*}

The proof is then complete.

%% file: prf_power.tex
Lemma \ref{lem: xu25} and Condition \ref{assumption: minimizer_local} imply that there exists an event $\cE_{1,n}$ that satifies $\PP(\cE_{1,n}^c) = O(n^{-100})$ such that for any $k \in \sbr{n}$,
\begin{align*}
    \lambda_{\min}\sbr{-\frac{1}{n} \sum_{i=1}^{n} w_{n}(\hX_k,X_i) dT^{Q_i}_{\hQ_{n}(\hX_k)}} \gtrsim \frac{1}{\polylog{n}}
\end{align*}
Under $\cE_{1,n}$, we have
\begin{align*}
    \cT_n \geq \frac{1}{\polylog{n}} \sum_{k=1}^{n} \Fnorm{\hQ_{n}(X_k) - \hQ_{n}(\hX_k)}^2
\end{align*}
Note that
\begin{align*}
    \hQ_{n}(X_k) - \hQ_{n}(\hX_k) = Q^*(X_k) - Q^*(\tX_k) + \underbrace{\hQ_{n}(X_k) - Q^*(X_k) + Q^*(\tX_k) - \hQ_{n}(\hX_k)}_{\Delta_k}
\end{align*}
Then we can obtain that
\begin{align*}
    &\sum_{k=1}^{n} \Fnorm{\hQ_{n}(X_k) - \hQ_{n}(\hX_k)}^2\\
    =&\sum_{k=1}^{n} \Fnorm{Q^*(X_k) - Q^*(\tX_k)}^2 + 2 \sum_{k=1}^{n} \inner{Q^*(X_k)-Q^*(\tX_k)}{\Delta_k} + \sum_{k=1}^{n} \Fnorm{\Delta_k}^2\\
    \geq &\sum_{k=1}^{n} \Fnorm{Q^*(X_k) - Q^*(\tX_k)}^2 - 2 \rbr{\sum_{k=1}^{n} \Fnorm{Q^*(X_k)-Q^*(\tX_k)}^2}^{1/2} \rbr{\sum_{k=1}^{n} \Fnorm{\Delta_k}^2}^{1/2} \\
    = & \sbr{\sum_{k=1}^{n} \Fnorm{Q^*(X_k) - Q^*(\tX_k)}^2} \cdot \sbr{1-2 \rbr{\frac{\sum_{k=1}^{n} \Fnorm{\Delta_k}^2}{\sum_{k=1}^{n} \Fnorm{Q^*(X_k) - Q^*(\tX_k)}^2}}^{1/2}}.
\end{align*}
Theorem \ref{thm: est}, Lemma \ref{lem: xu25} and \ref{lem: Q} imply that there exists an event $\cE_{2,n}$ that satisfies $\PP(\cE_{2,n}^c) = O(n^{-100})$ such that
\begin{align*}
    \sum_{k=1}^{n} \Fnorm{\Delta_k}^2 &\lesssim \polylog{n},\\
    \sum_{k=1}^{n} \Fnorm{Q^*(X_k) - Q^*(\tX_k)}^2 &\overset{(i)}{\gtrsim} \frac{n a_n^2}{2}.
\end{align*}
Here (i) follows from a similar proof as Lemma 50 in \citet{XuLi2025}, hence the details are omitted.

Combining the above results, we have that under $\cE_{1,n} \cap \cE_{2,n}$,
\begin{align*}
    \cT_n \gtrsim \frac{n a_n^2}{\polylog{n}}.
\end{align*}

To prove Theorem \ref{thm: power}, we claim in Lemma \ref{lem: qhat_bdd} below that $\hq_{1-\alpha}$ is of order $\polylog{n}$ with high probability whether or not the null hypothesis holds. The proof is deferred to Appendix \ref{sec: prf_qhat_bdd}.
\begin{lemma}
    \label{lem: qhat_bdd}
    Instate the assumptions of Theorem \ref{thm: power}. Fix $\alpha \in (0,1)$. There exists a sequence of positive numbers $\cbr{\tq_{1-\alpha,n}}$ such that 
    \begin{itemize}
        \item $\tq_{1-\alpha,n} \lesssim \polylog{n}$
        \item for each $n \geq 1$, there exists an event $\cE_{3,n}$ that satisfies $\PP(\cE_{3,n}^c) = O(n^{-100})$ and on which we have
        \begin{align*}
            \hq_{1-\alpha} \leq \tq_{1-\alpha,n}.
        \end{align*}
    \end{itemize}
\end{lemma}
Then we can obtain that
\begin{align*}
    \PP \rbr{\cT_n > \hq_{1-\alpha}} &\geq \PP \rbr{ \cbr{\cT_n > \tq_{1-\alpha,n}} \cap \cbr{\tq_{1-\alpha,n} > \hq_{1-\alpha}} \cap \cE_{1,n} \cap \cE_{2,n} \cap \cE_{3,n}}\\
    &\geq \PP \rbr{\cbr{\cT_n > \tq_{1-\alpha,n}} \cap \cE_{1,n} \cap \cE_{2,n}} - \PP \rbr{\sbr{\cbr{\tq_{1-\alpha,n} > \hq_{1-\alpha}} \cap \cE_{3,n}}^c}\\
    &= \PP \rbr{\cbr{\cT_n > \tq_{1-\alpha,n}} \cap \cE_{1,n} \cap \cE_{2,n}} - 1 + \PP \rbr{\cbr{\tq_{1-\alpha,n} > \hq_{1-\alpha}} \cap \cE_{3,n}}.
\end{align*}
Note that $polylog{n} = o\rbr{na_n^2}$, by letting $n\to \infty$, we have
\begin{align*}
    \liminf_{n \to \infty} \PP \rbr{\cT_n > \hq_{1-\alpha}} &\geq \liminf_{n \to \infty} \PP \rbr{\cE_{1,n} \cap \cE_{2,n}} - 1 + \liminf_{n \to \infty} \PP(\cE_{3,n}) = 1.
\end{align*}
The argument above holds for any $\PP \in \mathscr{P}_F$, hence the proof is then complete.

\subsection{Proof of Lemma \ref{lem: qhat_bdd}}
\label{sec: prf_qhat_bdd}
Denote $\hS:= \sum_{j=1}^n \hlambda_j w_j$. Note that since the kernel $\hat{\cK}_n$ is PSD, we have $\lambda_j \geq 0$ for any $j \in [n]$. By defininition, $\hq_{1-\alpha}$ is the $(1-\alpha)$-quantile of $\hS$. Since $\cbr{w_j}_{j \in [n]}$ is independent of $\cbr{(X_j,Q_j)}_{j \in [n]}$, we can obtain that $\EE \sbr{\hS \mid \cbr{(X_j,Q_j)}_{j \in [n]}} = \sum_{j \in [n]} \hlambda_j=\tr \hbK_n$. Therefore, by Markov's inequality, we have
\begin{align*}
    \PP \rbr{ \hS > 2\alpha^{-1} \tr \hbK_n \mid \cbr{(X_j,Q_j)}_{j \in [n]}} \leq \frac{\tr \hbK_n}{2\alpha^{-1}\tr \hbK_n} = \frac{\alpha}{2}.
\end{align*}
Therefore, we can obtain that
\begin{align*}
    \hq_{1-\alpha} &\leq 2 \alpha^{-1} \tr \hbK_n\\
    &=2\alpha^{-1} \frac{1}{n}\sum_{j=1}^{n} \hat{\cK}_n(X_i,Q_i; X_i,Q_i).
\end{align*}
The desired result then follows from Lemma \ref{lem: tau} and Conditions \ref{assumption: X}, \ref{assumption: bdd_Q}.

%% file: prf_example.tex

Recall notation $\cS^d_{++}$, $\cS^d_{+}$

\begin{align*}
    T_S^Q
    = Q^{1/2}\rbr[\big]{Q^{1/2} S Q^{1/2}}^{-1/2}Q^{1/2}
    = S^{-1/2}\rbr[\big]{S^{1/2} Q S^{1/2}}^{1/2}S^{-1/2}.
\end{align*}

\begin{align}
    W^2(S,Q) &= \tr{S} + \tr{Q} - 2 \tr \sbr[\Big]{\rbr[\big]{S^{1/2}QS^{1/2}}^{1/2}}, \quad \text{for any } S,Q \in \cS^d_{+}\\
    d_S W^2(S,Q) &= I_d - T_S^Q \in \cS^d.
\end{align}

\subsection{Example \ref{example_c} (commutative case)}

Without loss of generality, we can assume $U=I_d$. Denote $\phi_k(x)=a_k+ \beta^\top x$, write $Q^*_x:= Q^*(x)$ and $F_x(S):=F(x,S)$. Then we have
\begin{alignat*}{2}
    Q &= \diag(Q_{11},\ldots,Q_{dd}), &Q_{kk}&=\phi_k(X)^2V_k^2, \\
    Q^*_x &= \diag (Q^*_{x,11}, \ldots, Q^*_{x,dd}) &\qquad\qquad   Q^*_{x,kk}&= \phi_k(x)^2, \\
    F_x(S)&:=\mathbb E\bigl[w(x,X)W^2(S,Q)\bigr],
    &\qquad\qquad 
    w(x,X)&=1+3x^\top X.
\end{alignat*}

Note that the weight function $w(x,X)$ satisfies the reproducing perperty for affine functions, i.e. for any affine function $\psi$ and any $x \in \supp(X)$, we have
\begin{equation}
    \EE w(x,X)=1, \qquad \EE w(x,X)\psi(X) = \psi(x). \label{eqn: prf_ex1_w}
\end{equation}


\noindent\textbf{Assumption~\ref{assumption: X}:} 
Under the setting of Example~\ref{example_c}, $\supp(X) = [-1,1]^p$ and $\cov(X) = \tfrac{1}{3}I_p$, hence Assumption~\ref{assumption: X} is satisfied.

\noindent\textbf{Assumption~\ref{assumption: bdd_Q}:} 
For $p_y=p_z=2$ and $p=4$, $\delta_z \in [-1/4,1/4]$. Let $\ell_x := \ell(x):=\beta^\top x$. Then we have
\begin{align*}
    L_0:= \sup_{x \in [-1,1]^p} \abs{\ell(x)} = 0.1p_y + \abs{\delta_z}p_z  = 0.2 + \abs{\delta_z}2 \leq 0.7.
\end{align*}
Let $M_-:=a_1-L_0$, $M_+:=a_d+L_0$, $V_{\min}=0.9,\ V_{\max}=1.1$, 
$\underline{\lambda}:=V_{\min}^2\,M_-^2$, $\overline{\lambda}:=V_{\max}^2\,M_+^2$; the design gives $M_->0$, so
$\phi_k(x)\in[M_-,M_+]$ and $Q_{kk},\phi_k(x)^2\in[\underline{\lambda},\overline{\lambda}]$ a.s. Thus, Assumption~\ref{assumption: bdd_Q} is satisfied with $C_{\Lambda}=0$ and $c_{\lambda} = \max \{\overline{\lambda}, \underline{\lambda}^{-1}\}$.

\noindent\textbf{Assumption~\ref{assumption: minimizer_local} (second derivative):} To apply Lemma~\ref{lem: Q}, note that for diagonal matrices $S = \diag(s_1,\ldots,s_d)$ and $Q = \diag(q_1,\ldots,q_d)$, we have $Q^{1/2} S Q^{1/2} = \diag(s_1 q_1, \ldots, s_d q_d)$. By taking $U=I_d$ and $\lambda_i = q_i s_i$, for any perturbation $A \in \cS^d$,
\begin{align*}
    \Delta = Q^{1/2} A Q^{1/2}, \qquad \Delta_{kl}= \sqrt{q_k q_l} A_{kl}  \text{ for } k,l \in [d].
\end{align*}
Then Lemma~\ref{lem: Q} implies that 
\begin{align*}
    dT_S^Q(A) = -Q^{1/2} \Lambda^{-1/2} \delta \Lambda^{-1/2} Q^{1/2}, \qquad \text{where } \delta_{kl} = \frac{\sqrt{q_k q_l} A_{kl}}{\sqrt{q_k s_k} + \sqrt{q_l s_l}},
\end{align*}
which implies that
\begin{align*}
    \sbr[\big]{dT_S^Q(A)}_{kl} = \Finner{-Q^{1/2} \Lambda^{-1/2} \delta \Lambda^{-1/2} Q^{1/2}}{E_{kl}} = -\frac{\sqrt{q_k q_l}}{\sqrt{q_ks_k q_ls_l}}\delta_{kl} = -\frac{\sqrt{q_k q_l} }{\sqrt{s_k s_l}(\sqrt{q_k s_k} + \sqrt{q_l s_l})}A_{kl}.
\end{align*}
Hence $dT_S^Q$ is diagonal with respect to the symmetric coordinate basis $\{E_{kl}\}_{1 \leq k \leq l \leq d}$ ($E_{kl}=(e_k e_l^\top + e_l e_k^\top)/\sqrt{2}$ for $k<l$, $E_{kk}=e_ke_k^\top$).
Thus $H(x):= d^2 F(x,Q^*(x))=-\EE [w(x,X)dT_{Q^*(x)}^Q]$ is diagonal in the same basis with corresponding diagonal entries $H_{kl}(x)$. To verify Assumption~\ref{assumption: minimizer_local}, we consider the diagonal and off-diagonal entries separately.

\noindent\textit{Diagonal $k=l$:} at minimizer $S = Q_x^*$, plugging in $s_k = \phi_k(x)^2$ and $q_k = \phi_k(X)^2 V_k^2$, we can obtain that
\begin{align*}
    H_{kk}(x) &= \EE w(x,X) \frac{\sqrt{q_k}}{2s_k^{3/2}} = \EE w(x,X) \frac{\phi_k(X) V_k}{2 \phi_k(x)^3} = \frac{1}{2 \phi_k(x)^2}
\end{align*}
Thus
\begin{align*}
    H_{kk}(x) >  \frac{1}{2(a_d + L_0)^2}, \quad \forall x \in [-1,1]^p, k \in [d].
\end{align*}

\noindent\textit{Off-diagonal $k<l$:} at minimizer $S = Q_x^*$, plugging in $s_k = \phi_k(x)^2$ and $q_k = \phi_k(X)^2 V_k^2$, we have
\begin{align*}
    H_{kl}(x) = \EE w(x,X) \rho_{kl}(\ell_x,\ell(X),;V_k, V_l)
\end{align*}
where
\begin{align*}
    \rho_{kl}(\ell_x,u;V_k, V_l) = \frac{(a_k+u)(a_l+u)V_k V_l}{(a_k+\ell_x)(a_l+\ell_x)  D}, \quad D:=\sbr{(a_k+\ell_x)(a_k+u)V_k + (a_l+\ell_x)(a_l+u)V_l}.
\end{align*}

\begin{lemma} \label{lem: prf_ex1_A5_offdiag}
    we have
    \begin{align*}
        \inf \cbr{H_{kl}(x): x \in [-1,1]^p, k,l \in [d], k<l} > 0.
    \end{align*}
\end{lemma}
\begin{proof}
    See Section~\ref{sec: prf_ex1_A5_offdiag}.
\end{proof}
Combining the diagonal and off-diagonal cases, we have
\begin{align*}
    \inf_{x\in [-1,1]^p} \lambda_{\min}(H(x)) = \inf_{x\in [-1,1]^p} \min_{k\leq l} H_{kl}(x) =:\kappa >0.
\end{align*}
Thus, Assumption~\ref{assumption: minimizer_local} is verified with $c_{\lambda}=\kappa^{-1}$ and $C_{\lambda} =0$.

\noindent\textbf{Assumption~\ref{assumption: frechet} \& \ref{assumption: minimizer_global} (unique global minimizer):} Note that the Wasserstein distance is well-defined on $\cS^d_+$ (not just $\cS^d_{++}$). We define excess $\cE_x:\cS^d_+ \to \RR$ and $\Delta: \cS^d_+ \times \cS^d_+ \to \RR$ by
\begin{align*}
    \cE_x(S) &:= F_x(S) - F_x(Q_x^*),\\
    \Delta(S,Q) &:= \sbr[\Big]{\sum_{k=1}^d \sqrt{S_{kk} Q_{kk}}} - \tr \sbr[\big]{(S^{1/2} Q S^{1/2})^{1/2}}
\end{align*}
Under the setting of Example~\ref{example_c}, response $Q$ is diagonal
\begin{lemma} \label{lem: prf_ex1_uniq}
    Under the setting of Example~\ref{example_c}, suppose the strict slope condition
    \begin{equation}
        \sqrt p\,\norm{\beta} < c_1:=\frac{a_1(a_1-L_0)}{a_1+L_0}
        \label{eqn: prf_ex1_uniq_slope}
    \end{equation}
    holds. Then for every $x\in\supp(X)$,
    \[
    \cE_x(S)\ \ge\ 0\quad\text{for all }S\in\cS_+^d,\qquad\text{with equality iff }S=Q_x^{*}.
    \]
    In particular $Q_x^{*}$ is the unique global minimizer of $F_x$ over $\cS_+^d$; hence the
    Fr\'echet regression model holds and Assumption~\ref{assumption: frechet} and the first clause of Assumption~\ref{assumption: minimizer_global}
    are satisfied.
\end{lemma}
\begin{proof}
    See Appendix~\ref{sec: prf_ex1_uniq}.
\end{proof}
Condition \eqref{eqn: prf_ex1_uniq_slope} holds throughout the stated range
$\delta_z\in(-\tfrac14,\tfrac14)$. Indeed, writing $\delta:=|\delta_z|$, we have
$L_0(\delta)=0.2+2\delta$ and $\lVert\beta\rVert(\delta)=(0.02+2\delta^2)^{1/2}$, so
$\delta\mapsto\sqrt p\,\lVert\beta\rVert(\delta)$ is increasing while
$\delta\mapsto c_1(\delta)=a_1(a_1-L_0(\delta))/(a_1+L_0(\delta))$ is decreasing (as $L_0$
increases and $t\mapsto a_1(a_1-t)/(a_1+t)$ is decreasing). The inequality is therefore
binding at the endpoint $\delta=\tfrac14$, where $\sqrt p\,\lVert\beta\rVert=0.762<0.963=c_1$;
hence it holds for every $\delta\in[0,\tfrac14)$. Thus, the unique global minimizer condition is verified.

\noindent\textbf{Assumption~\ref{assumption: minimizer_global} (well-separated minimizer):}

\begin{lemma}\label{lem: sep}
    Instate \eqref{eqn: prf_ex1_uniq_slope} and $\kappa>0$. There is $\delta_F>0$ such that for every
    $x\in\supp(X)$ and all $0\le\delta\le\delta_F\le\Delta$,
    \[
    \inf_{\delta\le\Fnorm{S-Q_x^{*}}\le\Delta}\cE_x(S)\ \ge\ \frac{\kappa}{4}\,\delta^2 .
    \]
    Thus Assumption~\ref{assumption: minimizer_global} holds with $\alpha_F=2$ and $\gamma_F\equiv4/\kappa$ (so $c_F=4/\kappa$, $C_F=0$).
\end{lemma}

\begin{proof}
    See Appendix~\ref{sec: prf_sep}.
\end{proof}

\subsection{Proof of Lemma~\ref{lem: prf_ex1_A5_offdiag}}
\label{sec: prf_ex1_A5_offdiag}
First-order expansion of $u\mapsto \rho_{kl}(\ell_x,u;V_k,V_l)$ about $u=\ell_x$ defines the  residual
\begin{align} \label{eqn: prf_ex1_taylor}
    R_{kl}(X):=\rho_{kl}(\ell_x,\ell(X);V_k, V_l)-\rho_{kl}(\ell_x,\ell_x;V_k,V_l)-\partial_u\rho_{kl}(\ell_x,\ell_x;V_k,V_l)\,(\ell(X)-\ell_x). 
\end{align}
Multiplying this identity by $w(x,X)$ and taking expectations, the independence $V\perp X$ together with \eqref{eqn: prf_ex1_w} gives
\begin{align} \label{eqn: prf_ex1_split}
    H_{kl}(x)=&\underbrace{\EE_V \rho_{kl}(\ell_x,\ell_x;V_k,V_l)}_{=:\ \bar\rho_{kl}(x)}
    \;+\;\underbrace{\EE_V\big[\partial_u\rho_{kl}(\ell_x,\ell_x;V_k,V_l)\big]\,\EE \big[w(x,X)(\ell(X){-}\ell_x)\big]}_{=\,0\ \text{by \eqref{eqn: prf_ex1_w}}} \nonumber\\
    &\;+\; \underbrace{\EE \big[w(x,X)R_{kl}(X)\big]}_{=:\ r_{kl}(x)}.
\end{align}
Then it suffices to bound the leading term $\bar\rho_{kl}(x)$ and the remainder $r_{kl}(x)$:
\begin{itemize}[leftmargin=2em]
    \item Lower bound for the leading term $\bar\rho_{kl}(x)$: At $u=\ell_x$ the map collapses to $\rho_{kl}(\ell_x,\ell_x;V_k,V_l)=V_k V_l\big/\big((a_k+\ell_x)^2V_k+(a_l+\ell_x)^2V_l\big)$, so
    \begin{equation} \label{eqn: prf_ex1_leading}
        \bar\rho_{kl}(x)=\EE_V \frac{V_kV_l}{(a_k+\ell_x)^2V_k+(a_l+\ell_x)^2V_l}
        \ \ge\ \underline{\rho}_{kl}:=\frac{V_{\min}^2}{V_{\max}\big[(a_k+L_0)^2+(a_l+L_0)^2\big]}>0 .
    \end{equation}
    \item Upper bound for the remainder $r_{kl}(x)$: Computation of the second derivative gives
    \begin{align*}
        \abs{\partial_u^2 \rho_{kl}(\ell_x,u;V_k,V_l)} = \frac{2V_k^2V_l^2\,(a_l-a_k)^2}{D^3}
    \end{align*}
    Using $D\ge V_{\min}\big[(a_k+\ell_x)(a_k+u)+(a_l+\ell_x)(a_l+u)\big]\ge V_{\min}(\underline a_k^2+\underline a_l^2)$, we have
    \begin{align*}
        \abs{\partial_u^2 \rho_{kl}(\ell_x,u;V_k,V_l)} \leq \frac{2V_{\max}^4\,(a_l-a_k)^2}{V_{\min}^3\,(\underline a_k^2+\underline a_l^2)^3} =: B_{kl}, \quad \forall \ell_x, u \in [-L_0,L_0], V_k, V_l \in [V_{\min},V_{\max}].
    \end{align*}
    Then the remainder $R_{kl}(X)$ from \eqref{eqn: prf_ex1_taylor} can be bounded by
    \begin{align*}
        \abs{R_{kl}(X)} \le\tfrac12 B_{kl}(\ell(X)-\ell_x)^2 .
    \end{align*}
    Using $(\ell(X)-\ell_x)^2\le 2L_0|\ell(X)-\ell_x|$ and Cauchy--Schwarz, we have that 
    \begin{equation} \label{eqn: prf_ex1_remainder}
        |r_{kl}(x)|\le B_{kl}L_0\,(\EE w^2)^{1/2}\big(\EE(\ell(X)-\ell_x)^2\big)^{1/2}
        \le B_{kl}\,L_0\,\frac{(1+3p)\,\|\beta\|}{\sqrt3},
    \end{equation}
    uniformly in $x$. Here we used
    \begin{align*}
        \EE w(x,X)^2 = \EE \sbr{1 + 6 x^\top X + 9 (x^\top X)^2} = 1 + 3 \norm{x}^2 \leq 1 + 3p
    \end{align*}
    and bias-variance decomposition
    \begin{align*}
        \EE (\ell(X)-\ell_x)^2 = \sbr{\Var (\ell(X)-\ell_x)}^2 + [\EE \ell(X) - \ell_x]^2 = \frac{1}{3} \norm{\beta}^2 + (\beta^\top x)^2 \leq \rbr{\frac{1}{3}+p}\norm{\beta}^2.
    \end{align*}
\end{itemize}
Combining the bounds from \eqref{eqn: prf_ex1_leading} and \eqref{eqn: prf_ex1_remainder} gives
\begin{align*}
    \frac{\abs{r_{kl}(x)}}{\bar\rho_{kl}(x)} &\leq   \frac{(1+3p)\,\|\beta\|}{\sqrt3} L_0 \cdot \frac{B_{kl}}{\underline{\rho}_{kl}}\\
    &\leq \frac{(1+3p)\,\|\beta\|}{\sqrt3} L_0\cdot \frac{2 V_{\max}^5}{V_{\min}^5} g(a_k,a_l),\\
    \text{where} \quad & g(a_k,a_l):=\frac{(a_l-a_k)^2\big[(a_k+L_0)^2+(a_l+L_0)^2\big]}{\big[(a_k-L_0)^2+(a_l-L_0)^2\big]^3}.
\end{align*}
Then it suffices to show the above ratio is $< 1$ for any $1\leq k<l\leq d$.
We show in Claim~\ref{claim: prf_ex1_A5_offdiag} below that $\sup_{k<l} g(a_k,a_l) \leq g(a_1,a_4)=g(2,3.5)=0.0648$ for all $k,l$. Then using $\norm{\beta}\leq \sqrt{0.02+2\delta_z^2}\leq \sqrt{0.145}$, we have
\begin{align*}
    \frac{(1+3p)\norm{\beta}}{\sqrt{3}} L_0\cdot \frac{2 V_{\max}^5}{V_{\min}^5} g(a_k,a_l) &\leq \frac{13\sqrt{0.145}}{\sqrt{3}}0.7 \cdot 2\frac{1.1^5}{0.9^5}0.0648 = 0.708 < 1.
\end{align*}
The proof is then complete.

\begin{claim}[\ref{claim: prf_ex1_A5_offdiag}] \label{claim: prf_ex1_A5_offdiag}
    Under the parameter setting in Example~\ref{example_c}, we have
    \begin{align*}
        \sup_{1\leq k<l\leq d} g(a_k,a_l) = g(a_1,a_4) = g(2,3.5) = 0.0648.
    \end{align*}
\end{claim}
\begin{proof}[of Claim~\ref{claim: prf_ex1_A5_offdiag}]
Recall $L_0=0.7$ and $a_k=1.5+k/2$, so $a_1=2$, $a_2=2.5$, and $a_k-L_0\ge a_1-L_0=1.3>0$.
Two elementary reductions bound $g$ by a function of $a_l$ alone. For any $k<l$,
\[
(a_l-a_k)^2\le(a_l-a_r)^2,\quad (a_k+L_0)^2+(a_l+L_0)^2\le 2(a_l+L_0)^2,\quad
(a_k-L_0)^2\ge(a_r-L_0)^2\quad(a_k\ge a_r),
\]
the middle inequality because $a_k\le a_l$ and the last because $t\mapsto(t-L_0)^2$ increases on
$t\ge L_0$. Taking $r=1$ (valid for every $k\ge1$) and $r=2$ (valid for $k\ge2$) gives, respectively,
\[
g(a_k,a_l)\le\psi(a_l):=\frac{2(a_l-2)^2(a_l+L_0)^2}{\big[(2-L_0)^2+(a_l-L_0)^2\big]^3},
\]
\[
g(a_k,a_l)\le\chi(a_l):=\frac{2(a_l-2.5)^2(a_l+L_0)^2}{\big[(2.5-L_0)^2+(a_l-L_0)^2\big]^3}\qquad(k\ge2).
\]
We treat three exhaustive cases; because $a_l$ moves in steps of $0.5$, no grid value lies in $(5,5.5)$,
so $\{l\le7\}\cup\{l\ge8\}$ partitions row $k=1$.

\emph{Case $k\ge2$.} We show $\chi<0.0648$ on $[3,\infty)$ (the range of $a_l$ when $k\ge2$). Substituting
$u=a_l-L_0\ge2.3$, the sign of $(\log\chi)'$ equals that of the cubic
$q(u)=-2u^3+1.6u^2+28.08u-2.592$. Since $q'(u)=-6u^2+3.2u+28.08$ vanishes only at $u\approx2.447$ and
$q'(2.3)=3.70>0$, $q$ rises on $[2.3,2.447]$ then falls; as $q(2.3)=46.1>0$ and $q\to-\infty$, $q$ has a
single sign change on $[2.3,\infty)$, at $u^\ast\approx4.13$. Hence $\chi$ increases then decreases, with a
unique maximum at $a_l=u^\ast+L_0\approx4.83$; direct evaluation gives $\chi(4.83)=0.0398<0.0648$
(and $\chi(3)=0.011$, $\chi\to0$). Thus every pair with $k\ge2$ has $g<0.0648$.

\emph{Case $k=1$, $l\ge8$ (so $a_l\ge5.5$).} We show $\psi$ is strictly decreasing there. With $u=a_l-L_0$,
\[
(\log\psi)'\ \propto\ -\,C(u),\qquad C(u):=2u^3+0.4u^2-17.68u-0.338
\]
(the proportionality constant is positive). Now $C(4.8)=145.2>0$ and $C'(u)=6u^2+0.8u-17.68$ satisfies
$C'(4.8)=124.4>0$ with $C''(u)=12u+0.8>0$, so $C'>0$ and hence $C$ is increasing on $[4.8,\infty)$;
therefore $C(u)\ge C(4.8)>0$ and $\psi'<0$ for $a_l\ge5.5$. Consequently $\psi(a_l)\le\psi(5.5)=0.0623
<0.0648$ for all $l\ge8$, so these pairs are strictly below the target.

\emph{Case $k=1$, $l\le7$.} Here $\psi$ is too loose and we evaluate $g$ exactly. For $a_l\in\{2.5,3,3.5,4,4.5,5\}$,
\[
g(2,a_l)=(0.0366,\ 0.0617,\ 0.0648,\ 0.0590,\ 0.0511,\ 0.0436),
\]
each an exact rational (e.g.\ $g(2,3.5)=56092500/865523177=0.0648$), with a strict maximum at $a_l=3.5$,
i.e.\ $(k,l)=(1,4)$.

Combining the three cases, every pair other than $(1,4)$ satisfies $g<0.0648$, while $g(2,3.5)=0.0648$;
the supremum is thus attained uniquely at $(1,4)$, and since the cases cover all $l$ it is a supremum over
the infinite grid, independent of $d$.
\end{proof}

\subsection{Proof of Lemma~\ref{lem: prf_ex1_uniq}}
\label{sec: prf_ex1_uniq}

\begin{claim}[\ref{claim: prf_ex1_uniq_excess_identity}] \label{claim: prf_ex1_uniq_excess_identity}
    Under the parameter setting in Example~\ref{example_c}, for any $x\in \supp(X)$ and $S \in \cS^d_+$,
    \begin{equation}
        \cE_x(S)=\underbrace{\sum_{k=1}^d\big(\sqrt{S_{kk}}-\phi_k(x)\big)^2}_{=:\,G_{\mathrm{diag}}(x,S)}
        \;+\;\underbrace{2\,\EE\!\big[w(x,X)\,\Delta(S,Q)\big]}_{=:\,G_{\mathrm{off}}(x,S)},
        \label{eqn: excess}
    \end{equation}
    and both terms are nonnegative and jointly continuous on $\supp(X)\times\cS_+^d$.
\end{claim}

\begin{proof}[of Claim~\ref{claim: prf_ex1_uniq_excess_identity}]
    Since $W^2(S,Q) = \tr S+\tr Q -2\tr [(Q^{1/2}SQ^{1/2})^{1/2}]$, for any $S \in \cS^d_+$, we have
    \begin{align*}
        F_x(S) = \tr S - 2 \EE w(x,X)\tr\sbr[\big]{(Q^{1/2}SQ^{1/2})^{1/2}} + \EE w(x,X) \tr Q
    \end{align*}
    Specifically at $S=Q_x^*$, since $\EE w(x,X)\sbr[\big]{(Q_x^{*1/2}Q Q_x^{*1/2})^{1/2}}=  Q_x^*$, we can obtain
    \begin{align*}
        F_x(Q_x^*) = -\tr Q_x^* + \EE \tr\sbr{w(x,X) Q}.
    \end{align*}
    Then, let $D(A):=\diag(A_{11},\ldots,A_{dd})$, $D^{1/2}(A):= \diag(A_{11}^{1/2},\ldots,A_{dd}^{1/2})$ and take difference between $F_x(S)$ and $F_x(Q_x^*)$ to get
    \begin{align}
        \cE_x(S) &= F_x(S) - F_x(Q_x^*) \nonumber\\ 
        &= \tr S + \tr Q_x^* - 2 \EE w(x,X)\tr\sbr[\big]{(Q^{1/2}SQ^{1/2})^{1/2}} \nonumber\\
        &= \Fnorm[\big]{D^{1/2}(S) - D^{1/2} (Q_x^*)}^2 + 2 \tr D \rbr[\big]{\rbr[big]{Q^{*1/2} S Q^{*1/2}}^{1/2}} - 2 \EE w(x,X)\tr\sbr[\big]{\rbr[\big]{Q^{1/2}SQ^{1/2}}^{1/2}} \nonumber\\
        &= \Fnorm[\big]{D^{1/2}(S) - D^{1/2} (Q_x^*)}^2 + 2 \EE w(x,W) \underbrace{\sbr[\big]{\tr D \rbr{\rbr[big]{Q^{1/2} S Q^{1/2}}^{1/2}} - \tr \rbr[big]{Q^{1/2} S Q^{1/2}}^{1/2}}}_{= \ \Delta(S,Q)} \nonumber
    \end{align}
    Here the last equality uses the identity $\tr D (\rbr[big]{Q^{*1/2} S Q^{*1/2}}^{1/2}) = \EE w(x,X) \tr D \rbr[\big]{\rbr[big]{Q^{1/2} S Q^{1/2}}^{1/2}}$.
    The proof of Claim~\ref{claim: prf_ex1_uniq_excess_identity} is then complete.
\end{proof}

Clearly, the first term $G_{\diag}(x,S)$ in~\eqref{eqn: excess} is $\geq 0$, with equality if and only if
\[
    S_{kk}=Q_x^* = \phi_k(x)^2, \qquad k=1,\ldots,d.
\]
It remains to prove the 2nd term $G_{\mathrm{off}}(x,S)$ in~\eqref{eqn: excess} is $\geq 0$, with equality if and only if $S$ is diagonal.

\begin{claim}[\ref{claim: prf_ex1_uniq_Delta}] \label{claim: prf_ex1_uniq_Delta}
    For $S\in\cS_+^d$ and diagonal $Q\succeq0$, $\Delta(S,Q)\ge0$; if in addition $Q\succ0$ then
    $\Delta(S,Q)=0$ iff $S$ is diagonal. Moreover $S\mapsto\Delta(S,Q)$ is continuous on
    $\cS_+^d$.
\end{claim}
\begin{proof}[of Claim~\ref{claim: prf_ex1_uniq_Delta}]
    The claim follows from Lemma~\ref{lem: mtrx_sqrt} and the continuity of the matrix square root on~$\cS_+^d$ directly.
\end{proof}

\begin{claim}[\ref{lem: G_S}] \label{lem: G_S}
    For any fixed $S \succeq 0$, define $G_S: \RR^d_{++} \to \RR$ by
    \begin{align*}
        G_S(q) = \tr D \sbr[\big]{(\diag(q) S \diag(q))^{1/2}} - \tr \sbr[\big]{(\diag(q) S \diag(q))^{1/2}}
    \end{align*}
    where $\diag(q)$ denotes the diagonal matrix with entries $q_1,\ldots,q_d > 0$.
    Then
    \begin{enumerate}[label=\textup{(\arabic*)}, leftmargin=*]
        \item $G_S$ is positively homogeneous of degree $1$ in $q$: $G_S(cq) = c G_S(q)$ for $c > 0$.
        \item $G_S(q) \geq 0$ with equality if and only if $S$ is diagonal.
        \item $G_S$ is coordinatewise nondecreasing in $q$.
    \end{enumerate}
    Moreover, let
    \begin{align*}
        \sG_S(t) := \EE_V G_S(q(t,V)), \qquad q_k(t,V) = (a_k + t)V_k, k \in [d], \qquad \forall t \in [-T_0,T_0].
    \end{align*}
    If $a_d \geq a_{d-1} \geq \cdots \geq a_1 > T_0 >0$, then 
    \begin{enumerate}[label=\textup{(\arabic*)}, start=4, leftmargin=*]
        \item for any $-T_0 \leq s \leq t \leq T_0$,
        \begin{align*}
            \cG_S(t) \leq \frac{a_1 + t}{a_1 + s} \cG_S(s).
        \end{align*}
    \end{enumerate}
    
\end{claim}
See Appendix~\ref{sec: prf_G_S} for the proof.

Note that $\Delta(S,Q) = G_S(q(\beta^\top X,V))$ and $w(x,X)=1+3x^\top X$, then
\begin{align}
    \EE w(x,X) \Delta(S,Q) &= \EE \Delta(S,Q) + 3\EE (x^\top X) \Delta(S,Q) \nonumber\\
    &= \EE \sG_S(\beta^\top X) + 3 \EE(x^\top X) \sG_S(\beta^\top X) \label{eqn: prf_ex1_assump4_decomp}.
\end{align}
We bound the two terms in~\eqref{eqn: prf_ex1_assump4_decomp} separately. To apply Claim~\ref{lem: G_S}, we let $T_0 = \sup \cbr{\beta^\top x: x \in [-1,1]^p}$.
\begin{itemize}[leftmargin=*]
    \item First term: apply Claim~\ref{lem: G_S} (4) with $s=\beta^\top X$ and $t = L_0$ gives
    \begin{align*}
        \sG_S(\beta^\top X) \geq \frac{a_1 + \beta^\top X}{a_1 + L_0} \sG_S(L_0)
    \end{align*}
    which implies
    \begin{align}
        \EE \sG_S(\beta^\top X) \geq \frac{a_1}{a_1+L_0} \sG_S(L_0) \label{eqn: prf_ex1_assump4_1st_term}
    \end{align}    
    \item Second term: since $\EE \beta^\top X= 0$, we have
    \begin{align*}
        \EE(x^\top X) \sG_S(\beta^\top X) = \EE (x^\top X) \sbr{\sG_S(\beta^\top X) - \sG_S(0)}.
    \end{align*}
    Note that We have
    \begin{align*}
        \abs{\sG_S(t) - \sG_S(0)} \leq \frac{\abs{t}}{a_1- L_0} \sG_S(L_0), \qquad \abs{t} \leq L_0
    \end{align*}
    To see this, we consider $t\geq0$ and $t<0$ separately:
    \begin{itemize}
        \item If $t\geq 0$, Claim~\ref{lem: G_S} gives $\sG_S(r) \leq \frac{a_1+r}{a_1} \sG_S(0)$, which implies
        \begin{align*}
            \sG_S(r) - \sG_S(0) \leq \frac{t}{a_1} \sG_S(0) \leq \frac{t}{a_1-L_0} \sG_S(L_0).
        \end{align*}
        \item If $t <0$, Claim~\ref{lem: G_S} gives $\sG_S(0) \leq \frac{a_1}{a_1+t} \sG_S(t)$, which implies
        \begin{align*}
            \sG_S(0) - \sG_S(t) \leq \frac{-t}{a_1+t} \sG_S(0) \leq \frac{-t}{a_1-L_0} \sG_S(L_0).
        \end{align*}
    \end{itemize}
    Therefore, using $\Cov(X)=I_d/3$ gives
    \begin{align}
        \abs{3 \EE (x^\top X) \sG_S(\beta^\top X)} &\leq \frac{3 \sG_S(L_0)}{a_1-L_0} \EE \abs{x^\top X} \abs{\beta^\top X} \nonumber\\
        &\leq \frac{3 \sG_S(L_0)}{a_1-L_0} \sqrt{\EE (x^\top X)^2} \sqrt{\EE (\beta^\top X)^2} \nonumber\\
        &\leq \frac{\norm{x} \norm{\beta}}{a_1-L_0} \sG_S(L_0). \label{eqn: prf_ex1_assump4_2nd_term}
    \end{align}
\end{itemize}
Combining \eqref{eqn: prf_ex1_assump4_decomp}, \eqref{eqn: prf_ex1_assump4_1st_term} and~\eqref{eqn: prf_ex1_assump4_2nd_term} gives
\begin{align*}
    \EE w(x,X) \Delta(S,Q) \geq \sG_S(L_0) \sbr{\frac{a_1}{a_1+L_0} - \frac{\norm{x}\norm{\beta}}{a_1-L_0}}.
\end{align*}
Hence given $\frac{a_1}{a_1+L_0} - \frac{\norm{x}\norm{\beta}}{a_1-L_0} >0$ (which holds for all $x \in \supp X$ given parameters in Example~\ref{example_c}), we have $\EE w(x,X) \Delta(S,Q) \geq 0$ with equality if and only if $S$ is diagonal, which combined with \eqref{eqn: excess} shows that $Q_x^*$ is the unique global minimizer of $F_x(S)$ over $\cS^d_{+}$.

\subsection{Proof of Claim~\ref{lem: G_S}}
\label{sec: prf_G_S}

    (1) is obvious. (2) follows from Lemma~\ref{lem: mtrx_sqrt} with $B = \diag(q) S \diag(q)$ and $q_k >0$.

    \textit{Proof of (3):} We first assume $S \succ 0$.
    Since $\nabla_B \tr(B)=\frac{1}{2}B^{-1/2}$ for $B \succ 0$, by letting $B=\diag(q)S\diag(q)$, we can obtain that
    \begin{align*}
        \partial_{q_k} \tr \sbr[\big]{(\diag(q) S \diag(q))^{1/2}}
        &= \frac{1}{2} \Finner{B^{-1/2}}{(E_{kk} S \diag(q) + \diag(q) S E_{kk})}\\
        &=\frac{1}{2} \sbr{ \Finner{B^{-1/2}\diag(q) S}{E_{kk}} + \Finner{S\diag(q) B^{-1/2}}{E_{kk}} }
    \end{align*}
    Here $E_{kk}$ is a $d \times d$ matrix with $1$ at $(k,k)$ entry and $0$ elsewhere.
    Note that $\diag(q) S = B \diag(q)^{-1}$ and $S\diag(q) = \diag(q)^{-1} B$, thus $B^{-1/2}\diag(q)S = B^{1/2} \diag(q)^{-1}$ and $S\diag(q) B^{-1/2} = \diag(q)^{-1} B^{1/2}$. Therefore, we have
    \begin{align*}
        \Finner[\big]{B^{-1/2}\diag(q) S}{E_{kk}} + \Finner[\big]{S\diag(q) B^{-1/2}}{E_{kk}} &= \Finner[\big]{B^{1/2} \diag(q)^{-1}}{E_{kk}} + \Finner[\big]{\diag(q)^{-1} B^{1/2}}{E_{kk}}\\
        &= 2 \frac{[B^{1/2}]_{kk}}{q_k}
    \end{align*}
    which implies
    \begin{align*}
        \partial_{q_k} \tr \rbr[big]{\diag(q) S \diag(q)}^{1/2} = \frac{[B^{1/2}]_{kk}}{q_k}
    \end{align*}
    and
    \begin{align*}
        \partial_{q_k} G_S(q) = S_{kk}^{1/2} - \frac{(B^{1/2})_{kk}}{q_k} =: g_k
    \end{align*}
    Since $B \succeq 0$, Lemma~\ref{lem: mtrx_sqrt} gives $(B^{1/2})_{kk}\leq B^{1/2}_{kk} = q_k S_{kk}^{1/2}$, so $g_k \geq 0$.

    To extend the result to $S \succeq 0$, we exploit continuity. Specifically, for any $S \succeq 0$, and any $q,\tq \in RR_{++}^d$ with $q_k \leq \tq_k$. Define $S_\epsilon = S+ \epsilon I_d$ for any $\epsilon \geq 0$. Then $S_\epsilon \succ 0$ for any $\epsilon >0$, and hence $G_{S_\epsilon}(q) \leq G_{S_\epsilon}(\tq)$ for any $\epsilon >0$. Letting $\epsilon \to 0$ and using the continuity of $G_S(q)$ over $S\in \cS^d_{+}$ gives $G_S(q) \leq G_S(\tq)$.

    \textit{Proof of (4):}
    for any $-T_0 \leq s \leq t \leq T_0$, we have
    \begin{align*}
        q_k(t,V) = \frac{a_k + t}{a_k+s} q_k(s,V) \leq \frac{a_1+t}{a_1+s} q_k(s,V), \qquad k \in [d].
    \end{align*}
    which combined with coordinatewise monotonicity in part (3) and homogeneity in part (1) gives
    \begin{align*}
        G_S(q(t,V)) \leq G_S\rbr{\frac{a_1+t}{a_1+s} q(s,V)} = \frac{a_1+t}{a_1+s} G_S(q(s,V)).
    \end{align*}
    Averaging over $V$ gives the desired result.

\subsection{Proof of Lemma~\ref{lem: sep}}
\label{sec: prf_sep}

Write $r:=\Fnorm{S-Q_x^{*}}$. We use three claims.

\begin{claim}[\ref{cl:local}]\label{cl:local}
    There is $\delta_1>0$, independent of $x$, with $\cE_x(S)\ge\tfrac{\kappa}{4}r^2$ whenever
    $r\le\delta_1$.
\end{claim}
\begin{proof}[of Claim~\ref{cl:local}]
    Let $\delta_{\mathrm{band}}:=\tfrac12M_-^2$ and
    $\cM:=\cS^d(\tfrac12\underline{\lambda},2\overline{\lambda})$. If $r\le\delta_{\mathrm{band}}$, every point
    $S_\theta:=Q_x^{*}+\theta(S-Q_x^{*})$, $\theta\in[0,1]$, obeys
    $\tfrac12\underline{\lambda}\le\lambda_{\min}(S_\theta)\le\lambda_{\max}(S_\theta)<2\overline{\lambda}$, so
    $[Q_x^{*},S]\subset\cM$; as $Q\in\cS^d[\underline{\lambda},\overline{\lambda}]$ a.s., both $Q,S_\theta\in\cS^d(M^{-1},M)$
    with $M:=\max\{2\overline{\lambda},(\tfrac12\underline{\lambda})^{-1}\}$, on which $S\mapsto dT^{Q}_S$ is continuous with
    $\Onorm{dT^{Q}_S}\le\mathrm{poly}(M)$ \citet{XuLi2025}.
    Since $|w(x,X)|\le1+3\sqrt p\,\lVert X\rVert$ is integrable, dominated convergence shows
    that $(x,S)\mapsto d^2F_x(S)=-\EE[w(x,X)\,dT^{Q}_S]$, and hence
    $(x,S)\mapsto\lambda_{\min}(d^2F_x(S))$, is jointly continuous on the compact set
    \[
    \cB:=\big\{(x,S):x\in\supp(X),\ \Fnorm{S-Q_x^{*}}\le\delta_{\mathrm{band}}\big\}.
    \]
    We claim there is $\delta_1\in(0,\delta_{\mathrm{band}}]$, independent of $x$, with
    \begin{equation}
    \lambda_{\min}\big(d^2F_x(S)\big)\ \ge\ \tfrac{\kappa}{2}
    \qquad\text{whenever }\Fnorm{S-Q_x^{*}}\le\delta_1 .
    \label{eq:hessfloor}
    \end{equation}
    Otherwise there are $(x_n,S_n)\in\cB$ with $\Fnorm{S_n-Q_{x_n}^{*}}\to0$ and
    $\lambda_{\min}(d^2F_{x_n}(S_n))<\kappa/2$. By compactness pass to a subsequence
    $(x_n,S_n)\to(\bar x,\bar S)\in\cB$; since $x\mapsto Q_x^{*}$ is continuous (indeed
    polynomial), $\bar S=Q_{\bar x}^{*}$, and joint continuity gives
    $\lambda_{\min}(H_{\bar x})=\lambda_{\min}(d^2F_{\bar x}(\bar S))\le\kappa/2<\kappa$,
    contradicting $\kappa=\inf_x\lambda_{\min}(H_x)$.

    Now fix $x$ and $S$ with $r\le\delta_1$, and write $\Gamma:=S-Q_x^{*}$. As
    $Q_x^{*}\in\cS_{++}^d$ is an interior global minimizer,
    $dF_x(Q_x^{*})=0$, so Taylor's theorem in integral form gives
    \[
    \cE_x(S)=\int_0^1(1-\theta)\,\big\langle\Gamma,\ d^2F_x(S_\theta)[\Gamma]\big\rangle_F\,d\theta
    \ \ge\ \frac{\kappa}{2}\,\Fnorm{\Gamma}^2\int_0^1(1-\theta)\,d\theta
    \ =\ \frac{\kappa}{4}\,r^2,
    \]
    where the inequality uses \eqref{eq:hessfloor} at each $S_\theta$ (which lies in the same
    ball, as $\Fnorm{S_\theta-Q_x^{*}}=\theta r\le\delta_1$).
\end{proof}

\begin{claim}[\ref{cl:coerc} (coercivity)]\label{cl:coerc}
    With $t:=\tr S$, $\cE_x(S)\ge t-c\,t^{1/2}$ for $c:=2M_+\sqrt d+2(d\overline{\lambda})^{1/2}(1+3p)^{1/2}$,
    independent of $x$; hence $\cE_x(S)\ge1$ once $\tr S\ge T_\star:=\big(\tfrac12(c+\sqrt{c^2+4})\big)^2$.
\end{claim}
\begin{proof}[of Claim~\ref{cl:coerc}]
    We bound the two terms of \eqref{eqn: excess} separately, then add.
    Since $\phi_k(x)\le M_+$, Cauchy--Schwarz gives
    $\sum_k\phi_k(x)\sqrt{S_{kk}}\le M_+\sqrt d\,(\sum_kS_{kk})^{1/2}=M_+\sqrt d\,\sqrt t$,
    so, discarding $\sum_k\phi_k(x)^2\ge0$,
    \[
    G_{\mathrm{diag}}(x,S)=t-2\sum_k\phi_k(x)\sqrt{S_{kk}}+\sum_k\phi_k(x)^2
    \ \ge\ t-2M_+\sqrt d\,\sqrt t .
    \]
    For the second term, the subtracted trace in $\Delta$ is nonnegative, so
    $0\le\Delta(S,Q)\le\sum_k\sqrt{S_{kk}Q_{kk}}\le(\tr S\cdot\tr Q)^{1/2}\le(d\overline{\lambda})^{1/2}\sqrt t$
    a.s.\ (Cauchy--Schwarz and $Q\preceq\overline{\lambda} I_d$); since $w$ is signed, this upper bound on
    $\Delta$ together with $\EE|w(x,X)|\le(\EE w^2)^{1/2}=(1+3\lVert x\rVert^2)^{1/2}\le(1+3p)^{1/2}$
    yields
    \[
    G_{\mathrm{off}}(x,S)=2\,\EE[w\,\Delta]\ \ge\ -2\,\EE|w|\cdot(d\overline{\lambda})^{1/2}\sqrt t
    \ \ge\ -2(d\overline{\lambda})^{1/2}(1+3p)^{1/2}\sqrt t .
    \]
    Adding the two bounds via \eqref{eqn: excess} gives $\cE_x(S)\ge t-c\sqrt t$ with
    $c:=2M_+\sqrt d+2(d\overline{\lambda})^{1/2}(1+3p)^{1/2}$, a constant independent of $x$ and of $\Delta$.
    Finally $t-c\sqrt t\ge1$ iff $u^2-cu-1\ge0$ for $u:=\sqrt t$, i.e.\ iff
    $u\ge\tfrac12(c+\sqrt{c^2+4})$, i.e.\ iff $\tr S\ge T_\star$.
\end{proof}

\begin{claim}[\ref{cl:cpt} (compactness floor)]\label{cl:cpt}
    There is $\eta>0$ with $\cE_x(S)\ge\eta$ for every $x$ and every $S$ with $r\ge\delta_1$.
\end{claim}
\begin{proof}[of Claim~\ref{cl:cpt}]
    The set $K:=\{(x,S):x\in\supp(X),\tr S\le T_\star,r\ge\delta_1\}$ is
    compact; on $K$, $\cE_x(S)$ is jointly continuous and strictly
    positive (Lemma~\ref{lem: prf_ex1_uniq}: $S\ne Q_x^{*}$ and $Q_x^{*}$ is the unique minimizer over
    $\cS_+^d$), so it attains a minimum $\eta_1>0$. Put $\eta:=\min(1,\eta_1)$: if
    $\tr S>T_\star$ use Claim~\ref{cl:coerc}, else $(x,S)\in K$.
\end{proof}

\medskip
Combining results above, put $\delta_F:=\min\{\delta_1,(4\eta/\kappa)^{1/2}\}$. For
$\delta\le r\le\Delta$: if $r\le\delta_1$, Claim~\ref{cl:local} gives
$\cE_x(S)\ge\tfrac{\kappa}{4}r^2\ge\tfrac{\kappa}{4}\delta^2$; if $r>\delta_1$,
Claim~\ref{cl:cpt} gives $\cE_x(S)\ge\eta\ge\tfrac{\kappa}{4}\delta_F^2\ge\tfrac{\kappa}{4}\delta^2$.
This holds for any $x \in \supp(X)$ and completes the proof of Lemma~\ref{lem: sep}.

%% file: tech_lemmas.tex
Recall that by definition, $w(x, X)=1+(x-\mu)^{\top} \Sigma^{-1}(X-\mu)$ and $w_{n}(x)= 1 + (x-\bar{X})\hSigma^{-1}(X_i - \bar{X})$. For any vector $x \in \RR^p$, denote $\vec{x}=(1, x^\top)^\top$ and
\begin{align*}
    \Xi &:= \EE \Vec{X}\Vec{X}^\top =
    \begin{pmatrix}
        1 & \mu^\top\\
        \mu & \Sigma +\mu \mu^\top
    \end{pmatrix}\\
    \hXi &:=n^{-1} \sum_{i=1}^{n} \vec{X}_i \vec{X}_i^\top = 
    \begin{pmatrix}
        1 & \hmu^\top\\
        \hmu & \hSigma +\hmu \hmu^\top
    \end{pmatrix}
\end{align*}

\begin{lemma} \label{lem: matrix}
    Given a positive definite covariance matrix $\Sigma \in \cS_p^{++}$,
    \begin{enumerate}[leftmargin=3em,label=\textup{(\arabic*)},ref=Part \textup{(\arabic*)}]
        \item \label{enum: matrix1}
        If 
        $\Sigma=\begin{pmatrix}
            \Sigma_{11} & \Sigma_{12} \\
            \Sigma_{21} & \Sigma_{22}
        \end{pmatrix}$, 
        denote $\Sigma_{22.1} = \Sigma_{22} - \Sigma_{21} \Sigma_{11}^{-1} \Sigma_{12}$. Then
        \begin{align*}
            \Sigma^{-1}
            = 
            \begin{pmatrix}
                \Sigma_{11}^{-1} + \Sigma_{11}^{-1} \Sigma_{12} \Sigma_{22.1}^{-1} \Sigma_{21} \Sigma_{11}^{-1} & -\Sigma_{11}^{-1} \Sigma_{12} \Sigma_{22.1}^{-1} \\
                -\Sigma_{22.1}^{-1} \Sigma_{21} \Sigma_{11}^{-1} & \Sigma_{22.1}^{-1}
            \end{pmatrix}
        \end{align*}
        and
        \begin{align*}
            \begin{pmatrix}
                y^\top & z^\top
            \end{pmatrix}
            \Sigma^{-1}
            \begin{pmatrix}
                Y \\ Z
            \end{pmatrix}
            = y^\top \Sigma_{11}^{-1} Y + (z-\Sigma_{21} \Sigma_{11}^{-1} y)^\top \Sigma_{22.1}^{-1} (Z-\Sigma_{21} \Sigma_{11}^{-1} Y).
        \end{align*}
        \item \label{enum: matrix2}
        The weight function $w$ and $w_n$ satisfy
         \begin{align*}
            w(x,X)&= \vec{x}^\top \Xi^{-1} \vec{X} \\ 
            w_{n}(x,X) &= \vec{x}^\top \hat{\Xi}^{-1} \vec{X} 
        \end{align*}
        \item \label{enum: matrix3} Given $x=(y,z)$, denote
        \begin{align*}
            \tx&=(y, \mu_Z + \Sigma_{21}\Sigma_{11}^{-1}(y-\mu_Y))\\
            \hx&=(y, \bar{Z} + \hSigma_{21}\hSigma_{11}^{-1}(y-\bar{Y}))
        \end{align*}
        then the following holds
        \begin{align*}
            w(\tx,X) &= \tw(y,Y)\\
            w_n(\hx,X) &= \tw_n(y,Y)
        \end{align*}
    \end{enumerate}

\end{lemma}

\begin{proof}
    \ref{enum: matrix1} follows from linear algebra and direct computation. 

    \ref{enum: matrix2} follows by applying Property \ref{enum: matrix1} to $\Xi$ and $\hat{\Xi}$ with $y=1$ and $z=x$. 

    \ref{enum: matrix3} follows by applying Property \ref{enum: matrix1} to $\Sigma$ and $\hSigma$ with $x=(y,z)$. To see this, note that
    \begin{align*}
        w(\tx,X)&= 1 + 
        \begin{pmatrix}
            \rbr{y-\mu_Y}^\top & \rbr{\Sigma_{21}\Sigma_{11}^{-1}(y-\mu_Y)}^\top
        \end{pmatrix}
        \Sigma^{-1}
        \begin{pmatrix}
            Y-\mu_Y \\ Z - \mu_Z
        \end{pmatrix}\\
        &=1 + \rbr{y-\mu_Y}^\top \Sigma_{11}^{-1} (Y-\mu_Y)\\
        &+ \rbr{\Sigma_{21}\Sigma_{11}^{-1}(y-\mu_Y) - \Sigma_{21}\Sigma_{11}^{-1}(y-\mu_Y)}^\top \Sigma_{22.1}^{-1} \rbr{Z - \mu_Z - \Sigma_{21}\Sigma_{11}^{-1}(Y-\mu_Y)}\\
        &= 1 + \rbr{y-\mu_Y}^\top \Sigma_{11}^{-1} (Y-\mu_Y)\\
        &= \tw(y,Y)
    \end{align*}
    Similarly for $w_n(\hx,X)$, we have
    \begin{align*}
        w_n(\hx,X)&= 1 + 
        \begin{pmatrix}
            \rbr{y-\bar{Y}}^\top & \rbr{\hSigma_{21}\hSigma_{11}^{-1}(y-\bar{Y})}^\top
        \end{pmatrix}
        \hSigma^{-1}
        \begin{pmatrix}
            Y-\bar{Y} \\ Z - \bar{Z}
        \end{pmatrix}\\
        &= 1 + \rbr{y-\bar{Y}}^\top \hSigma_{11}^{-1} (Y-\bar{Y})\\
        &+ \rbr{\hSigma_{21} \hSigma_{11}^{-1}(y-\bar{Y}) - \hSigma_{21}\hSigma_{11}^{-1}(y-\bar{Y})}^\top \hSigma_{22.1}^{-1} \rbr{Z - \bar{Z} - \hSigma_{21}\hSigma_{11}^{-1}(Y-\bar{Y})}\\
        &=  1 + \rbr{y-\bar{Y}}^\top \hSigma_{11}^{-1} (Y-\bar{Y})\\
        &=\tw_n(y,Y)
    \end{align*}
    This completes the proof of Lemma \ref{lem: matrix}.
\end{proof}

\begin{lemma}
    \label{lem: arun22}
    If $X_1,\ldots,X_n$ are independent, mean zero random variables with $\norm{X_i}_{\psi_\alpha} \leq K$ for all $i \in [n]$ and some $\alpha>0$, then the following bounds hold true:
    \begin{itemize}[leftmargin=2em]
        \item $\alpha \in (0,1]$:
        \begin{equation*}
            \PP \rbr{ \frac{1}{n}\abs[\Big]{\sum_{i=1}^{n} X_i} > C_1 K \sqrt{\frac{t}{n}} + C_2  K\frac{t^{1/\alpha}}{n}} \leq 2 e^{-t} \quad \text{for all} \quad t \geq 0
        \end{equation*} 
        where $C_1,C_2>0$ are constants that depends only on $\alpha$.
        \item $\alpha > 1$
        \begin{equation*}
            \PP \rbr{ \frac{1}{n}\abs[\Big]{\sum_{i=1}^{n} X_i} > C_3 K \sqrt{\frac{t}{n}} + C_2  K \rbr{\frac{t}{n}}^{1/\alpha}} \leq 2 e^{-t} \quad \text{for all} \quad t \geq 0
        \end{equation*} 
        where $C_1,C_2>0$ are constants that depends only on $\alpha$.
    \end{itemize}
\end{lemma}
\begin{proof}
    See \citet[Theorem 3.1]{arun22}.
\end{proof}

\begin{lemma} \label{lem: Q}
    The following properties hold for $W(\cdot,\cdot)$ and $Q^*(\cdot)$:
    \begin{enumerate}[leftmargin=3em,label=\textup{(\arabic*)},ref=Part \textup{(\arabic*)}]
        \item For any $Q,S \in \cS^d_{++}$ and $X,Y \in \cS^d$, the following holds
        \begin{itemize}[leftmargin=*]
            \item $W^2(Q,S)$ is infinitely differentiable with respect to $Q$ and $S$ and we have
            \begin{align*}
                d_S W^2(S,Q)(X) &= \Finner{I-T_S^Q}{X}\\
                d^2_S W^2(S,Q)(X, Y) &= - \Finner{X}{dT_S^Q(Y)}
            \end{align*}
            Here $dT_S^Q: \cS^d \to \cS^d$ is a linear operator on symmetric matrices. Specifically, for any $H \in \cS^d$,
            \begin{align*}
                dT_S^Q(H) := -Q^{1/2}U^\top \Lambda^{-1/2} \delta \Lambda^{-1/2} U Q^{1/2} 
            \end{align*}
            where 
            \begin{itemize}[leftmargin=*]
                \item $U^\top \Lambda U$ is an eigenvalue decomposition of $Q^{1/2}SQ^{1/2}$ with $UU^\top =U^\top U = I_d$ and $\Lambda=\diag(\lambda_1,\ldots,\lambda_d)$;
                \item $\delta=(\delta_{ij})_{i,j \in [d]}$ with
                \begin{align*}
                    \delta_{ij} = \begin{cases}
                        \frac{\Delta_{ij}}{\sqrt{\lambda_i} + \sqrt{\lambda_j}}, & i, j \leq \rank(Q)\\
                        0 & \text{otherwise}
                    \end{cases}
                    , \qquad \text{and} \qquad
                    \Delta = U Q^{1/2} H Q^{1/2} U^\top.
                \end{align*}
            \end{itemize}

            \item if $Q,S \in \cS^d(M^{-1}, M)$, then we have
            \begin{align*}
                \max \cbr{\norm[\big]{d^k T^Q_S}, k=0,\ldots,3} \lesssim \poly{M}
            \end{align*}
        \end{itemize}
        \item Assume Assumption~\ref{assumption: X} and \ref{assumption: bdd_Q} holds. Then the following inequality holds
        \begin{align*}
            \gamma_{\Lambda}(\norm{x-\mu})^{-1} I_d \preceq & \; Q^*(x) \preceq  \gamma_{\Lambda}(\norm{x-\mu}) I_d, \qquad \forall x \in \supp X 
        \end{align*}
    \end{enumerate}
\end{lemma}
\begin{proof}
    see \citet[Theorem 25, 33]{XuLi2025} and \citet[Lemma A.2]{aap21}.
\end{proof}

\begin{lemma}
    \label{lem: xu25}
    Let $1 \leq L_n \leq C_L (\log n)^{1/2}$ for some constant $C_L>0$.
    Suppose Condition~\ref{assumption: X}-\ref{assumption: minimizer_local} hold. Then for any fixed constant $\tau > 0$, with probability at least $1-C_3 n^{-\tau}$, we have
        \begin{align*}
            \sup_{\norm{x -\mu}\leq L_n} \Fnorm{\hQ_n(x)-Q^*(x)} \leq  C_1 \frac{ \log^{C_2} {n}}{\sqrt{n}}
        \end{align*}
        for some constants $C_1, C_2, C_3>0$ independent of $n$.
\end{lemma}
\begin{proof}
    See \citet[Theorem 6]{XuLi2025}.
\end{proof}

For $\tau$ introduced in Theorem \ref{thm: null} and its estimator $\htau$ before Theorem \ref{thm: size}, we have the following properties:
\begin{lemma}[properties of $\tau$, $\htau$] \label{lem: tau}
    Let $x, x' \in \RR^p$, $q, q' \in \cS^d_{++}$,
    \begin{enumerate}[leftmargin=3em,label=\textup{(\arabic*)}, ref=Part \textup{(\arabic*)}]
        \item  \label{enum: tau_bdd}
        There exists a constant $C_1>0$ independent of $n$ such that the following holds
        \begin{align*}
            \Fnorm{\tau(x,q;x')} &\lesssim \sbr{\rbr{1 \vee \norm{x}} \rbr{1 \vee \Onorm{q} \vee \Onorm{q^{-1}}} \rbr{1 \vee \norm{x'}}}^{C_1}
        \end{align*}
        \item \label{enum: tau_null}
        Suppose Conditions \ref{assumption: X}-\ref{assumption: minimizer_local} hold. Assume the null hypothesis (\ref{eqn: null}) holds. Then we have
        \begin{align*}
            \EE \tau(X,Q; x)&= 0, \quad \forall x \in \RR^p
        \end{align*}
        Moreover, $\cK$ is a centered RKHS kernel.
        
        \item  \label{enum: tau_est}
        Suppose Conditions \ref{assumption: X}-\ref{assumption: minimizer_local} hold. Then there exists an event $\cE_0$ with probability at least $1-O(n^{-100})$ under which the following holds
        \begin{align*}
            \Fnorm{\htau(x,q;X_i) - \tau(x,q;X_i)} &\lesssim \frac{\polylog{n}}{\sqrt{n}} \sbr{\rbr{1 \vee \norm{x}} \rbr{1 \vee \Onorm{q} \vee \Onorm{q^{-1}}} }^{C_2}, \quad \forall i \in [n]
        \end{align*}
        and
        \begin{align*}
            &\abs{\inner{\htau(x,q;X_i)}{\htau(x',q';X_i)} - \inner{\tau(x,q;X_i)}{\tau(x',q';X_i)}}\\
            \lesssim  &\frac{\polylog{n}}{\sqrt{n}} \sbr{\rbr{1 \vee \norm{x}} \rbr{1 \vee \Onorm{q} \vee \Onorm{q^{-1}}} \rbr{1 \vee \norm{x'}} \rbr{1 \vee \Onorm{q'} \vee \Onorm{q'^{-1}}} }^{C_2}, \quad \forall i \in [n]
        \end{align*}
        and
        \begin{align*}
            \hat{\cK}_n(x,q;x',q') \lesssim \sbr{\rbr{1 \vee \norm{x}} \rbr{1 \vee \Onorm{q} \vee \Onorm{q^{-1}}} \rbr{1 \vee \norm{x'}} \rbr{1 \vee \Onorm{q'} \vee \Onorm{q'^{-1}}} }^{C_2}
        \end{align*}
        Here $C_2>0$ is a constant independent of $n$.
        
    \end{enumerate}
\end{lemma}

\begin{proof}
    \ref{enum: tau_bdd} and \ref{enum: tau_est} follow from the definition of $\tau$, Condition \ref{assumption: bdd_Q} and Lemma \ref{lem: Q}.

    \textit{Proof of \ref{enum: tau_bdd}.} Condition \ref{assumption: X}, \ref{assumption: bdd_Q} and Lemma \ref{lem: Q} imply the desired bound directly.

    \textit{Proof of \ref{enum: tau_null}.} The optimality condition at $x$ gives that
    \begin{align*}
        \EE_{(X,Q)} \vx^\top \Xi^{-1} \vX \rbr{T^{Q}_{Q^*(x)} - I_d} &= \EE w(x, X)\rbr{T^{Q}_{Q^*(x)} - I_d}= 0
    \end{align*}
    Hence $\EE \tau_0(X,Q;x) = 0$. Similarly, we have $\EE \tau_1(X,Q;x)=0$ by noticing that
    \begin{align*}
        \EE_{(X,Q)} \vy^\top \Xi_{11}^{-1} \vY \rbr{T^{Q}_{Q^*(x)} - I_d} &= \EE w(\tx, X)\rbr{T^{Q}_{Q^*(x)} - I_d} \overset{(i)}{=}  \EE w(\tx, X)\rbr{T^{Q}_{Q^*(\tx)} - I_d} = 0
    \end{align*}
    Here (i) holds under the null hypothesis.

    To show that $\cK$ is PSD, note that for any $m \geq 1$, $a_1,\ldots,a_m \in \RR$ and $(x_1,q_1),\ldots,(x_m,q_m) \in \RR^p \times \cQ$, we have
    \begin{align*}
        \sum_{i,j=1}^{m} a_i a_j \cK((x_i,q_i),(x_j,q_j)) &= \EE \; \Fnorm[\Big]{\sum_{i=1}^m a_i \tau(x_i,q_i;X)}^2 \geq 0
    \end{align*}
    The centeredness follows from $\EE \tau(X,Q;x)=0$ for any $x$.

    \textit{Proof of \ref{enum: tau_est}} can be proved by noticing that $\htau$ is a plug-in estimator of $\tau$. Then applying Lemma \ref{lem: xu25}, the smoothness of $Q \mapsto T^Q_{Q'}$ (Lemma \ref{lem: Q}) and the light tail of $X$ (Condition \ref{assumption: X}) gives the desired results.
\end{proof}

\begin{lemma}
    \label{lem: quantile_conv}
    Given one dimensional random variables $X,X_1,X_2,\ldots$. Let $F_X$ be the CDF of $X$, and $F_n$ the CDF of $X_n$. Fix $\alpha \in (0,1)$. Define quantiles
    \begin{align*}
        q_{n}:=F_n^{-1}(\alpha)=\inf\cbr{x: F_n(x) \geq \alpha}, \quad q:=F_X^{-1}(\alpha)=\inf\cbr{x: F_X(x) \geq \alpha}
    \end{align*}
    If $X_n \xrightarrow{d} X$ and $F_X$ is continuous and strictly increasing,
    then $q_{n} \rightarrow q$.

    Specifically, if $X=\sum_{j=1}^{\infty} \lambda_j w_j$ and $X_n=\sum_{j=1}^{\infty} \lambda_{n,j} w_j$ where $w_j$ are i.i.d. $\chi_1^2$ random variables, $\lambda_j \geq 0$, $\lambda_{n,j} \geq 0$ with $\sum_{j=1}^{\infty} \lambda_j < \infty$ and $\sum_{j=1}^{\infty} \lambda_{n,j} < \infty$. If $\sum_{i=1}^{\infty} \lambda_{n,j} \to \sum_{i=1}^{\infty} \lambda_j$ and $\sum_{i=1}^{\infty} |\lambda_{n,j} - \lambda_j|^2 \to 0$, then $X_n \xrightarrow{L_1} X$ and $q_n \to q$.
\end{lemma}

\begin{proof}
    For the first part of the lemma, fix any $\epsilon>0$. Since $F_X$ is continuous and strictly increasing, we have $F_X(q-\epsilon) < \alpha < F_X(q+\epsilon)$. By the convergence in distribution, we have $F_n(q-\epsilon) \to F_X(q-\epsilon)$ and $F_n(q+\epsilon) \to F_X(q+\epsilon)$. Hence for sufficiently large $n$, we have
    \begin{align*}
        F_n(q-\epsilon) < \alpha < F_n(q+\epsilon)
    \end{align*}
    Thus for sufficiently large $n$, we have $q_n \in [q-\epsilon, q+\epsilon]$. Since $\epsilon$ is arbitrary, we have $q_n \to q$.

    For the second part of the proof, note that $\lambda_j, \lambda_{n,j} \geq 0$,  hence for any $J \geq 1$, we have
    \begin{align*}
        \EE \abs{X_n - X} &\leq \sum_{j=1}^{\infty} \abs{\lambda_{n,j} - \lambda_j} \overset{(i)}{\leq} \sum_{j=1}^{J} \abs{\lambda_{n,j} - \lambda_j} + \sum_{j>J}^{\infty} \lambda_{n,j} + \sum_{j>J}^{\infty} \lambda_j \\
        &\leq \sum_{j=1}^{J} \abs{\lambda_{n,j} - \lambda_j} + 2\sum_{j>J}^{} \lambda_j + \abs[\Big]{\sum_{j>J}^{\infty} \lambda_{n,j} - \sum_{j>J}^{\infty} \lambda_j}\\
        &\leq 2\sum_{j=1}^{J} \abs{\lambda_{n,j} - \lambda_j} + 2\sum_{j>J}^{} \lambda_j + \abs[\Big]{\sum_{j=1}^{\infty} \lambda_{n,j} - \sum_{j=1}^{\infty} \lambda_j}\\
        &\leq 2\sqrt{J} \sum_{i=1}^{\infty} |\lambda_{n,j} - \lambda_j|^2 + 2\sum_{j>J}^{} \lambda_j + \abs[\Big]{\sum_{j=1}^{\infty} \lambda_{n,j} - \sum_{j=1}^{\infty} \lambda_j}
    \end{align*}
    Here (i) follows from the fact that $\lambda_{n,j}, \lambda_j \geq 0$. Hence we have
    \begin{align*}
        \limsup_{n \to \infty} \EE \abs{X_n - X} &\leq 2\sum_{j>J}^{} \lambda_j, \quad \forall J \geq 1
    \end{align*}
    Letting $J \to \infty$ yields $\EE \abs{X_n - X} \to 0$. Then the convergence of quantiles follows from the first part of the lemma.
\end{proof}

\begin{lemma} \label{lem: mtrx_sqrt}
    For any PSD matrix $\RR^{d\times d} \ni B \succeq 0$, we have
    \begin{align*}
        \sbr{B_{kk}}^{1/2} \geq \sbr[\big]{B^{1/2}}_{kk}, \quad \forall k \in [d].
    \end{align*}
    As a result, we have $\sum_{k\in [d]} [B_{kk}]^{1/2} \geq \tr(B^{1/2})$,
    with equality if and only if $B$ is diagonal.
\end{lemma}

\begin{proof}
    Let $B = U \Lambda U^\top$ be the eigen-decomposition of $B$ where $U \in \OO_{d}$ is orthogonal and $\Lambda = \diag(\lambda_1, \ldots, \lambda_d)$. Then for any $k \in [d]$, we have
    \begin{align*}
        B_{kk} = \sum_{j=1}^d U_{kj}^2 \lambda_j, \qquad\text{and} \qquad [B^{1/2}]_{kk} = \sum_{j=1}^d U_{kj}^2 \lambda_j^{1/2}.
    \end{align*}
    Since $U$ is orthogonal, we have $\sum_{j=1}^d U_{kj}^2 = 1$. Hence by the strict concavity of the square root function and Jensen's inequality, we have
    \begin{align*}
        [B_{kk}]^{1/2} \geq [B^{1/2}]_{kk},
    \end{align*}
    with equality if and only if $\lambda_j = \lambda_{j'}$ for all $j,j'$ such that $U_{kj}^2 >0$ and $U_{kj'}^2 >0$. We denote such value as~$\theta_j$. Note that at equality, we have
    \begin{align*}
        [B^2]_{kk} = [U\Lambda^2 U^\top]_{kk}= \sum_{j=1}^d U_{kj}^2 \lambda_j^2 = \theta_k^2 = [B]_{kk}^2.
    \end{align*}
    Hence when $\sum_{k\in [d]} [B_{kk}]^{1/2} = \tr(B^{1/2})$, $B$ is diagonal.
\end{proof}

\subsection{Reproducing kernel Hilbert space}
\label{sec: tech_rkhs}

\input{RKHS.tex}

\subsection{U- and V-Statistics}
\input{u_stat.tex}


%% file: RKHS.tex

Let $(\cX,\rho)$ be a metric space. A symmetric bivariate function $\cK: \cX \times \cX \to \RR$ is positive semidefinite (PSD) if, for all integers $n \geq 1$ and any collection of points
$\cbr{x_i}_{i=1}^n \subset \cX$, the $n \times n$ matrix with entries $\cK(x_i, x_j)$ is positive semidefinite. 
Every PSD kernel function $\cK$ induces a unique Hilbert space $\cH_{\cK}$ that has the following reproducing property: for any $x \in \cX$, the function $\cK(\cdot,x)$ belongs to $\cH_{\cK}$ and satisfies
\begin{equation*}
    \inner{\cK(\cdot,x)}{f}_{\cH_{\cK}} = f(x), \quad \forall f \in \cH_{\cK}.   
\end{equation*}
This space $\cH_{\cK}$ is called the reproducing kernel Hilbert space (RKHS) associated with $\cK$.
The following lemma characterizes when the RKHS $\mathcal{H}_{\cK}$ consists entirely of continuous functions.
\begin{lemma} \label{lem: berlinet}
    Every function $h \in \cH_{\cK}$ is continuous if and only if $\cK$ satisfies the following two conditions
    \begin{itemize}[leftmargin=2em]
    \item $\forall x \in \cX$, $\cK(., x)$ is continuous.
    \item $\forall x \in \cX$, $\exists r > 0$, such that the function
        \begin{align*}
            \cX &\to \RR^+\\
            y &\mapsto \cK(y,y)
        \end{align*}
        is bounded on the open ball $B(x, r)$. 
    \end{itemize}
\end{lemma}
\begin{proof}
    See \citet[Chapter 1, Theorem 17]{berlinet}.
\end{proof}

Now let $\mu$ be a measure on the Borel sets of $\cX$. The kernel $\cK$ is said to be square integrable with respect to $\mu$ if $\int\int_{\cX \times \cX} \cK^2(x,y)d\mu(x) d\mu(y)< + \infty$. When $\cK$ is square integrable, we can define the integral operator~$\mathscr{T}_{\cK}: L^2(\cX;\mu) \to L^2(\cX;\mu)$ (associated with $\cK$) by
\begin{equation*}
    (\mathscr{T}_{\cK} f) (\cdot) := \int \cK(\cdot, y) f(y) d\mu(y), \quad \forall f \in L^2(\cX; \mu).
\end{equation*}
This operator $\mathscr{T}_{\cK}$ is bounded and compact, with operator norm  satisfying
\begin{equation*}
    \Onorm{\cK}^2 \leq \int\int_{\cX \times \cX} \cK^2(x,y)d\mu(x) d\mu(y) = \norm{\cK}_{\mathrm{HS}}^2,
\end{equation*}
where $\norm{\cdot}_{\mathrm{HS}}$ denotes the Hilbert-Schmidt norm. 


Throughout this paper, we assume that the kernel $\cK$ is PSD, continuous and square integrable with respect to $\mu$.
When $\cX$ is locally compact, Mercer's theorem provides a spectral decomposition.
\begin{theorem}[Mercer] \label{thm: mercer}
    Let $\cX$ be locally compact and $\cK: \cX \times \cX \to \RR$ be a symmetric, PSD and continuous kernel. Then there exist:
    \begin{itemize}[leftmargin=2em]
        \item A sequence of eigenfunctions $\cbr{\phi_j}_{j=1}^\infty$ forming an orthonormal basis of $L^2(\cX;\mu)$,
        \item Non-negative eigenvalues $\cbr{\lambda_j}_{j=1}^\infty$ in non-increasing order
    \end{itemize}
    such that $\mathscr{T}_{\cK} \phi_j= \lambda_j \phi_j$ for all $j \geq 1$, and
    \begin{equation*}
        \cK(x,y) = \sum_{j=1}^{\infty} \lambda_j \phi_j(x) \phi_j(y),
    \end{equation*}
    where the series converges absolutely and uniformly on compact subsets of $\cX \times \cX$.
\end{theorem}
\begin{proof}
    See \citet[Theorem 4.6.5]{hsing}.
\end{proof}

The spectral decomposition induces a natural feature map $\Phi: \cX \to \ell^2(\NN)$ defined by
\begin{align*}
    x \mapsto \Phi(x) = 
    \begin{pmatrix}
        \sqrt{\lambda_1} \phi_1(x) &
        \sqrt{\lambda_2} \phi_2(x) &
        \cdots
    \end{pmatrix}^\top.
\end{align*}
This map satisfies $\cK(x,y) = \inner{\Phi(x)}{\Phi(y)}_{\ell^2(\NN)}$.

When the kernel is trace-class, meaning  $\int \cK(x,x)d\mu(x) < \infty$, we have the embedding $\cH_{\cK} \subset L^2(\cX;\mu)$. Indeed, for any $h \in \cH_{\cK}$, we have
\begin{align*}
    \abs{{h(x)}} = \abs{\inner{h}{\cK_x}_{\cH_{\cK}}} \leq \norm{h}_{\cH_{\cK}} \sqrt{\cK(x,x)},
\end{align*}
which implies that $h \in  L^2(\cX;\mu)$.

We refer readers to the monographs by \citet{berlinet,hsing} and \citet{wainwright}, as well as \citet{lax}, for detailed expositions of RKHS theory.

%% file: u_stat.tex
Let $\psi: \cX^k \to \RR$ be a symmetric kernel of order $k \geq 1$ and $X_1,\ldots,X_n \overset{\text{iid}}{\sim} P \in \cP(\cX)$. Then the U-statistic associated with $\psi$ is
\begin{align*}
    U_n = \binom{n}{k}^{-1} \sum_{1 \leq i_1 < i_2 < \cdots < i_k \leq n} \psi\rbr{X_{i_1}, X_{i_2}, \ldots, X_{i_k}}.
\end{align*}
The V-statistic associated with $\psi$ is
\begin{align*}
    V_n = \frac{1}{n^k} \sum_{i_1,\ldots, i_k=1}^n \psi\rbr{X_{i_1}, X_{i_2}, \ldots, X_{i_k}}.
\end{align*}

\begin{lemma}
    \label{lem: v_decomp}
    When $\psi$ is a symmetric kernel of order 3, we have
    \begin{align*}
        n^3 V_n = 6 \binom{n}{3} U_n^{(3)} + 6 \binom{n}{2} U_n^{(2)} + \binom{n}{1} U_n^{(1)},
    \end{align*}
    with kernels
    \begin{align*}
        \phi_{(3)}(x_1,x_2,x_3) &=  \psi(x_1,x_2,x_3),\\
        \phi_{(2)}(x_1,x_2) &= \frac{1}{2} \rbr{\psi(x_1,x_1,x_2) + \psi(x_1,x_2,x_2)},\\
        \phi_{(1)}(x_1) &= \psi(x_1,x_1,x_1).
    \end{align*}
\end{lemma}
\begin{proof}
    See \citet[Example 2, \S 4.2]{lee1990u}.
\end{proof}

Define for $c=1,2,\ldots,k$ the conditional expectations
\begin{align*}
    \psi_c(x_1,\ldots,x_c) = \EE \sbr{\psi(x_1,\ldots,x_c,X_{c+1},\ldots,X_k)}
\end{align*}
and their variances
\begin{align*}
    \sigma_c^2 = \Var \sbr{\psi_c(X_1,\ldots,X_c)}.
\end{align*}
Define also $\sigma_0^2=0$. Let $\theta=\EE U_n$, then we have the following results.
\begin{lemma}
    \label{lem: u_asym_nondeg}
    If $\sigma_1^2>0$, then $n^{1/2}(U_n - \theta) \xrightarrow{d} \cN(0, k^2 \sigma_1^2)$.
\end{lemma}
\begin{proof}
    See \citet[Theorem 1, \S 3.1]{lee1990u}.
\end{proof}

\begin{lemma}
    \label{lem: u_asym_deg}
    Let $U_n$ be a U-statistic based on a kernel $\psi(x_1,\ldots,x_k)$, and suppose $\sigma_1^2=0$, $\sigma_2^2>0$. Then
    \begin{align*}
        n(U_n - \theta) \xrightarrow{d} \binom{k}{2}\sum_{\nu=1}^\infty \lambda_\nu (Z_{\nu}^2 -1),
    \end{align*}
    where $Z_\nu$ are i.i.d. $N(0,1)$ and the $\lambda_\nu$ are the eigenvalues of the integral function
    \begin{align*}
        \int_{-\infty}^{\infty} \sbr{\psi_2(x_1,x_2) - \psi_1(x_1) - \psi_1(x_2) + \theta} f(x_2) dP(x_2) = \lambda f(x_1).
    \end{align*}
\end{lemma}
\begin{proof}
    See \citet[Corollary 1, \S 3.2]{lee1990u}.
\end{proof}

\begin{lemma}
    \label{lem: u_var}
    The function $var(nU_n)$ is decreasing in $n$.
\end{lemma}

\begin{proof}
    See \citet[Theorem 5, \S 1.3]{lee1990u}.
\end{proof}